\documentclass{naturep}
\usepackage[utf8]{inputenc}
\usepackage{lineno}
\usepackage{setspace}
\usepackage{color}
\usepackage{graphicx}
\usepackage{booktabs}
\usepackage{amssymb}
\usepackage{amsmath}
\usepackage{hyperref}
\usepackage{adjustbox}
\usepackage{subcaption}
\usepackage{tablefootnote}
\usepackage[margin=0.6in]{geometry}
\usepackage{verbatim}
\usepackage{caption}
\usepackage{paralist}
\usepackage{multirow}

\newcommand{\ours}{\textsc{VORTEX}}

\newcommand\Heading[1]{
  \noindent\textbf{\Large{#1}}
}

\newcommand\hheading[1]{
  \noindent\textbf{#1}
}

\title{\raggedright{\textbf{AI-driven Non-Destructive 3D Spatial Transcriptomics}}}
\title{\textbf{AI-driven 3D Spatial Transcriptomics}}
\author{\raggedright Cristina Almagro-Pérez$^{\ast,1,2,3,4}$, Andrew H. Song$^{\ast,1,2,3}$, Luca Weishaupt$^{1,2,3,4}$, Ahrong Kim$^{1,5}$, Guillaume Jaume$^{1,2,3}$, Drew F.K. Williamson$^{1,6}$, Konstantin Hemker$^{1, 7}$, Ming Y. Lu$^{1,2,3,8}$,  Kritika Singh$^{9}$, Bowen Chen$^{1,2,3}$, Long Phi Le$^{1}$, Alexander S. Baras$^{10,11}$, Sizun Jiang$^{12,13,14}$, Ali Bashashati$^{15,16}$, Jonathan T.C. Liu$^{17}$, and Faisal Mahmood$^{\dagger,1,2,3,18}$}
    
\makeatletter
\let\saved@includegraphics\includegraphics
\AtBeginDocument{\let\includegraphics\saved@includegraphics}

\makeatother

\begin{document}
\maketitle
\begin{affiliations}
 \item Department of Pathology, Mass General Brigham, Harvard Medical School, Boston, MA, USA
 \item Cancer Program, Broad Institute of Harvard and MIT, Cambridge, MA, USA
 \item Data Science Program, Dana-Farber Cancer Institute, Boston, MA, USA
\item Harvard-MIT Division of Health Sciences and Technology, Massachusetts Institute of Technology, Cambridge, MA, USA
\item Department of Pathology, Pusan National University, Busan, South Korea
\item Department of Pathology and Laboratory Medicine, Emory University School of Medicine, Atlanta, GA, USA
\item Department of Computer Science \& Technology, University of Cambridge, Cambridge, UK
\item Department of Electrical Engineering and Computer Science, Massachusetts Institute of Technology, Cambridge, MA, USA
\item Rutgers Robert Wood Johnson Medical School, New Brunswick, NJ
\item Department of Pathology, Johns Hopkins Hospital, Baltimore, MD, USA
\item Department of Biomedical Engineering, Johns Hopkins Hospital, Baltimore, MD, USA 
\item Broad Institute of Harvard and MIT, Cambridge, MA, USA
\item Center for Virology and Vaccine Research, Beth Israel Deaconess Medical Center, Harvard Medical School, Boston, MA, USA
\item Department of Pathology, Dana Farber Cancer Institute, Boston, MA, USA
\item Department of Pathology and Laboratory Medicine, University of British Columbia, Vancouver, BC, Canada
\item School of Biomedical Engineering, University of British Columbia, Vancouver, BC, Canada
\item Department of Mechanical Engineering, Bioengineering, and Laboratory Medicine \& Pathology, University of Washington, Seattle, WA, USA
\item Harvard Data Science Initiative, Harvard University, Cambridge, MA, USA
\item[$^{\ast}$] Equal contribution\\
\textbf{$\dagger$ Corresponding author}: Faisal Mahmood (FaisalMahmood@bwh.harvard.edu)
\end{affiliations}

\clearpage

\Heading{Abstract}

\begin{spacing}{1.2}
\noindent\textbf{
\noindent A comprehensive three-dimensional (3D) map of tissue architecture and gene expression is crucial for illuminating the complexity and heterogeneity of tissues across diverse biomedical applications\cite{erturk2024deep}. However, most spatial transcriptomics (ST) approaches remain limited to two-dimensional (2D) sections of tissue\cite{staahl2016visualization, marx2021method, moses2022museum}. Although current 3D ST methods hold promise, they typically require extensive tissue sectioning, are complex, are not compatible with non-destructive 3D tissue imaging technologies, and often lack scalability\cite{wang2023construction, tang2024search, schott2024open, mo2024tumour}. Here, we present VOlumetrically Resolved Transcriptomics EXpression (VORTEX), an AI framework that leverages 3D tissue morphology and minimal 2D ST to predict volumetric 3D ST. By pretraining on diverse 3D morphology–transcriptomic pairs from heterogeneous tissue samples and then fine-tuning on minimal 2D ST data from a specific volume of interest, VORTEX learns both generic tissue-related and sample-specific morphological correlates of gene expression. This approach enables dense, high-throughput, and fast 3D ST, scaling seamlessly to large tissue volumes far beyond the reach of existing 3D ST techniques. By offering a cost-effective and minimally destructive route to obtaining volumetric molecular insights, we anticipate that VORTEX will accelerate biomarker discovery and our understanding of morphomolecular associations and cell states in complex tissues.  Interactive 3D ST volumes can be viewed at \url{https://vortex-demo.github.io/}. 
}
\end{spacing}

\clearpage
\begin{spacing}{1.35}
\Heading{Introduction}

Understanding intratumoral morphological and molecular heterogeneity in human tissue is critical for developing personalized treatments and predicting therapeutic responses\cite{song2023artificial, marusyk2020intratumor, vitale2021intratumoral, fu2021spatial, bagaev2021conserved, arora2023spatial}.
Spatially-resolved transcriptomics (ST) provides expression profiles for many genes at high spatial resolution on two-dimensional (2D) tissue sections\cite{staahl2016visualization, rao2021exploring, rodriques2019slide, marx2021method, moses2022museum, palla2022spatial, ren2024spatial}. 
By analyzing ST with its associated high-resolution tissue morphology, researchers can holistically characterize intratumoral heterogeneity with multimodal views, investigate how changes in molecular profile influence underlying morphology, and vice versa.

The molecular and morphological traits captured within a 2D tissue section only represent a small fraction of the tissue volume and the patient\cite{liu2021harnessing, song2023artificial, braxton20243d, mo2024tumour, erturk2024deep, wang20243d, mathur2024glioblastoma}. Therefore, increasing attention has recently been directed toward extending molecular characterization from within a single tissue section to many adjacent tissue sections or across a larger volume. Recent three-dimensional (3D) pathology studies, fueled by substantial advances in high-resolution 3D tissue imaging modalities such as micro computed tomography (microCT) or open-top light-sheet microscopy\cite{withers2021x, liu2021harnessing, bishop2024end} showed that 3D morphological characterization can lead to better patient prognostication or cancer biomarker discovery\cite{song2024analysis, xie2022prostate, erturk2024deep}. 
Parallel efforts have been devoted to creating 3D molecular atlases of tissue, either with \textit{in situ} sequencing\cite{wang2018three, wang2021easi, fang2024three, sui2024scalable,doi:10.1126/science.adq2084} or by registering serial sections of 2D ST data meticulously obtained from a single tissue volume\cite{dong2022deciphering, vickovic2022three, zeira2022alignment, zhou2023integrating,wang2023construction, lin2023multiplexed, schott2024open, tang2024search,shu2024efficient}. While promising, \textit{in situ} approaches remain limited in terms of capture area and depth, and require long processing times (e.g., capture area of $3\times 4\, mm^2$ and depth of $\sim 200 \mu m$). Serial section-based approaches provide discontinuous coverage along the axial dimension (i.e., 2.5D ST characterization) of thick tissues. Such approaches are impractical for scaling to whole-volume transcriptomic profiling in terms of cost and effort, with up to several days of processing for a single clinical sample.

An AI-based computational predictive framework offers an attractive alternative for characterizing the molecular landscapes of tissue specimens.
Evidence of the close relationship between spatially variable genes and underlying tissue morphology \cite{edsgard2018identification, svensson2018spatialde, sun2020statistical, binder2021morphological, ash2021joint, hu2021spagcn, song2023artificial} suggests that such \textit{morphomolecular} links can be modeled, especially when leveraging the powerful capabilities of deep learning. Coupled with the increasing availability of paired high-resolution 2D tissue images and 2D ST data\cite{moses2022museum, marx2021method, jaume2024hest, chen2024stimagekm}, recent AI-based frameworks have demonstrated success in directly learning morphomolecular links and predicting transcript expression and localization from morphological data alone\cite{he2020integrating, hu2021spagcn, bergenstraahle2022super, xie2024spatially, chung2024accurate, coleman2024unlocking, kueckelhaus2024inferring, zhang2024inferring, lee2024Path}. 
However, these models are exclusively restricted to 2D tissue sections, and designing 3D ST prediction frameworks based on 3D tissue morphology necessitates additional consideration. 

Here, we present an AI-based computational framework called $\ours$, \textbf{VO}lumetrically \textbf{R}esolved \textbf{T}ransciptomics \textbf{EX}pression. $\ours$ enables scalable and efficient 3D ST prediction of large volumes from 3D pathology datasets.
$\ours$ is pretrained on 3D morphology and 2D ST data pairs from diverse volumes of the same cancer and is further fine-tuned on data pairs from a specific volume of interest. This learning paradigm takes advantage of both generic morphomolecular links prevalent across diverse volumes and volume-specific links that are difficult to learn due to inter-volume heterogeneity.
A distinguishing feature of $\ours$ is its ability to adapt to different 3D imaging modalities and tissue sizes for 3D ST prediction.
To handle diverse non-destructive 3D tissue imaging approaches\cite{withers2021x, palermo2025investigating, glaser2017light, bishop2024end} with ST data confined to 2D tissue sections, $\ours$ performs cross-modal registration and integration between 3D tissue images, 2D tissue images, and 2D ST. Furthermore, $\ours$ can easily scale up to performing ST predictions in large tissue volumes, vastly exceeding typical ST capture areas with little additional cost and processing time. To highlight the versatility of $\ours$, we also demonstrate it for ST predictions in 2.5D tissue images constructed from serial 2D tissue sections, a commonly utilized approach that is compatible with current histopathology workflows.

\Heading{Results}

\hheading{AI-based 3D ST prediction with $\ours$}

$\ours$ is a deep learning model that enables 3D ST prediction for 3D tissue images captured with high-resolution non-destructive 3D pathology modalities\cite{liu2021harnessing, song2024analysis, wang20243d}, which are anticipated to become more common as a complementary approach to serial tissue sectioning\cite{kiemen2022coda, braxton20243d}. Non-destructive imaging preserves tissues for downstream assays, thereby facilitating morphomolecular analyses \cite{bishop2024end, li2023feasibility}.
Upon modeling the link between 3D tissue morphology and corresponding spatially resolved gene expression profiles in local 3D regions (or patches), $\ours$ processes each 2D section of the test volume (or volume of interest, VOI), and its neighboring sections together to provide 2D ST predictions for all sections. This stack of predicted 2D ST images constitutes a 3D ST prediction for thick tissue specimens, accommodating any tissue volume size.
The 2D ST measurements are obtained with the Visium platform\cite{staahl2016visualization, rao2021exploring}, which captures aggregate gene expression profiles from several neighboring cells for each sequencing spot (55-micron spot size). Small 2D or 3D tissue image patches centered around each sequencing spot represent the localized tissue morphology, providing the morphology and transcriptomics data pairs that $\ours$ operates on.

\begin{figure*}
\centering 
\includegraphics[width=\textwidth]{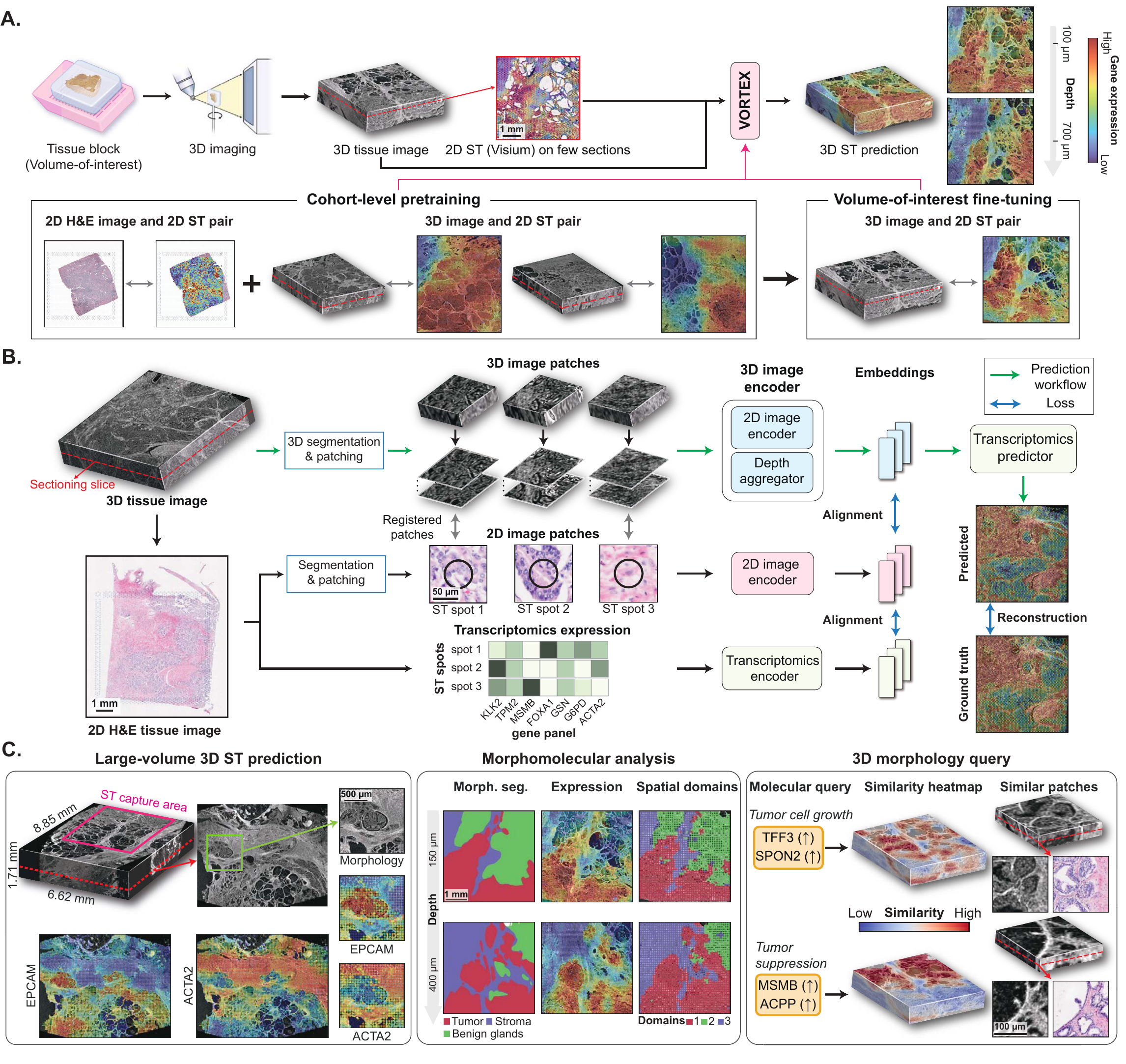}
\caption{\textbf{Overview of $\ours$}. 
\textbf{(a)} Workflow of 3D ST prediction with $\ours$ on a test volume (or volume of interest, VOI). $\ours$ provides efficient whole volume 3D ST prediction for gene sets of interest based on the 3D tissue images and ST measurements. A 3D tissue image is obtained with a non-destructive 3D imaging modality (microCT chosen as an illustrative example). ST is performed on a few 2D tissue sections from the same tissue volume (Visium chosen as an illustrative example).  $\ours$ is trained over two stages. It is first pretrained on a disease-specific cohort of 2D (or 3D) tissue images and 2D ST data pairs. It is further fine-tuned on data pairs of 2D (or 3D) tissue images and 2D ST acquired from the VOI. $\ours$ can also be extended to 2.5D tissue images comprised of serial tissue sections.
\textbf{(b)} Illustration of $\ours$ architecture. All deep learning components of $\ours$ are trained with a combination of \textit{ST reconstruction} and \textit{cross-modal alignment} loss. The green arrows indicate the prediction workflow of $\ours$ once trained.
\textbf{(c)} Applications for $\ours$ on efficient large 3D ST prediction, joint morphology and ST analysis, and 3D morphology query. ST: Spatial transcriptomics. Morph. Seg.: Morphological segmentation.
}
\label{fig:main}
\end{figure*} 

The predictive capacity of $\ours$ is further enhanced by fine-tuning the model on 2D ST captured from tissue sections within the VOI. While the model input is still the same 3D tissue image, the ST prediction stage can benefit from the incorporation of subtle volume-specific attributes (\textbf{Figure~\ref{fig:main}A}). 
The structure of $\ours$ offers two avenues for data scaling. First, scaling the training dataset to include different tissue volumes of the same disease or tissue type increases the training data scale and the statistical power of transcriptomics analyses\cite{stuart2019comprehensive, korsunsky2019fast}. This also helps $\ours$ extract generic morphomolecular signatures for relevant tissue types that are preserved across heterogeneous examples of a tissue type/disease of interest. Next, incorporating the 3D morphological context and 2D ST measurements from a VOI through a principled model-based approach increases the predictive performance. To facilitate concurrent analysis for correspondence between the distinct tissue morphologies and spatial transcriptomic expression patterns, we also provide an efficient AI-based 3D morphological segmentation mask constructed from annotations on 2D tissue sections\cite{kirillov2023segment}.
 
To ensure good predictive performance, we first pretrain $\ours$ on 2D hematoxylin and eosin (H\&E) tissue sections and 2D ST data pairs. This step allows the model to accurately learn the relationship between transcriptomic expression and its 2D morphological correlates, which forms the basis for 3D ST prediction. $\ours$ is composed of four main components, the 2D and 3D \textit{image encoders} for extracting low-dimensional embedding of 2D and 3D image patches, the \textit{transcriptomics encoder} for extracting low-dimensional embedding of ST, as well as the \textit{transcriptomics predictor} for predicting ST from patch embeddings (\textbf{Figure~\ref{fig:main}B}).
To extract representative histology patch and transcriptomics embeddings, the image and transcriptomic encoders are initialized with a pathology foundation model CONCH\cite{lu2024visual} (pretrained on millions of histology image and text pairs) and a single-cell foundation model (scGPT)\cite{cui2024scgpt} (pretrained on single-cell transcriptomics data from millions of cells of various cancer types), respectively. The entire model is trained in a multi-task setting, combining the contrastive loss to align the image and transcriptomic embeddings and the reconstruction loss to predict ST from the image embedding\cite{yu2022coca}.
Once pretrained, $\ours$ is extended to integrate the 3D morphological context, by employing a lightweight module on top of the image encoder to aggregate neighboring tissue regions at different depths. Further details on the model architecture can be found in \textbf{Online Methods} section \textbf{Model Architecture}.

$\ours$ presents a fundamentally different mechanism for 3D ST prediction from other frameworks. Specifically, existing works meticulously align multiple 2D tissue sections with 2D ST measurements from the same volume to construct a 2.5D ST heatmap\cite{dong2022deciphering, zeira2022alignment, zhou2023integrating,wang2023construction, tang2024search, shu2024efficient,li2024high, lin2024bridging}.
Consequently, obtaining a 3D or 2.5D ST profile of a sample still results in high costs and turnaround time from having to sequence a large number of tissue sections. Moreover, these approaches do not have extrapolation capacity across the plane, restricting the predicted ST coverage to within the ST planar capture area. In contrast, $\ours$ operates on continuous 3D tissue morphology as input, based on the underlying morphomolecular links learned by the models. $\ours$ can provide 3D ST for each volume with orders of magnitude less cost and time because it requires significantly fewer ST measurements from a VOI for fine-tuning. Here we show that fine-tuning the model on a single 2D ST capture area from the VOI can help predict the ST profile for any other tissue regions outside the capture area, across the plane and at varying depths. Consequently, $\ours$ can operate on tissue volumes of any size (\textbf{Figure~\ref{fig:main}C}). 

\hheading{3D ST prediction for prostate cancer}

To evaluate the performance of $\ours$ on clinical tissue specimens, we apply the model to 3D ST prediction of prostate cancer volumes.
We use microCT\cite{withers2021x, palermo2025investigating} with an isotropic resolution of $4~\mu m$/voxel to acquire 3D high-resolution images for 11 tissue volumes from 11 different patients from Mass General Brigham, with each image covering $7 \times 11 \times 3\, mm^3$ field-of-view. After imaging each volume with microCT, we obtain both Visium ST and H\&E tissue images from those volumes.  For five of the volumes, we obtained two pairs of tissue sections spaced apart by 250~$\mu m$.  For the other six volumes, we obtained a single section each, resulting in a total of 16 sections and 65,715 training pairs (morphology patch with corresponding 2D ST spot) (\textbf{Figure~\ref{fig:main}A}). We additionally curated a public dataset of 2D H\&E sections with corresponding 2D Visium and Spatial Transcriptomics ST data from various studies, encompassing 49 sections (72,832 spots).
The data integration shows that $\ours$ can bridge the modality gaps between 2D ST, 2D H\&E tissue images, and 3D tissue images. 
Additionally, it shows the scale of data that typical 3D non-destructive tissue imaging modalities produce (typically on the order of several hundred 2D image sections) can be handled by $\ours$\cite{song2024analysis, coleman2024unlocking, erturk2024deep}. Further details on the prostate cancer dataset can be found in \textbf{Extended Data Table~\ref{tab:dataset}} and \textbf{Online Methods} in section \textbf{Datasets}.

\begin{figure*}[]
\centering
\includegraphics[width=\textwidth]{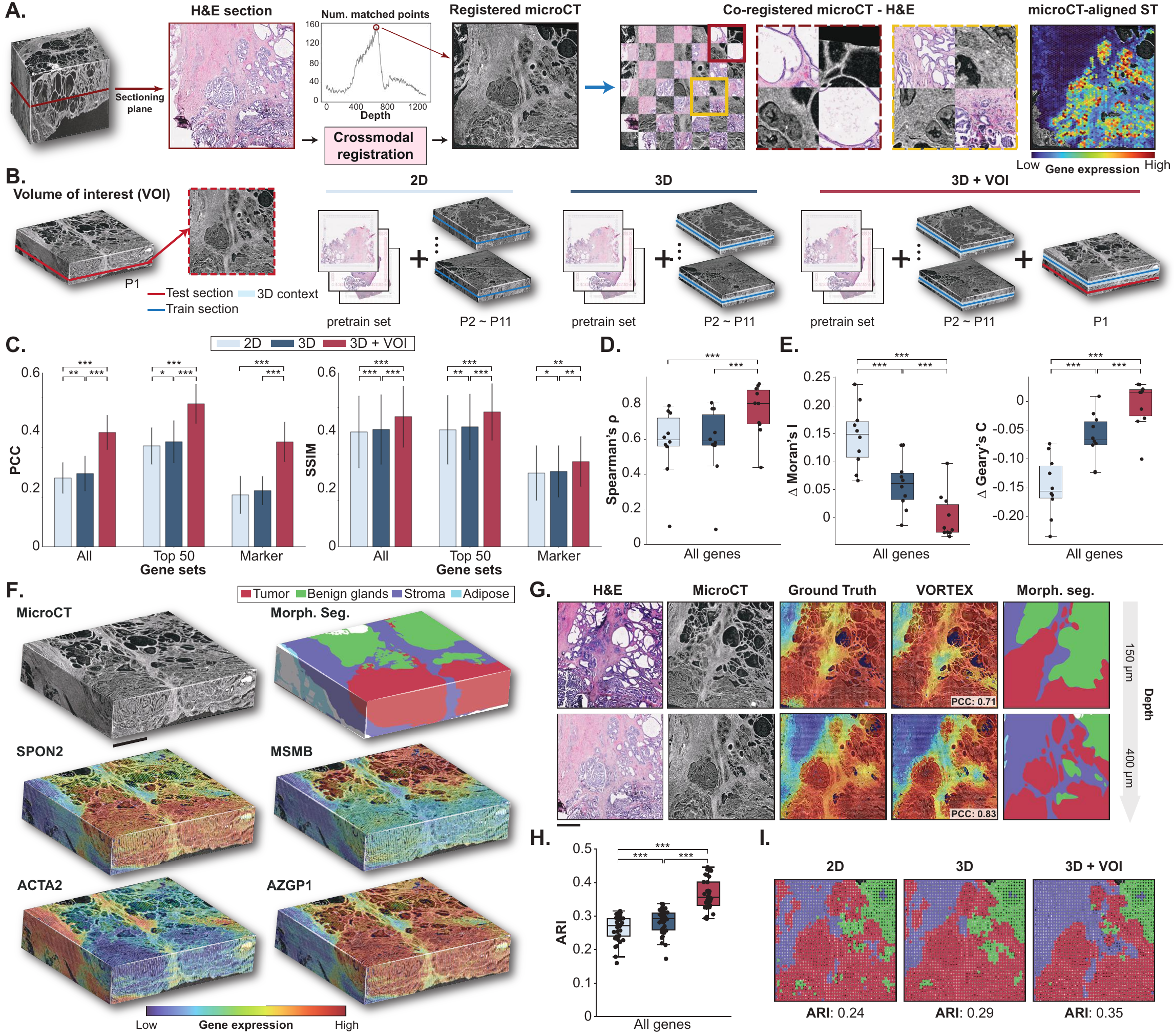}
\caption{\textbf{$\ours$ analysis on prostate cancer}. 
\textbf{(a)} Cross-modal registration between the 3D microCT tissue image (4$\mu m$/voxel) and the H\&E-stained tissue sections with 2D ST.
Checkerboard visualization of co-registered microCT and H\&E images.
\textbf{(b)} Schematics for different training scenarios: 2D image and 2D ST pairs (\textit{2D}), adding 3D image and 2D ST pairs  (\textit{3D}), and further adding 3D image and 2D ST pairs from the VOI (\textit{3D + VOI}).
\textbf{(c)} PCC and SSIM between the predicted and the measured expression for three gene sets for five patients: All Genes (264 genes), the top 50 highly predictive genes, and marker genes. Error bars indicate one standard deviation from the mean, over ten sections across five patients.
\textbf{(d)} Spearman's $\rho$ between the variance of measured and predicted expressions across all genes. Each black dot represents a tissue section. 
\textbf{(e)} Difference in Moran's I and Geary's C metric between measured and predicted expressions aggregated across all genes in five patients.
\textbf{(f)} 3D ST prediction heatmap for select genes, with 3D morphological segmentation masks. Additional examples can be found in \textbf{Extended Data Figure~\ref{fig:ext_prostate_3D}}.
\textbf{(g)} Cross-section visualizations of morphology, measured and predicted ST for \textit{AZGP1}, and morphological segmentation masks.
\textbf{(h)} ARI metrics across the depth of tissue volume. The ARI metric is measured between the segmentation mask and predicted spatial domains. Each black dot represents a tissue section.
\textbf{(i)} The spatial domains identified by $\ours$ for the plane at $400 \mu m$. All scalebars are $1\,mm$.  Statistical significance was assessed with the Wilcoxon signed-rank test. $^{\ast}p\leq 0.05$, $^{\ast\ast}p\leq 0.01$, $^{\ast\ast\ast}p\leq 0.001$. Whiskers extend to data points within 1.5$\times$ the interquartile range. VOI: Volume of interest. PCC: Pearson Correlation Coefficient. SSIM: Structural Similarity Index Measure. ARI: Adjusted Rand Index. 
}
\label{fig:prostate}
\end{figure*}

As the first preprocessing step for $\ours$, we perform cross-modal registration between 2D and 3D tissue images. While the 2D ST measurements are spatially registered to 2D H\&E tissue sections by default, registering the sectioned images to the microCT tissue volume is often non-trivial.
We use landmark-based registration pipelines\cite{bay2006surf, fischler1981random, gatenbee2023virtual,chicherova2014histology} to estimate the depth and the angle at which tissue sections were cut from the tissue volume. Using the estimated parameters, we register 2D tissue image and the ST sequencing spots onto the tissue volume (\textbf{Figure~\ref{fig:prostate}A}).
As a result, all 16 tissue sections with 2D ST measurements are registered to respective tissue volumes.
We then create a 2D H\&E patch of $112\times112~\mu m$ ($112\times 112$ pixels) and a 3D microCT patch of $448\times 448 \times 84~\mu m$ ($112 \times 112 \times 21$ pixels), centered around each ST spot as the corresponding 2D and 3D morphological context. The depth of the 3D patch ensures that large benign prostatic glands are covered. Further details on cross-modal registration and data preprocessing can be found in \textbf{Online Methods} in section \textbf{3D image data preprocessing}.

We evaluate the predictive performance of $\ours$ for five volumes for which we obtained two 2D ST sections at different depths (samples P1$\sim$P5). Performance is assessed using a leave-one-volume-out approach, with one volume assigned as VOI and the rest as the training set. Three different training scenarios are examined for delineating data-related effects: \textit{2D}, \textit{3D}, and \textit{3D+VOI} (\textbf{Figure~\ref{fig:prostate}B}). In the \textit{2D} scenario, the training set combines public 2D H\&E and 2D ST data pairs with internal 2D microCT image and 2D ST pairs. For the \textit{3D} scenario, 3D microCT images, instead of 2D microCT images, are used to provide 3D morphological context for ST prediction. The \textit{3D + VOI} scenario further fine-tunes the model by incorporating an additional data pair from the VOI, in the form of a single 2D ST section. The remaining ST section not used for fine-tuning (the second ST section for each specimen) is used for evaluation, with the roles subsequently switched to yield two predictions for each VOI. These experiments are designed to elucidate the benefits of 3D morphological context (\textit{3D} vs. \textit{2D}) and integrating VOI-specific training pairs on top of training pairs from generic volumes of the same cancer/tissue type (\textit{3D + VOI} vs. \textit{3D}).
We use an average of Pearson Correlation Coefficient (PCC) and Structural Similarity Index Measure (SSIM) across two sections for which ST profiles are available. PCC is computed for an individual gene between the measured and predicted expressions across all spots in a plane, with a higher correlation indicating better ST predictions. SSIM assesses the structural similarity between the measured and predicted expressions by treating them as images, with a higher value indicating a more similar spatial structure\cite{wang2004image, zhang2024inferring, wang2025benchmark}.
To assess the robustness of $\ours$ to the choice of different genes for prediction, we choose three genes sets: 19 genes curated from Oncotype DX\cite{klein201417, cullen2015biopsy} and Decipher\cite{klein2016decipher} (\textit{marker genes}, \textbf{Extended Data Table~\ref{tab:marker_genes}}), the top 50 genes with the highest PCC (or SSIM) across the patients (\textit{Top 50 genes}), and the 264 genes that comprise the union of the 250 highly-expressed genes (HEG) and marker genes (\textit{all genes}).
Further details on model evaluation can be found in \textbf{Online Methods} section \textbf{Evaluation Metrics}.

We observe that \textit{3D + VOI} setting achieves the best PCC across all patients with an average of 0.46 for all, 0.57 for the top 50 predictive, and 0.42 for marker genes, and outperforming the 3D (\textit{$P\leq 0.001$} for all 0.29, top-50 0.41, marker 0.23) and the \textit{2D} settings (\textit{$P\leq 0.001$} for all 0.27, top-50 0.39, marker 0.21) (\textbf{Figure~\ref{fig:prostate}C}). Evaluation with SSIM provides the same conclusion with the \textit{3D+VOI} setting (all 0.56, top-50 0.58, marker 0.37) outperforming the \textit{3D} (all 0.51 \textit{$P\leq 0.001$}, top-50 0.52 \textit{$P\leq 0.001$}, marker 0.33 \textit{$P\leq 0.01$}) and the \textit{2D} settings (all 0.50 \textit{$P\leq 0.001$}, top-50 0.51 \textit{$P\leq 0.001$}, marker 0.32 \textit{$P\leq 0.01$}).
The trend is also maintained with the high-variable genes (HVG) or when the gene set size is expanded to 1,000 HEG, demonstrating the robustness of $\ours$ to gene sets (\textbf{Extended Data Figure~\ref{fig:ext_1000HEG_and_HVG}}). Next, we sought to assess whether $\ours$ can accurately capture the variance of expression levels across different spatial locations. First, we use Spearman's $\rho$ to compute the correlation between the variance of the measured expression levels and the variance of $\ours$-predicted expression levels across all 264 genes within each of 10 tissue sections (\textbf{Figure~\ref{fig:prostate}D}). 
Additionally, to assess whether $\ours$ can capture spatially heterogeneous expression patterns of the genes, we compute Moran's I\cite{moran1950notes} and Geary's C\cite{geary1954contiguity} for all genes to evaluate spatial autocorrelation. Specifically, we compute the difference of Moran's I and Geary's C between the ground truth and $\ours$-prediction, with smaller values indicating that $\ours$ better captures the expression heterogeneity (\textbf{Figure~\ref{fig:prostate}E}). We observe that \textit{3D + VOI} faithfully captures the expression variability, with Spearman's $\rho$ achieving the highest value and the two other metrics achieving the median value closest to 0. Examples of gene expression variance show that $\ours$ with \textit{3D+VOI} setting identifies and predicts gene expressions across a wide spectrum of variance (\textbf{Extended Data Figure~\ref{fig:ext_variance_ablation}}).

These results collectively indicate two important data scaling trends. 
First, incorporating depth context enhances the predictive performance, suggesting that the 3D context provides more morphological cues for predicting accurate transcriptomics expression. The second trend indicates that integrating VOI-specific morphomolecular information is crucial for ST prediction. Integrating measured ST data from a VOI apparently enables $\ours$ to learn VOI-specific morphomolecular links that are not represented in other volumes due to heterogeneity between cases (even of similar diseases or tissue types).

\hheading{Morphomolecular analysis of prostate cancer tissue volume}

\begin{figure*}[]
\centering  
\includegraphics[width=\textwidth]{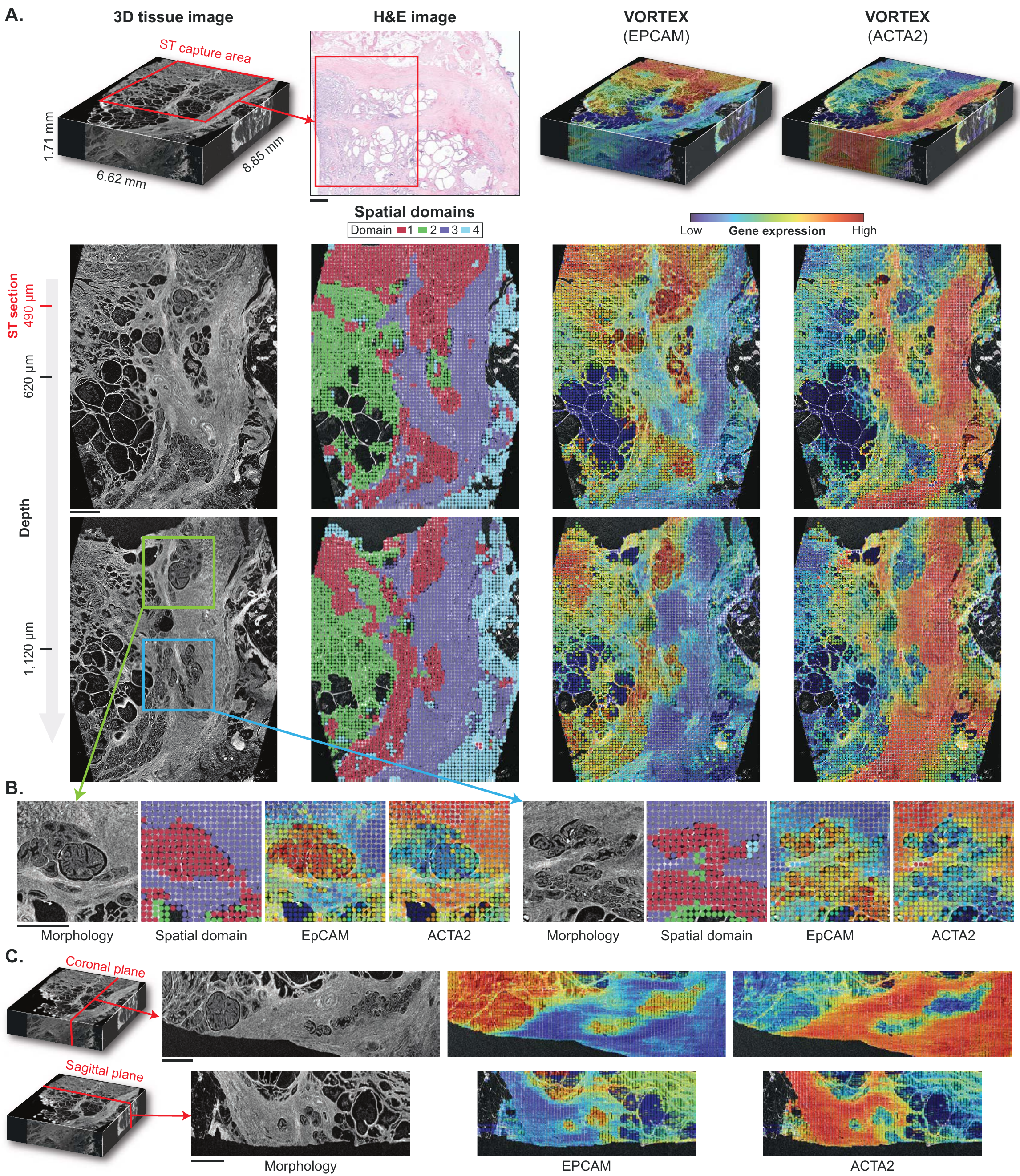}
\caption{\textbf{$\ours$ on large prostate cancer tissue}. \textbf{(a)} 3D ST prediction by $\ours$ on large prostate cancer tissue volume for \textit{EpCAM} and \textit{ACTA2} genes, with the spatial domains identified by $\ours$. Cross-sections at 620 $\mu m$ and 1,120 $\mu m$ are also displayed. The red box indicates the ST capture area at the depth of 490 $\mu m$. The corresponding H\&E tissue image based on which the ST capture area was selected is also shown.
\textbf{(b)} Zoomed-in regions-of-interests from the tissue section at depth 1,120 $\mu m$.
\textbf{(c)} The coronal and sagittal plane of the tissue volume and the corresponding prediction for \textit{EpCAM} and \textit{ACTA2}.
Additional examples can be found in \textbf{Extended Data Figure~\ref{fig:ext_prostate_3D_largeFOV}}. Scalebar is 1 mm.
}
\label{fig:ext_prostate_large}
\end{figure*}

To better understand the spatial distribution of predicted expressions and their links to the underlying tissue morphology, we construct 3D ST prediction heatmaps and 3D morphological segmentation masks. We construct two types of 3D ST prediction heatmaps, either based on the gene of interest or the gene set. We utilize spatial domains for the gene set, obtained by K-means clustering gene set prediction into distinct molecular groups. To render 3D ST prediction with higher spatial resolution than Visium sequencing spots, we apply the super-resolution framework TESLA\cite{hu2023deciphering}, which imputes ST for non-sequenced regions by aggregating expression levels of neighboring spots, on $\ours$-predicted expressions for each axial plane. 3D segmentation mask, generated based on the combination of pathologist's annotations on a few tissue sections and AI-based segmentation\cite{kirillov2023segment, ravi2024sam2segmentimages}, provides a morphological reference for comparison with molecular counterparts throughout the tissue volume. We create four to five morphological categories, dependent on the prostatic tissue subtypes present on each volume. For example, for P1 (\textbf{Figure~\ref{fig:prostate}F)}, we define tumor, benign glands, stroma, and adipose regions.
Utilizing both ST heatmaps and segmentation masks, we observe the consistency between the spatial distributions of the measured (ground truth) and predicted ST, both of which are localized to specific morphological categories (\textbf{Figure~\ref{fig:prostate}F, G, Extended Data Figure~\ref{fig:ext_prostate_3D}}). For example, the predicted \textit{SPON2} expression is upregulated in tumoral regions, \textit{MSMB} is upregulated in normal glands, and ACTA2 is upregulated in stromal regions, aligning with previous findings\cite{whitaker2010rs10993994, qian2012spondin, berglund2018spatial}. Further details on how 3D morphological segmentation masks are built can be found in \textbf{Online Methods} section \textbf{3D morphological segmentation}.

Visualization of ST heatmaps also enables analyzing inter-tumoral heterogeneity in 3D. A notable example is \textit{AZGP1}, part of OncotypeDx and the marker gene set, for which down-regulation in prostatic tumor glands is associated with poor prognosis and shorter biochemical recurrence (BCR)\cite{lapointe2004gene, burdelski2016reduced, kristensen2019predictive} (\textbf{Extended Data Figure~\ref{fig:ext_intertumoral}}). We observe that $\ours$ captures this inter-tumoral heterogeneity, and predicts different expression patterns based on the BCR status of each sample. P1 (low-risk for BCR) has high \textit{AZGP1} expression for both tumor and benign glands and P3 and P4 (high-risk for BCR) have low and high \textit{AZGP1} expression for tumor and benign glands, respectively. This reaffirms earlier observations that despite identifying cohort-wise consistent morphomolecular links, VOI fine-tuning can help further identify volume- or patient-specific as well as conflicting links (tumor glands with high and low \textit{AZGP1} expression for low and high BCR risk, respectively). 

Finally, we quantify the agreement between the predicted spatial domains and underlying morphology. Using 3D morphological segmentation masks as the ground truth annotation for each level, we compute the adjusted rand index (ARI) for the spatial domains along all levels of the axial dimension across three training scenarios (\textbf{Figure~\ref{fig:prostate}H}). Reflecting the previous trends, we observe that \textit{3D + VOI} results in the best ARI metric, with \textit{3D} and \textit{2D} achieving similar performance. Visual inspection of the spatial domains also confirms that the spatial domains discovered by \textit{3D + VOI} agrees the most with the segmentation masks (\textbf{Figure~\ref{fig:prostate}I}, \textbf{Extended Data Figure~\ref{fig:ext_spat_cluster}}). 

\hheading{3D ST prediction for large tissue volume}

The 3D ST prediction with $\ours$ can be easily scaled to a larger tissue volume captured with 3D imaging modalities, for which the planar field-of-view exceeds that of a typical ST capture area. As an example, $\ours$ fine-tuned on one tissue section with ST measurements at the depth of 490 $\mu m$ of the VOI can produce prediction for the large microCT tissue volume with larger planar area (tissue area $6.62\times 8.85\, mm^2$ vs. ST capture area for this sample $5.3 \times 6.5 \,mm^2$) and 171 times larger depth along the axial dimension (thickness for tissue $1.71 \,mm$ vs. one ST section $5 \,\mu m$) (\textbf{Figure~\ref{fig:ext_prostate_large}A}). Even in regions far apart from the tissue section with the ST measurement (plane at depth of 1,120 $\mu m$, with the distance of 630 $\mu m$ from the ST section at the depth of 490 $\mu m$), we observe that the expression of representative genes, \textit{EPCAM} and \textit{ACTA2}, show consistent overlap with tumor glands and stroma, respectively. This is also reflected in the alignment of the spatial domains with the different morphological classes, where domains 1, 2, 3, and 4 correspond to tumoral glands, benign glands, stroma, and adventitia (or adipose), respectively.
The spatial domains and gene expression patterns segment along fine-grained morphology, such as pockets of tumoral glands surrounded by stroma (\textbf{Figure~\ref{fig:ext_prostate_large}B}).
With access to continuous 3D morphology in the tissue volume, it is also easy to examine different views of the tissue, such as sagittal and coronal planes (\textbf{Figure~\ref{fig:ext_prostate_large}C}). Additional examples of $\ours$ scaling to larger tissue samples can be found in \textbf{Extended Data Figure~\ref{fig:ext_prostate_3D_largeFOV}}.
We emphasize that other 2.5D or 3D ST approaches lack such scalability. Tissue sections with ST measurements on both sides and close to the test tissue area are required for smooth interpolation, requiring a large number of ST measurements to profile thick tissue. Furthermore, prediction via extrapolation outside the ST capture area for each axial section is non-trivial.  

Finally, we assess the generalization capability of $\ours$ by applying the model to 3D ST prediction for a prostate cancer image captured with open-top lightsheet microscopy (OTLS), another non-destructive imaging modality that can provide H\&E-like appearance in 3D\cite{bishop2024end}. To this end, we apply $\ours$ pretrained on 2D H\&E morphology and ST pairs and predict the ST profile for each OTLS image plane, imaged at 1~$\mu m$/voxel, across the axial dimension (\textbf{Extended Data Figure~\ref{fig:ext_otls}}). We observe close agreement of the expression pattern with underlying morphology throughout the volume, such as \textit{KLK3} with tumoral glands and \textit{COL1AI} with stroma. This demonstrates $\ours$'s ability to handle different 3D imaging modalities flexibly and also underscores its generalizability even in the absence of ST measurements from a VOI.

\hheading{$\ours$ analysis for serial tissue sections}

Although 3D ST prediction based on 3D tissue images is the primary focus of $\ours$, the same principles can be applied to 2.5D tissue images consisting of serial tissue sections. While the sparse number of sections at certain intervals provides discontinuous and insufficient coverage of the tissue volume compared to 3D tissue images, easy integration into the current tissue processing workflow makes the serial section approach a practical alternative. To make $\ours$ compatible with 2.5D tissue images, we make two small adjustments. First, we replace the cross-modal registration with serial tissue section registration using VALIS\cite{gatenbee2023virtual} (\textbf{Extended Data Figure \ref{fig:breast_crc}A}). Next, with 3D morphological context for localized ST prediction infeasible due to non-contiguous sections, we construct a 2.5D context with equidistant neighboring sections instead, using the same depth aggregation module. Additional information about the 2.5D context can be found in \textbf{Online Methods} section \textbf{2.5D image data preprocessing}.

We validate $\ours$ on publicly available breast and colorectal cancer cohort volumes with serial tissue sections.
For the breast cancer cohort, we curate 101 H\&E-stained tissue sections with ST (58,263 spots) aggregated from several studies\cite{andersson2021spatial,he2020integrating,staahl2016visualization}. The curated dataset contains a mixture of serial sections cut at 32 $\mu m$ intervals for eight volumes and single sections from the remaining volumes.  $\ours$ evaluations are performed on four volumes for which six serial sections are available\cite{andersson2021spatial}. We use immediate neighboring sections placed at +32 $\mu m$ and -32 $\mu m$ as 2.5D context for each section. For the colorectal cancer cohort, we curate 26 H\&E-stained tissue sections with 2D ST (72,042 spots) from two studies\cite{valdeolivas2024profiling,mirzazadeh2023spatially}. Compared to the other cohorts for which 3D or serial 2D tissue images are available, the colorectal cancer cohort has only two sections with morphology and 2D ST from each tissue volume. Further details about the datasets can be found in \textbf{Extended Data Table \ref{tab:dataset}} and in the \textbf{Online Methods} section \textbf{Datasets}.

To understand how the data scaling trend \textit{across} tissue volumes influences the predictive performance, we investigate the performance of $\ours$ for the breast cancer cohort only with cohort pretraining (2.5D), only with training data pairs from VOI (VOI), and the combination of both (2.5D + VOI). We additionally evaluate how different amounts of VOI training pairs affect the performance, to understand the data scaling trend \textit{within} tissue volume (\textbf{Extended Data Figure~\ref{fig:breast_crc}B}). Specifically, for evaluating the top section of the volume with 2.5D context (S2), we gradually increase the number of ST sections from the bottom section with 2.5D context (S5) used for fine-tuning $\ours$, and vice versa. We evaluate the averaged performance for four patients with PCC and SSIM across three similar gene sets as before, adapted for this cohort with the marker gene set curated from HER2DX and anti-HER2 therapy\cite{prat2020multivariable, smith2021her2+}. 
First, we observe the data scaling trend within a tissue volume, where the predictive performance is increased as more tissue sections with paired morphology and 2D ST from the VOI are integrated. Including the maximum-allowable three sections with 2.5D context achieves the best performance across all gene sets, regardless of whether or not $\ours$ is pretrained with cohort data.
Furthermore, we observe the data scaling trend across tissue volumes, with \textit{2.5D + VOI} outperforming both \textit{2.5D} and $\textit{VOI}$ only (\textbf{Extended Data Figure~\ref{fig:breast_crc}C}). This trend is also maintained when the gene set is expanded to 1,000 HEG (\textbf{Extended Data Figure~\ref{fig:ext_1000HEG_and_HVG}C}).
For the same number of VOI fine-tuning tissue sections, the \textit{2.5D + VOI} setting always outperforms the \textit{VOI} setting, emphasizing the importance of cohort pretraining.
For the analyses on colorectal cancer, despite being restricted to \textit{2D} and \textit{2D+VOI} due to the lack of neighboring sections, we observe similarly that \textit{2D + VOI} performs better in both PCC and SSIM than the non-fine-tuning alternative of \textit{2D} setting (\textbf{Extended Data Figure~\ref{fig:breast_crc}D}). Additional analysis on the expanded gene set of 1,000 HEG also preserves the trend (\textbf{Extended Data Figure~\ref{fig:ext_1000HEG_and_HVG}D}).

Visualization of the predicted 2.5D ST heatmap shows that $\ours$ can reliably predict and localize morphological correlates of transcriptomic expressions for breast cancer, such as overexpression of \textit{ESR1} and \textit{COX6C} \cite{kim2017epithelial} in tumor (\textbf{Extended Data Figure~\ref{fig:breast_crc}E, \ref{fig:ext_CRC_3D}A}). The same holds true for the colorectal cancer cohort, where $\ours$ accurately captures the morphomolecular relationship such as up-regulation of \textit{EpCAM}, \textit{CEACAM5} and \textit{KRT8} in tumoral regions\cite{xiao2024integrating}(\textbf{Extended Data Figure~\ref{fig:breast_crc}F}, \textbf{Extended Data Figure~\ref{fig:ext_CRC_3D}B}).

Finally, we assess whether $\ours$ pretrained on a colorectal cancer cohort can generalize to an unseen large colorectal cancer tissue specimen. To this end, we use publicly available 22 serially-sectioned large H\&E tissue images at the intervals of roughly 25$\mu m$ from colorectal adenocarcinoma patient\cite{lin2023multiplexed} as an input volume to $\ours$, with the axial plane area of $12 \times 10\, mm^2$ easily exceeding typical ST capture area (\textbf{Extended Data Figure~\ref{fig:breast_crc}G}, \textbf{Extended Data Figure~\ref{fig:ext_CRC_3D}C}). While the lack of measured ST prevents quantitative evaluation of prediction quality, it provides a valuable validation for $\ours$ generalization capacity to unseen large volumes.
The five spatial domains identified by $\ours$ show close agreement with the five morphological clusters across the volume, achieving high and consistent ARI values across the depth. The localization of representative gene expression patterns to specific morphological regions supports this observation, such as \textit{EpCAM}, known to be up-regulated in tumoral regions\cite{spizzo2011epcam}, and \textit{ZG16}, known to be down-regulated in tumoral regions\cite{xu2020identification}. This localization captures subtle variations in fine-grained morphology, exemplified by a heterogeneous region with a thin `cord-like' structure consisting of stroma (left half) and normal mucosa (right half) surrounded by adenocarcinoma (\textbf{Extended Data Figure~\ref{fig:breast_crc}H}). $\ours$ successfully identifies three distinct spatial domains within this region, also with the predicted \textit{ZG16} and \textit{EpCAM} showing high expression for normal mucosa and adenocarcinoma, respectively. These results collectively underscore the capacity of $\ours$ to generalize to unseen volumes of large physical dimensions.

\hheading{Morphological biomarker exploration with $\ours$}

\begin{figure*}[!h]
\centering
\includegraphics[width=\textwidth]{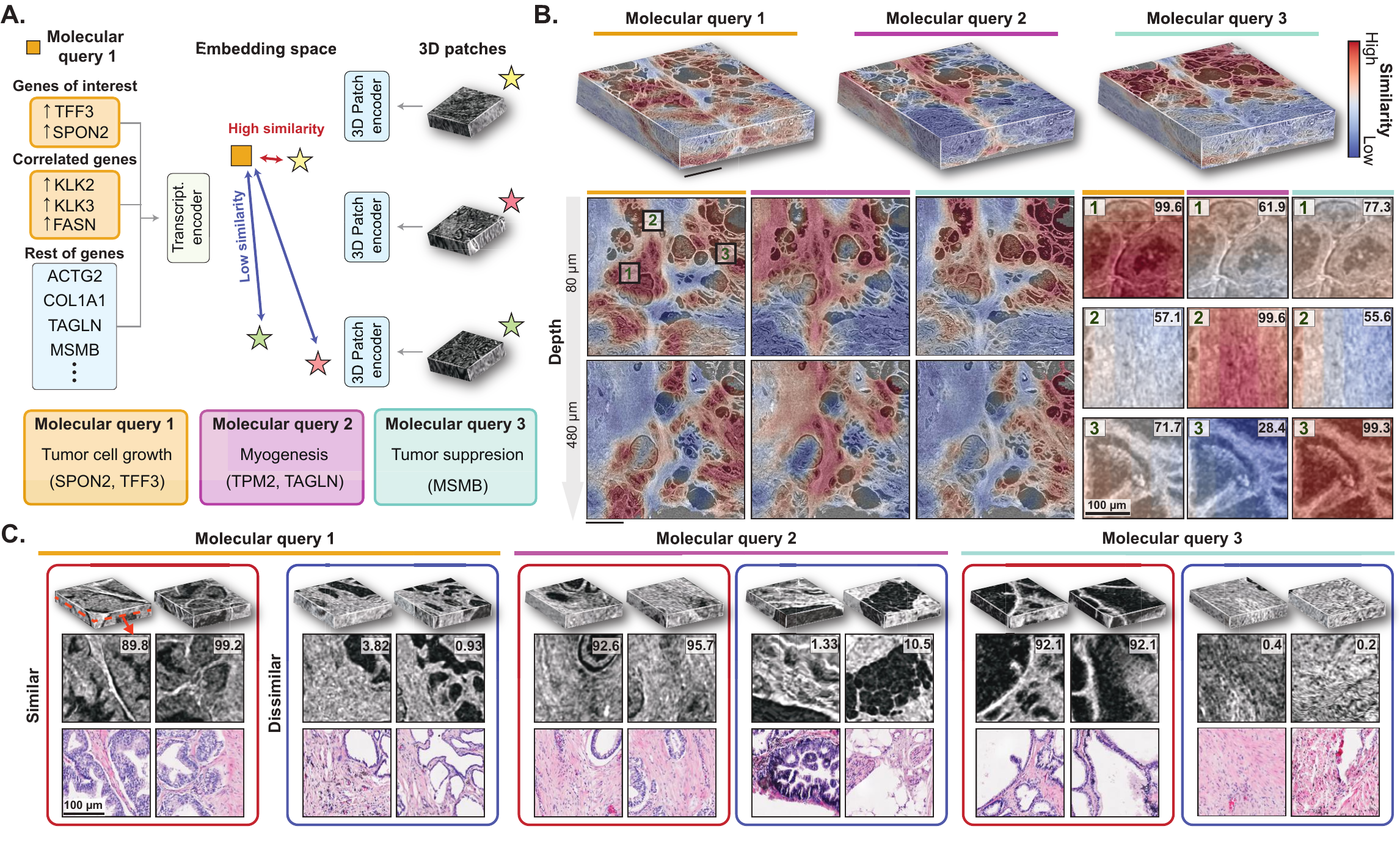}
\caption{\textbf{Cross-modal morphology retrieval with $\ours$}. Aligned morphology-transcriptomics embedding space, formed by 3D image encoder and transcriptomics encoder of $\ours$, enables cross-modal retrieval. Given a molecular query of transcriptomics profile based on biological functions of interest, the corresponding 3D morphology regions are retrieved, as determined by the closest Euclidean distance in the embedding space. 
\textbf{(a)} A schematic for 3D morphological retrieval with molecular queries. Genes are divided into three categories, well-known genes for a biological function (\textit{genes of interest}), the additional set of genes for which the expression levels across the ST spots are correlated with those genes (\textit{correlated genes}), and the remaining genes in the Visium sequencing panel shared across other samples in the cohort (\textit{rest of genes}). ST spot embeddings are filtered based on the expression levels of \textit{spatial filter genes} (combination of \textit{genes of interest} and \textit{correlated genes}) and averaged to form each molecular query embedding.
\textbf{(b)} 3D similarity heatmap, cross-sections, and representative patches from VOI. The number in the top-right corner of each patch indicates the similarity percentile rank of the patch for a given molecular query, with the higher percentile indicating the higher similarity. 
\textbf{(c)} Additional examples of similar and dissimilar 3D patches for each molecular query. For each 3D patch, the 2D microCT and H\&E patches from the middle section are also displayed. Unless specified otherwise, scalebar is 1mm.
}
\label{fig:recall}
\end{figure*} 

In addition to 3D ST prediction, $\ours$ can retrieve closely related 3D morphological regions for a transcriptomic profile query, referred to as cross-modal retrieval in a zero-shot setting.
Such a task can naturally be extended to biomarker exploration for identifying 3D morphological underpinnings of a specific transcriptomic profile, such as co-expression of genes of interest from specific functional gene sets.
This process involves using transcriptomic profiles as query inputs to obtain morphological outputs, representing the reverse of $\ours$ prediction workflow, where morphology is used as input to predict transcriptomic profiles.
Instead, this leverages the aligned image and transcriptomic embedding space from training $\ours$ with the contrastive loss, where the most similar 3D VOI patch embeddings are retrieved based on the Euclidean distance to the transcriptomic query embedding.

We design three molecular queries based on well-known biological processes. For well-known \textit{genes of interest} up-regulated in a biological function of interest, we identify a set of highly-correlated (co-expressed) genes across the ST spots. The union of the two gene sets (\textit{spatial filter genes}) is then used for filtering ST spots in the VOI, for curation of coherent expression profiles within the volume. Finally, we obtain the molecular query embedding by averaging the transcriptomic embeddings of the filtered spots.
This process ensures that spot-wise technical variations are removed and the rest of the genes in the query are at the baseline expression levels. 
The biological functions of the three queries correspond to \textit{Tumor cell growth} (query 1) with key genes \textit{TFF3}, and \textit{SPON2} associated with the \textit{PI3K/AKT/mTOR} pathway\cite{liu2018overexpression,zhang2024biological}, \textit{Myogenesis}\cite{liberzon2015molecular, ma2023prostate} (query 2) with the key genes \textit{TPM2}, and \textit{TAGLN}, and \textit{Tumor suppression} (query 3) with key genes \textit{MSMB}\cite{sjoblom2016microseminoprotein}, and \textit{ACPP}\cite{veeramani2005cellular}, respectively (\textbf{Figure~\ref{fig:recall}A}). More details on designing molecular queries and filtering process can be found in \textbf{Online Methods} Section \textbf{Molecular queries for cross-modal retrieval}.

We observe that the corresponding 3D similarity heatmaps and the representative 3D image patches agree with the morphological correlates of the gene subsets, as highlighted in red regions (\textbf{Figure~\ref{fig:recall}B}). The most similar morphology for the first query corresponds to the tumoral glands, which is expected due to the up-regulation of \textit{PI3K/AKT/mTOR} pathway involved in tumor growth\cite{berglund2018spatial, shorning2020pi3k}. The stromal regions are retrieved for the second query, with the \textit{Myogensis} pathway linked to stroma composition remodeling\cite{dakhova2014genes, ma2023prostate}. Finally, the benign glands are retrieved for the third query, with \textit{MSMB} and \textit{ACPP} involved in tumor suppression\cite{sjoblom2016microseminoprotein,veeramani2005cellular} and known to be up-regulated compared to the cancerous glands\cite{bjartell2007association, dahlman2011evaluation}.
To further quantify similarity, we rank the patches based on their distance to the molecular query in the embedding space, with a higher percentile indicating higher similarity. 
We observe that morphology of high similarity for one molecular query shows a lower similarity for other queries. For instance, the most similar stromal patch that ranks in the 99.6\textsuperscript{th} percentile for molecular query 2 is placed at the 57.1\textsuperscript{st} and 55.6\textsuperscript{th} for other queries (\textbf{Figure~\ref{fig:recall}B}). Additional examples of similar and dissimilar patches, for the tissue sections with corresponding H\&E images available, confirm the separation of the embedding space along different morphological concepts and the alignment with different transcriptomic profiles (\textbf{Figure~\ref{fig:recall}C}).
We additionally evaluate on the large CRC volume with serial sections, which reaffirms that $\ours$ can reliably retrieve morphological correlates of transcriptomic profiles (\textbf{Extended Data Figure~\ref{fig:ext_CRC_query}}).

These examples hint at $\ours$'s potential as a biomarker discovery tool for unknown 3D morphological correlates of custom transcriptomics profiles. Additionally, clinicians can triage sections or regions of interest of high or low similarity for further examination or molecular tests, based on the similarity heatmap provided by $\ours$.

\Heading{Discussion}

We introduce $\ours$, an AI-based 3D spatial transcriptomics (ST) prediction framework based on non-destructive 3D tissue images. 
With emerging 3D ST technologies, through \textit{in-situ} sequencing\cite{wang2018three, wang2021easi, fang2024three, sui2024scalable} or aligning serial sections with 2D ST measurements\cite{vickovic2022three, zeira2022alignment, zhou2023integrating,wang2023construction, lin2023multiplexed, schott2024open}, there are significant hurdles due to cost and limited coverage volume. $\ours$ provides a cost-effective and rapid alternative by leveraging the ability of deep learning models to learn fine-grained morphological correlates of gene expression, or \textit{morphomolecular links}.
Training $\ours$ on a collection of morphology and 2D ST data pairs from the same disease/tissue type results in a 3D ST prediction model that generalizes across unseen samples (3D CRC, \textbf{Extended Data Figure~\ref{fig:breast_crc}G}). This suggests that generic morphomolecular links, which are conserved across diverse patients, can be captured with sufficient training data. 
Additionally, fine-tuning the pretrained model on ST measurements of a few tissue sections from a volume of interest (VOI) further enhances the performance by integrating sample-specific morphomolecular links.
While the primary focus of $\ours$ is on non-destructive 3D tissue images, we demonstrate the model's flexibility in handling serial 2D tissue sections, a common approach for obtaining pseudo-3D ST.
Furthermore, leveraging the aligned morphological and transcriptomic embedding space from contrastively trained image and transcriptomics encoders, $\ours$ can retrieve corresponding morphological regions for custom molecular query, hinting at potential biomarker exploration tasks.

$\ours$ presents a scalable paradigm for 3D molecular analyses, building on the latest advances in deep learning. Current 3D ST computational approaches align and integrate ST measurements from multiple tissue sections within the same volume, without relying on the underlying tissue morphology\cite{zeira2022alignment, zhou2023integrating, wang2023construction,mo2024tumour, schott2024open, tang2024search}.
Consequently, each volume requires abundant ST measurements spread across multiple sections within each volume, and the model must be trained separately for each sample.
This prevents the reuse of shared disease- or organ-specific molecular traits across the samples, even within the same disease or organ cohort, confining the 3D molecular analysis to a relatively small number of samples due to financial and time constraints. 
In this context, $\ours$ enables the integration and reuse of morphomolecular traits both \textit{across} and \textit{within} the samples, leveraging 3D tissue morphology as the common binding factor. The model can then match and integrate morphological characteristics across different imaging modalities and dimensions (e.g., 3D imaging modality and 2D H\&E) and learn the correspondence between such morphological characteristics and transcriptomics expression commonly shared or specific to each sample. This allows $\ours$ to easily scale and generalize to other samples of similar disease profiles of any physical size.
Finally, the predictive performance is expected to further increase with continued developments in computational pathology\cite{jaume2024hest, chen2024towards, zimmermann2024virchow, campanella2024clinical} and single-cell foundation models\cite{cui2024scgpt, hao2024large}, on top of increasing availability for ST-morphology data pairs.

A limitation of the study is the relatively coarse resolution of microCT and Visium, which is insufficient for achieving single-cell resolution. With advances in non-destructive tissue 3D imaging and the increasing availability of single-cell sequencing data paired with high-resolution morphology data, we expect the resolution constraint to be temporary. On the imaging front, the next-generation microCT\cite{frohn20203d, walsh2021imaging,palermo2025investigating} and other 3D imaging modalities such as open-top light-sheet microscopy\cite{glaser2017light, bishop2024end} and holotomography\cite{kim2024holotomography}, have shown promise in capturing morphological details at the cellular level. Furthermore, increasing the availability of paired morphology and single-cell sequencing data through concerted efforts\cite{jaume2024hest, chen2024stimagekm} will provide sufficient data for $\ours$ training. Combined with emerging works showing that single-cell gene expression can be reliably predicted from histomorphology\cite{comiter2023inference, chadoutaud2024scellst, zhang2024inferring}, we expect $\ours$ to easily extend to 3D ST prediction at the single-cell level.
In conclusion, $\ours$ paves the way for a scalable approach to 3D ST prediction. 
We envision $\ours$ to assist clinicians and biomedical researchers in the 3D multimodal analysis of tissue at a large scale towards novel biomarker exploration.
\clearpage
\Heading{Online Methods}

\hheading{Datasets}

\noindent\textbf{Prostate Cancer} Archived formalin-fixed and paraffin-embedded (FFPE) prostatectomy specimens were collected from $n=11$ patients and imaged with micro-computed tomography (microCT). Following imaging, either one or two histological sections per sample were prepared using a microtome depending on the sample thickness, stained with H\&E, and subsequently subjected to Visium spatial transcriptomics profiling. Overall, we obtained 65,715 spots. 
In addition to the internal cohort, we compiled data from multiple public sources.  The dataset includes three sources: (1) a study where Visium profiling was applied to 22 frozen H\&E-stained histological sections from 2 patients with prostate adenocarcinoma, yielding 58,906 spots\cite{erickson2022spatially}, (2) a study that used Spatial Transcriptomics technology on 24 needle biopsies snap frozen from three patients with prostate adenocarcinoma, resulting in 3,969 spots\cite{marklund2022spatio}, and (3) the public 10x Genomics repository, which provides three FFPE prostate samples from both healthy and diseased individuals, totaling 9,957 spots. Collectively, the public cohort we compile consisted of 72,832 spots from 49 histological sections. Overall, from public and in-house cohorts, we obtain 138,547 spots from 68 sections and 20 patients. Further details can be found in \textbf{Extended Data Table \ref{tab:dataset}}.

\noindent\textbf{Breast Cancer} We compiled 58,263 H\&E image patches with associated transcriptomic profiles from 101 histological sections and 35 patients from three previous works\cite{andersson2021spatial, staahl2016visualization, he2020integrating} as well as the 10x Genomics public repository. The samples are a mixture of frozen and FFPE histological sections with associated transcriptomics profiles from a 
sequencing-based ST technology (Visium or Spatial Transcriptomics). 
2.5D tissue images consisting of serial tissue sections were obtained from a previous study \cite{andersson2021spatial}. The cohort includes samples from 8 patients (A-H): four patients (A-D) with 6 serial sections each, and four patients (E-H) with 3 serial sections each. The histological sections were acquired at the thickness of 16~\textmu m, with a spacing of 32~\textmu m between each section.  Overall, 6,577 paired H\&E image patches with associated 2.5D morphology and ST spots from 16 histological sections were availble. Further details can be found in \textbf{Extended Data Table \ref{tab:dataset}}.

\noindent\textbf{Colorectal Cancer} We curated a dataset of paired ST-H\&E samples from colon, rectum, and cecum areas of human donors that had been subjected to a sequencing-based ST technology (Visium or Spatial Transcriptomics). The dataset is a compilation of three sources of publicly available data:  two studies on colorectal cancer \cite{valdeolivas2024profiling,mirzazadeh2023spatially}, and the publicly available samples on the 10x Genomics website. Overall, the dataset consists of 72,042 ST spots with paired H\&E image patches from 26 histological sections and 13 patients. 
For validation, we used the public cohort \cite{valdeolivas2024profiling} which includes 
samples from 7 patients, each with 2 histological sections, totaling 20,708 spatial transcriptomic spots (Dataset CRC I in \textbf{Extended Data Table \ref{tab:dataset}}). 

We additionally apply the model to publicly available large CRC tissue volume sample~\cite{lin2023multiplexed}. The tissue volume has dimensions of 1.6~cm~x~1.6~cm~x~0.5~cm and consists of 22 serial H\&E-stained histological sections obtained from a specimen with poorly differentiated stage IIIB adenocarcinoma and adjacent normal tissue resected from the cecum of a 69-year old male.

\hheading{Data acquisition}

\noindent\textbf{MicroCT} MicroCT imaging was performed on the collection of FFPE prostate cancer tissue blocks using the Versa 620 X-ray Microscope (Carl Zeiss, Inc., Pleasanton, California, USA) at a resolution of 4 $\mu m$/voxel. For each scan, a microfocus X-ray source operating at a tube voltage of 40~kV and a filament current of 75~mA (3~watts) was utilized. A total of 4,501 projection images were acquired, with the sample rotated by 0.08 degrees per projection (360 degrees/4,501). The images were captured using a 16-bit flat panel detector with a resolution of 3,064 $\times$ 1,928 pixels, resulting in a stack of 1,300 2D images along the depth dimension. Each projection was averaged from 15 frames, with an exposure time of 0.5 seconds per frame (totaling 7.5 seconds per projection) to enhance the signal-to-noise ratio. The detector recorded raw grayscale intensity values for each voxel. Scanning each sample required 11 hours, to cover a field-of-view measuring 12.8~mm~$\times$~ 7.68~mm$\times$~ 5.2~mm (3,200~$\times$~1,920~$\times$~1,300 voxels). 
To ensure a consistent image intensity scale across samples, we normalize the 3D tissue image for each sample with the lower threshold of 25,000 intensity value and the upper threshold to the top 1\% of each tissue’s intensity value.

\noindent\textbf{Visium spatial transcriptomics on Prostate samples}
Following the imaging with microCT, two sections spaced 250 $\mu$m apart were obtained from samples P1~$\sim$~P5, and one section was cut in samples P6 $\sim$ P11. Each section was subject to Visium ST, where the RNA expression measurements were performed in spots with a diameter of 55~$\mu$m arranged in a grid with a centre-to-centre distance of 100~$\mu$m. The number of spots in each section ranged from 3,402 to 4,760. DV200 was first performed on the tissue sections for quality control, followed by the Tissue Adhesion Test outlined in the 10x Genomics protocol. The sections (5 $\mu$m in thickness) were placed on a Visium Spatial Gene Expression Slide following the Visium Spatial Protocols–Tissue Preparation guide. Sections were left drying  and deparaffinized following the  protocol for Visium Spatial Gene Expression for FFPE – Deparaffinization, H\&E Staining, Imaging \& Decrosslinking
(10x Genomics, CG000409 Rev D). Tissue sections were stained with H\&E and imaged at 0.62 $\mu$m/pixel resolution. Decrosslinking was immediately carried out for H\&E-stained sections. Subsequently, human whole transcriptome probe panels were applied to the tissue. After probe hybridization and target gene ligation, the ligation products were released through RNase treatment and tissue permeabilization. Ligation probes were finally hybridized to the spatially barcoded oligonucleotides on the capture area.  Spatial transcriptomics libraries were constructed from the probes and sequenced using an Illumina NovaSeq 6000 system 300 cycle with an S4 flow cell. High-resolution tissue images were captured with Olympus BX51 scope and DP74 camera.

\hheading{3D image data preprocessing}

\noindent\textbf{Cross-modal registration} To align genomic expression data with volumetric microCT images, a registration pipeline was deployed to map H\&E images (and corresponding spot coordinates) onto microCT volumes. Given the intensity and color differences between microCT and H\&E-stained histologies, a feature-based approach was chosen over intensity-based registration. The registration paradigm consists of two main steps: first, identifying the (virtual) plane within the microCT volume that corresponds to the H\&E-stained image thereby determining the exact tilt angle during the sectioning process; and second, achieving precise 2D alignment between the H\&E-stained image and its corresponding microCT plane. In more detail, we first downscaled the H\&E images to match the resolution of the microCT images (4 $\mu$m/pixel). Then,
we followed a previous work\cite{chicherova2014histology} to extract Speeded Up Robust Features (SURF) descriptors \cite{bay2006surf} for each axial microCT plane and matched those to the histology image using a second-nearest-neighbor-criteria. Following this, RANSAC plane fitting \cite{fischler1981random, dong2022deciphering} was used to fit a plane to the 3D point cloud of matching points. The resulting plane matching the H\&E-stained histological section was virtually cut from the microCT volume. For the second step, non-elastic registration\cite{gatenbee2023virtual} was applied to the microCT plane to achieve pixel-wise alignment between corresponding images from both modalities. After microCT – histology registration, the coordinates of the spots containing the gene expression levels were readily aligned to the volumetric microCT scans.
 
\noindent\textbf{Image patches} We downsample all H\&E whole-slide-images (WSIs) to 1 $\mu m$/pixel to ensure the image resolutions are consistent. Next, we crop $112\times112$ pixel image patch centered around each ST spot to obtain data pairs of 2D patch and 2D ST. 
Following image co-registration of prostate microCT images to their corresponding H\&E images with associated ground truth ST expression, we crop $112\times 112$ pixel patch ($4 \mu m$/pixel) at the same axial locations as in the H\&E images. In the \textit{2D} training setting, we only consider the 2D microCT plane image for which the ST measurement data is available, resulting in 2D patches of $112\times 112\times 1$ voxels ($448~\mu m \times 448~\mu m \times 4~\mu m $). In the \textit{3D} and \textit{3D+VOI} settings, 20 adjacent planes (10 planes above and below the central plane with 2D ST data) are incorporated to capture tissue context, forming 3D patches with $112\times 112\times 21$ voxels ($448~\mu m \times 448~\mu m \times 84~\mu m$). The intensity in each patch is then normalized to [0,1].  

\hheading{2.5D image data preprocessing}

\subsection{Image registration - Serial H\&E sections}
Serial histological sections from the breast and colorectal cancer cohorts were co-registered to generate 2.5D digital tissue samples. The images were 
first downsampled to 1~$\mu$m/pixel and then aligned sequentially, with the middle section of the image stack serving as the reference image. Alignment was performed using a two-step process: a landmark-based rigid registration followed by a non-rigid registration for enhanced precision.  For the breast cohort, where each histological section has corresponding 2D ST data, ST spots of consecutive sections were aligned by applying the same registration transformation that had been used for image alignment, using the VALIS registration framework \cite{gatenbee2023virtual}.

\subsection{Image patches}
Upon serial registration of tissue sections, in the breast cancer cohort, we construct 2.5D patch of $112\times 112\times 3$ pixel from three consecutive neighboring sections that are $32\mu m$ apart. Specifically, the tissue section for 2D ST prediction was assigned the central section, and the sections immediately above and below it were considered. Consequently, for samples with six sections, we only considered ST prediction for four middle sections for which 2.5D context was available (S2$\sim$S5). For samples with three sections, we only considered the single middle section. In the colorectal cancer samples used for model training and evaluation, we crop
2D image patches of $112\times 112$ pixels at 1 $\mu m$/pixel resolution. For the colorectal cancer sample consisting of 22 serial histological sections with no ST data available, we crop non-overlapping $112~\mu m\times112~\mu m$ image patches on each plane.

\hheading{Transcriptomic data preprocessing}

\noindent\textbf{ST spot filtering and expression normalization} Spatial transcriptomics spots are first filtered based on the number and type of expressed genes. Spots containing gene expression of at least 25 genes and with less than 20\% of mitochondrial genes were considered. For ST prediction, we preprocess the gene counts with a series of two transformations. First, we normalize the total gene expression of each spot to a library size of 10,000 to equalize the sequencing depth across different samples and spots. This is then followed by a log transformation. The normalized gene expression for each spot is smoothed by averaging its expression values with that of its immediate spot neighbors ($\sim$10 closest neighbors) for removing spot-specific measurement noise\cite{he2020integrating, chung2024accurate, zhang2024inferring}.

\noindent\textbf{Gene expression panel for evaluation} 
Different ST technologies provide the expression levels of different sets of genes. For example, in the prostate cancer cohort, comprised of data from four sources, the number of sequenced genes per spot ranges from 17,943 to 33,538 (\textbf{Extended Data Table \ref{tab:dataset}}). To integrate data from multiple sources, we first identified the intersection of gene expression panels for each cancer cohort, considering only genes common to all spots. This process resulted in 8,136 genes for the prostate cohort, 8,034 for the breast cohort, and 10,765 for the colorectal cancer cohort. For the main experiments, we first curated the 250 genes with the highest mean expression (HEG) for a given tissue cohort, in line with the previous studies\cite{he2020integrating, chung2024accurate}. We subsequently incorporated a set of genes with prognostic value for each tissue cohort into the 250 gene set. For the prostate cancer cohort, we utilized the genes from Oncotype DX and Decipher, which are molecular assays used for evaluating prostate cancer risk, resulting in \textit{All genes} panel with 264 genes. For the breast cancer cohort, we incorporated the genes from the prognostic score HER2DX\cite{prat2020multivariable}, as well as a set of genes involved in evading anti-HER2 therapy\cite{smith2021her2+}, yielding an \textit{All genes} panel with 269 genes. For the colorectal cancer cohort, we considered a set of mutated driver genes and genes significant in several key pathways from two studies\cite{nunes2024prognostic,valdeolivas2024profiling}, resulting in \textit{All genes} panel with 276 genes. To assess the robustness of $\ours$ on different gene expression panels, we also analyze a gene expression panel with 1,000 HEGs and also with 250 highly-variable genes (HVG). Marker gene names for each cancer cohort are included in \textbf{Extended Data Table \ref{tab:marker_genes}}.

\noindent\textbf{Gene expression input processing} To prepare the transcriptomics for being encoded with the scGPT transcriptomics encoder\cite{cui2024scgpt}, we applied the default preprocessing transformations required for fine-tuning this gene encoder. Specifically, the 1,200 most HVGs were selected from each cancer cohort and filtered for each spot. This was followed by a log1p transformation and a value binning technique to convert expression counts into relative values.

\hheading{$\ours$ architecture}

$\ours$ combines the two directions of ST evaluation, the direct regression-based approaches\cite{he2020integrating} and cross-modal alignment approaches\cite{xie2024spatially, min2024multimodal}, resulting in the model with ST prediction and image-ST alignment branch, similar to CoCa\cite{yu2022coca} in vision-language literature. Aligning ST to the corresponding image modalities, 2D H\&E image patches and 3D microCT patches, allows $\ours$ to perform cross-modal retrieval tasks in addition to ST prediction, making $\ours$ a flexible framework for diverse tasks. 
$\ours$ is comprised of four main components: \textit{2D image encoder}, \textit{transcriptomics encoder}, \textit{3D image encoder}, and \textit{transcriptomics predictor}. 

\noindent\textbf{2D image encoder} We choose CONCH\cite{lu2024visual} as the \textit{2D image encoder} for two reasons. First, it was pretrained on histology regions with diverse types and stains, including frozen tissue, FFPE, and immunohistochemistry, yielding image features robust to different tissue processing protocols across data sources. Next, it was shown to be one of the most competitive models for predicting transcriptomic profiles for diverse cancer types on public HEST-1K benchmark\cite{jaume2024hest} and showing robust performance across different tissue stains and textures\cite{filiot2025distilling}. 
Instead of directly using 512-dimensional image patch embedding from CONCH, we use embeddings from the pre-contrastive module, which is a set of 196 ($= 14 \times 14$) patch token embeddings, each of which is dimension $\mathbb{R}^{768}$. This provides additional flexibility in using image encoder output embeddings for different downstream tasks, as further mentioned in \textbf{Attentional Poolers}.
To aggregate training pairs from diverse sources, $\ours$ needs to handle batch effects arising from integration of diverse data sources. Besides using image and transcriptomics encoders pretrained on diverse data sources, we include a lightweight MLP to encode the source/batch ID to distill biological variations while removing batch-associated variations during training through a domain adaptation loss, following previous work\cite{ganin2015unsupervised, vaidya2024demographic, cui2024scgpt}. Upon training, the MLP module is discarded for downstream tasks. 

\noindent\textbf{3D image encoder} To encode a 3D patch, we use a transformer-based architecture (ViT-B/16) pretrained on natural images (ImageNet-1K) as the backbone of the 3D image encoder. Instead of treating 3D patch as a volume, we treat it as a stack of 2D patches. Specifically, the image encoder first extracts a set of 196 2D patch token features for every 2D section of the 3D patch. A depth-specific learnable embedding is then added to each set of token features. Following the works in video processing\cite{alayrac2022flamingo}, the same learnable embedding is added to all the token features in the same depth, without additional 2D positional embeddings. Subsequently, the sets are merged to form a larger set of patch token embeddings\cite{alayrac2022flamingo}. For example, a 3D patch with depth 21 would result in 4,116 ($=196\times 21$) patch token features. The choice of ImageNet-pretrained VIT as the image backbone is motivated by the prior study\cite{song2024analysis}, which demonstrated that image encoders pretrained on natural images provided better transfer performance for the microCT data compared to other radiology-specific image encoders, due to inherent texture and resolution differences between MRI/CT and microCT. Notably, $\ours$ is flexible in its components, allowing easy replacement with more powerful modality-specific 3D imaging foundation models as they become available. For encoding 2.5D patches from the serial tissue sections dataset, we instead use the CONCH image encoder, as both 2D and 2.5D modalities are the same. 

\noindent\textbf{Attentional Poolers} The output of the image encoders is a set of token features for each image patch. For the 2D image encoder, this amounts to 196 ($=14\times 14$) token features per 2D image patch. For the 3D image encoder, this amounts to 4,116 ($196\times 21$ sections) token features per 3D image patch. A single-layer Transformer, termed \textit{attentional pooler}, facilitates the encoding of interactions between the token features set and a set of learnable embeddings (queries), each of which is dimension $\mathbb{R}^{768}$. Next, each query is projected to a lower dimension of $\mathbb{R}^{512}$ through a linear layer. The encoded queries are then used for subsequent downstream tasks.
In $\ours$, we introduce two attentional poolers, one for the ST reconstruction task and the other for the contrastive task, inspired by CoCa framework\cite{yu2022coca}. For a given $i^{\text{th}}$ image patch, the \textit{contrastive attentional pooler} for cross-modal alignment with constrastive learning uses a single query ($n_{\text{contrast}}=1$) to encapsulate the global representation of the patch, resulting in $\mathbf{h}_i^{\text{2D, cont.}}\in\mathbb{R}^{512}$ and $\mathbf{h}_i^{\text{3D, cont.}}\in\mathbb{R}^{512}$, for 2D and 3D patch, respectively. The \textit{reconstruction attentional pooler} uses $n_{\text{recon}}=32$ queries to capture more localized and fine-grained image details for ST prediction, resulting in $\{\mathbf{h}_{i,j}^{\text{2D, rec.}}\}_{j=1}^{32}$ and $\{\mathbf{h}_{i,j}^{\text{3D, rec.}}\}_{j=1}^{32}$ with $\mathbf{h}_{i,j}^{\text{2D, rec.}}, \mathbf{h}_{i,j}^{\text{3D, rec.}}\in\mathbb{R}^{512}$, for 2D and 3D patch,respectively.

\noindent\textbf{Transcriptomics encoder}
We encode ST data using a modified version of scGPT \cite{cui2024scgpt}, a single-cell foundation model pretrained on transcriptomics data from millions of cells of various cancer types. While initially developed for single-cell data, we adapt scGPT to encode transcriptomics data from Visium and Spatial Transcriptomics spots, which typically contain about 10 and 20 cells, respectively\cite{elosua2021spotlight}. This follows the successful adaptations of single-cell foundation models to encode transcriptomics data beyond single-cell, such as tissue bulk RNA expression and spatial transcriptomics data through model fine-tuning\cite{vaidya2025molecular,lee2024pathomclip}. ScGPT features three key components: a \textit{gene-name} encoder, an \textit{expression-value} encoder, and a \textit{transformer} encoder. The \textit{gene-name} encoder comprises an embedding layer that maps each gene to a fixed-length embedding vector of dimension 512. The \textit{expression-value} encoder consists of two fully connected layers with ReLU activation, which transform each gene expression value into a 512-dimensional vector. The output of the \textit{gene-name} encoder and the \textit{expression-value} encoder are then combined through element-wise addition, forming the input to the \textit{transformer} encoder, which is a stack of 12 Transformer layers, each with eight attention heads. The \texttt{<CLS>} token from the last transformer layer in fed into a single fully-connected layer for the transcriptomics embedding $\mathbf{g}_i\in\mathbb{R}^{512}$ for $i^{\text{th}}$ sequencing spot.
scGPT encoders are initialized from the \texttt{pancancer} checkpoint (pretrained on 5.7 million cells of various cancer types) and the projection head is randomly initialized.

\noindent\textbf{Transcriptomics predictor} We use a Transformer with a single layer followed by a single fully-connected layer as the transcriptomics predictor $f_{\text{pred.}}$, which takes $\{\mathbf{h}_{i,j}^{\text{3D, rec.}}\}_{j=1}^{32}$ as the input to predict the ST expression levels, i.e., $\widehat{\mathbf{y}}_i = f_{\text{pred.}}(\{\mathbf{h}_{i,j}^{\text{3D, rec.}}\}_{j=1}^{32})$. Consequently, $\widehat{\mathbf{y}}_{i,j}$ corresponds to the $j^{\text{th}}$ gene expression prediction for $i^{\text{th}}$ spot. The predictor can also operate on 2D patch embeddings $\{\mathbf{h}_{i,j}^{\text{2D, rec.}}\}_{j=1}^{32}$, for earlier pretraining stages.

\hheading{$\ours$ training}

$\ours$  is trained over three stages designed to gradually build the capacity of 3D ST prediction for the volume-of-interest (VOI). The first two stages utilize 2D and 3D images of all the volumes except VOI in the same cancer cohort. If the 2D ST measurements from VOI are available, the third stage is performed to fine-tune the model. All three stages use adaptations of loss functions used in CoCa\cite{yu2022coca} designed to predict transcriptomics profiles from image embeddings while also aligning them with transcriptomics embeddings.

\noindent\textbf{Stage I: 2D Pretraining on cancer-specific heterogeneous samples} During the pretraining stage, we leverage all available paired 2D H\&E-stained histology and ST data for a given cancer cohort. At this stage, $\ours$ takes three types of data as inputs: 2D morphology from 112$\times$112~\textmu m histology image patches (112$\times$112 pixels) centered at the location of each of the ST spots, transcriptomics expression data after preprocessing, and the source ID for correcting batch effects. The 2D morphology and transcriptomics data are encoded with the 2D image encoder and transcriptomics encoder, respectively. The contrastive and reconstruction attentional poolers are randomly initialized and trained. The last three transformer layers from the 2D image encoder and transcriptomic encoder are also fine-tuned to provide task-specific embeddings. We apply a conventional data augmentation scheme to image patches such as horizontal flip, vertical flip, and color jittering. 

We use a combination of three loss functions: symmetric cross-modal contrastive learning objective ($\mathcal{L}_{\text{cont.}, \mathrm{I}}$), ST reconstruction loss ($\mathcal{L}_{\text{rec.},\mathrm{I}}$), and domain adaptation loss ($\mathcal{L}_{\text{da}}$).

\textbf{Contrastive loss ($\mathcal{L}_{\text{cont.}, \mathrm{I}}$)} We align the embedding space of the 2D image encoder and transcriptomic encoder using a symmetric cross-modal contrastive learning objective.  Specifically, for a batch of $M$ pairs $\left\{ \left( \mathbf{h}_{i}^{\text{cont.}}, \mathbf{g}_i \right) \right\}_{i=1}^{M}$ with $\mathbf{g}_{i}\in\mathbb{R}^{512}$ and $\mathbf{h}^{\text{cont.}}_{i}\in\mathbb{R}^{512}$ denoting the $i^{\text{th}}$ transcriptomic and histology (single query from the contrastive attentional pooler) normalized embeddings respectively, the loss function is defined as:
\begin{equation}
\mathcal{L}_{\text{cont.}, \mathrm{I}} = 
- \frac{1}{2M} \sum_{i=1}^{M} \log \frac{\exp\left(\tau (\mathbf{h}_{i}^{\text{2D, cont.}})^{\top} \mathbf{g}_i \right)}{\sum_{j=1}^{M} \exp\left(\tau (\mathbf{h}_{i}^{\text{2D, cont.}})^{\top} \mathbf{g}_j \right)}
- \frac{1}{2M} \sum_{j=1}^{M} \log \frac{\exp\left(\tau \mathbf{g}_j^{T} \mathbf{h}_{j}^{\text{2D, cont.}} \right)}{\sum_{i=1}^{M} \exp\left(\tau \mathbf{g}_j^{T} \mathbf{h}_{i}^{\text{2D, cont.}} \right)},
\end{equation}
where $\tau$ is the temperature parameter. The first term represents histology-to-gene loss, and the second represents gene-to-histology loss. The loss function $\mathcal{L}_{\text{cont.,I}}$ aims to minimize the distance between paired embeddings while maximizing the distance between unpaired embeddings.

\textbf{Reconstruction loss ($\mathcal{L}_{\text{rec.},\mathrm{I}}$)} In addition to the contrastive loss, we use the reconstruction loss to minimize the error between the predicted gene expression and the ground truth ST profiles. Specifically, we minimize the mean squared error (MSE) between the (smoothed) ground truth gene expression ($\mathbf{y}_i$) and the predicted expression obtained from the histology image embeddings $\{\mathbf{h}_{i,j}^{\text{2D, rec.}}\}_{j=1}^{32}$,
\begin{equation}
\mathcal{L}_{\text{rec., I}} = \frac{1}{M} \sum_{i=1}^{M} \left\| \mathbf{y}_i - f_{\text{pred.}}\left(\{\mathbf{h}_{i,j}^{\text{2D, rec.}}\}_{j=1}^{32}\right) \right\|_2^2.
\end{equation}

\textbf{Domain adaptation loss ($\mathcal{L}_{\text{da}}$)} To address potential batch effects by integrating ST samples from multiple data sources, we train the MLP classifier to infer the batch source ID from the transcriptomic embedding ($\mathbf{g_i}$) and use a cross-entropy loss. Batch source IDs are defined based on the data sources included in \textbf{Extended Data Table \ref{tab:dataset}}. As the aim is to make the model invariant to the batch attribute, the negative of the attribute prediction loss is back-propagated, making the model poor in predicting the data source. 

The total loss minimized during this stage is defined as:
\begin{equation}
    \mathcal{L}_{\mathrm{I}} = \lambda_{\text{cont.},\mathrm{I}}\cdot\mathcal{L}_{\text{cont.},\mathrm{I}} + 
\lambda_{\text{rec.},\mathrm{I}}\cdot\mathcal{L}_{\text{rec.},\mathrm{I}} +
\lambda_{\text{da}}\cdot\mathcal{L}_{\text{da}}, 
\end{equation}
where we use $\lambda_{\text{cont.},\mathrm{I}}=\lambda_{\text{rec.},\mathrm{I}}=1$ and $\lambda_{\text{da}}=0.1$.
The model is trained with a batch size of 512 for 25 epochs. The initial five epochs are used for warmup, where the learning rate is linearly increased from 0 to $1 \times 10^{-5}$. Next, the cosine scheduler is applied with the learning rate decaying from $1 \times 10^{-5}$ down to 0 by the end of training. The weight decay is set to 0.01 and the AdamW optimizer is used with $\beta$ values of (0.9, 0.999). Further details on hyperparameters and training settings are provided in \textbf{Extended Data Table \ref{tab:stageI}}.

\noindent\textbf{Stage II: 3D pretraining}
Upon establishing the morphomolecular link between 2D H\&E histology and transcriptomics, the second stage focuses on further fine-tuning $\ours$ to capture the relationship between the morphology present in 3D tissue imaging data, and transcriptomics.
Specifically, we encode the morphology of 3D tissue image data with the 3D image encoder and align the embedding to the corresponding 2D H\&E and ST embeddings. In doing so, we also fine-tune the transcriptomics predictor such that the model can transition from predicting ST from 2D H\&E patch to predicting ST from 3D microCT patch. To preserve the morphology-transcriptomics embedding space from the previous stage, the 2D image encoder and transcriptomics encoder are kept frozen. To account for the smaller size of microCT and ST data pairs compared to the pretraining dataset, we also keep the 3D image encoder frozen to prevent overfitting, instead opting to train the randomly initialized contrastive and reconstruction attentional poolers.

We use a combination of three loss functions: a symmetric cross-modal contrastive learning objective ($\mathcal{L}_{\text{cont., II}}$), a direct alignment loss ($\mathcal{L}_{\text{dir}}$), and a reconstruction loss ($\mathcal{L}_{\text{rec., II}}$).

\textbf{Contrastive loss ($\mathcal{L}_{\text{cont.}, \mathrm{II}}$)} Similar to the alignment between histology and transcriptomic embeddings in the first stage, we align the embedding space of the 3D image encoder to that formed between the 2D image encoder and transcriptomic encoder, using a dual symmetric cross-modal contrastive learning objective
\begin{equation}
\begin{aligned}
\mathcal{L}_{\text{cont., II}} = 
& - \frac{1}{2M} \sum_{i=1}^{M} \log \frac{\exp\left(\tau (\mathbf{h}_{i}^{\text{3D, con.}})^{\top} \mathbf{g}_i \right)}{\sum_{j=1}^{M} \exp\left(\tau (\mathbf{h}_{i}^{\text{3D, con.}})^{\top} \mathbf{g}_j \right)} 
- \frac{1}{2M} \sum_{j=1}^{M} \log \frac{\exp\left(\tau \mathbf{g}_j^{\top} (\mathbf{h}_{j}^{\text{3D, con.}}) \right)}{\sum_{i=1}^{M} \exp\left(\tau \mathbf{g}_j^{\top} (\mathbf{h}_{i}^{\text{3D, con.}}) \right)} \\
& - \frac{1}{2M} \sum_{k=1}^{M} \log \frac{\exp\left(\tau (\mathbf{h}_{k}^{\text{3D, con.}})^{\top} (\mathbf{h}_{k}^{\text{2D, con.}}) \right)}{\sum_{l=1}^{M} \exp\left(\tau (\mathbf{h}_{k}^{\text{3D, con.}})^{\top} (\mathbf{h}_{l}^{\text{2D, con.}}) \right)} 
- \frac{1}{2M} \sum_{l=1}^{M} \log \frac{\exp\left(\tau (\mathbf{h}_{l}^{\text{2D, con.}})^{\top} (\mathbf{h}_{l}^{\text{3D, con.}}) \right)}{\sum_{k=1}^{M} \exp\left(\tau \mathbf{h}_{l}^{\text{2D, con.}})^{\top} (\mathbf{h}_{k}^{\text{3D, con.}}) \right)}.
\end{aligned}
\end{equation}

\textbf{Direct Alignment ($\mathcal{L}_{\text{dir.}}$)} The 3D microCT modality presents different intensity, texture, and resolved structures compared to the 2D H\&E images. Therefore, for accurate ST prediction with microCT data, it is imperative to minimize the gap between different imaging modalities and significantly leverage the first pretraining stage based on 2D H\&E imaging modality. To this end, we introduce a second alignment loss ($\mathcal{L}_{\text{dir}}$) that minimizes the Euclidean distance between the 2D image patch token embeddings $\{\mathbf{h}_{i,j}^{\text{2D, rec.}}\}_{j=1}^{32}$ and 3D image patch token embeddings $\{\mathbf{h}_{i,j}^{\text{3D, rec.}}\}_{j=1}^{32}$ from the reconstruction attentional pooler. The alignment of two modalities through minimizing Euclidean distance, instead of the contrastive approach, is inspired by an alternate approach for aligning multiple modalities\cite{yang2021multi}. The loss can be written as
\begin{equation}
\mathcal{L}_{\text{dir.}} = \frac{1}{M} \sum_{i=1}^{M} \sum_{j=1}^{32} \left\| \mathbf{h}_{i,j}^{\text{2D, rec.}} - \mathbf{h}_{i,j}^{\text{3D, rec.}} \right\|_2^2.
\end{equation}

\textbf{ST reconstruction ($\mathcal{L}_{\text{rec.},\mathrm{II}}$)} We minimize the MSE between the ground truth and the $\ours$-predicted gene expression. The loss can be written as
\begin{equation}
\mathcal{L}_{\text{REC}, \mathrm{II}} = \frac{1}{M} \sum_{i=1}^{M} \left\| \mathbf{y}_i - f_{\text{pred.}}\left(\{\mathbf{h}_{i,j}^{\text{3D, rec.}}\}_{j=1}^{32}\right) \right\|_2^2.
\end{equation}

The total loss minimized during this stage is defined as
\begin{equation}
    \mathcal{L}_{\mathrm{II}} = \lambda_{\text{cont.},\mathrm{II}}\cdot\mathcal{L}_{\text{cont.},\mathrm{II}} + 
\lambda_{\text{dir.}}\cdot\mathcal{L}_{\text{dir.}} +
\lambda_{\text{rec.},\mathrm{II}}\cdot\mathcal{L}_{\text{rec.}, \mathrm{II}},
\end{equation}
where $\lambda_{\text{cont.},\mathrm{II}}=\lambda_{\text{rec.},\mathrm{II}}=1$. The strength of direct alignment, $\lambda_{\text{dir.}}$ is set to 1 for microCT and 0 for 2.5D serial tissue sections as the 2D and 2.5D image data are composed of the same imaging modality. 
To maintain the number of training iterations consistent with the previous stage, the model is trained with a reduced batch size of 128. This is to account for a smaller number of available 3D morphology and ST pairs compared to the H\&E morphology and ST pairs curated from diverse sources. The model is trained for 15 epochs using a cosine scheduler that decays the learning rate from $1 \times 10^{-5}$ down to 0 by the end of training. The weight decay and optimizer settings are the same as in the first stage. Hyperparameters and training settings are provided in \textbf{Extended Data Table \ref{tab:stageII}}.

\noindent\textbf{Stage III: VOI fine-tuning}
In the final stage, we fine-tune $\ours$ with \textit{sample-specific} data to better capture the morphomolecular links of the VOI. During this stage, we fine-tune all layers that were trainable during previous stages. This covers the last three transformer layers of the 2D image encoder, the last three transformer layers of the transcriptomics encoder, the trainable layers of the 3D image encoder from the previous stage (none for microCT or three for serial tissue sections), the contrastive and reconstruction attentional poolers in both the 2D image encoder and 3D image encoder, and the transcriptomics predictor. 

During this stage, $\ours$ is trained using the same direct ($\mathcal{L}_{\text{dir.}}$) loss and reconstruction ($\mathcal{L}_{\text{rec.}, \mathrm{II}}$) loss as in training stage II. In addition, the contrastive loss is defined as the sum of the symmetric contrastive losses from stages I and II: $\mathcal{L}_{\text{cont.}} 
 = \mathcal{L}_{\text{cont.},\mathrm{I}} + \mathcal{L}_{\text{cont.},\mathrm{II}}$. 
To have a consistent number of training iterations as in the earlier stages, we use a reduced batch size of 16. This adjustment ensures a comparable number of batches for each training epoch, with data pairs from VOI significantly smaller than those from previous stages.
The model is fine-tuned for 10 epochs using a cosine scheduler that decays the learning rate from $1 \times 10^{-5}$ down to 0 by the end of training. The weight decay and optimizer settings are the same as in the first stage. Hyperparameters and training settings are provided in \textbf{Extended Data Table \ref{tab:stageIII}}.

\hheading{$\ours$ evaluation scenarios} 

We design three training strategies to evaluate $\ours$ to understand the effect of two important data scaling trends for 3D ST prediction for prostate cancer cohort: (1) the benefit of 3D morphological context to predict ST profiles (\textit{3D} or \textit{3D + VOI} scenario) as opposed to considering the 2D morphology (\textit{2D} scenario), and (2) the benefit of integrating VOI-specific training pairs (ST acquired on another tissue section from the VOI, 250 $\mu m$ from the section being evaluated) on top of training pairs from other volumes in the same cancer cohort (\textit{3D + VOI} scenario).

For the three scenarios, $\ours$ training stage I is commonly used. For the \textit{2D} setting, the model is further trained in stage II with 2D microCT sections and the corresponding 2D ST measurements obtained from the samples, excluding VOI. For encoding 2D microCT sections, we use the 3D image encoder with the input depth of 1, instead of 21. For the \textit{3D} setting, 3D microCT volumes of depth 21, instead of 2D microCT sections, around the 2D ST section are considered to incorporate 3D morphological context in stage II. 
For the \textit{3D+VOI} setting, we further fine-tune the model through Stage III with \textit{3D setting} on one of the two tissue sections with 2D ST measurement from VOI (P1-P5). We use the other remaining tissue section for evaluation. Next, we swap the roles of these two sections and re-perform fine-tuning and evaluation to obtain two measurements for each sample.  
All models are evaluated for each section with a tissue section leave-out-cross-validation strategy, resulting in ten distinctive results (two tissue sections each for five patients).

\hheading{Molecular queries for cross-modal retrieval}

To design a molecular query based on transcriptomic profiles of biological functions of interest, such as \textit{tumor cellular growth}, we start by combining two sets of genes. First, we identify one or two key genes (\textit{genes of interest}) involved in the biological process based on literature. Then, for these \textit{genes of interest}, we identify the set of  \textit{correlated genes}, defined as the genes whose expression levels across the ST spots are correlated with each of the \textit{genes of interest} with a Pearson Correlation Coefficient (PCC) greater than 0.5. PCC is evaluated for all genes that commonly exist across all samples in the same cancer cohort. In the prostate cancer cohort, for example, this consists of 8,136 genes, as explained in the previous  \textbf{Gene expression panel for evaluation} section. \textit{Genes of interest} and \textit{Correlated genes} for each molecular query in the prostate cancer cohort can be found in \textbf{Extended Data Table~\ref{tab:query_genes}}.

The union of \textit{genes of interest} and \textit{correlated genes}, which we refer to as \textit{spatial filter genes}, are then used for ST spot filtering within VOI to curate coherent expression profiles defining the molecular query. Specifically, we retain only the ST spots where at least half of the genes in the \textit{spatial filter genes} exhibit high expression, defined as being above the 75th percentile of expression values across all ST spots in VOI. Finally, we obtain the molecular query embedding by averaging the transcriptomic embeddings of the filtered spots.
The averaging operation is intended to remove inherent noise in individual measurements, similar to how text prompt embeddings are averaged in vision-language zero-shot cross-modal tasks\cite{radford2021learning, lu2024visual, ding2024multimodalslidefoundationmodel}. 
The molecular query is defined per VOI if ST data is available for the VOI or across the whole training set, otherwise. The former is the case of the prostate cancer cohort (analysis in \textbf{Figure \ref{fig:recall}}) and the latter is the case of the large CRC with serial histological sections (analysis in \textbf{Extended Data Figure \ref{fig:ext_CRC_query}}).

To identify morphological regions in the VOI that are most representative of the molecular queries, we divide the VOI into 3D patches and compute the cosine similarity between the normalized molecular query embedding and 3D patch embeddings. To minimize the sharp transition in similarity values for voxels at the boundary of neighboring patches, we employ the following sequence of operations: 3D image patches are created with 75\% overlap, cosine similarity is computed per patch, and the similarity values are averaged in the overlapping regions to achieve a smoother appearance in the similarity heatmaps. 
A coolwarm colormap, with red and blue colors indicating high and low similarity values, is then applied to the cosine similarity values and overlaid on the raw 3D image with a transparency value of 0.35. 
The minimum and maximum values for the colormap are set to the 10\textsuperscript{th} and 90\textsuperscript{th} percentile of the cosine similarity values for each molecular query and volume. The patches with the highest cosine similarity are also visualized as representative of the molecular query.

\hheading{Spatial domain identification}

We identify spatial domains in the tissue volumes by clustering the 3D patches based on 
their predicted gene expression information\cite{hu2021spagcn, dong2022deciphering, xiao2024integrating}. Specifically, we use the transcriptomic embedding before the last fully-connected layer of the transcriptomis predictor, immediately after the single-layer Transformer module. This yields 512-dimensional embedding, the dimensionality of which is independent of the final number of predicted genes. We subsequently cluster the set of transcriptomic embeddings in the tissue, by aggregating all the embeddings in the tissue volume and performing \textit{k}-means algorithm with four or five clusters, depending on the sample. This approach divides the tissue into functionally distinct regions in an unsupervised manner, guided by their transcriptomic expression profiles.
We evaluate the quality of the spatial domains by computing the Adjusted Rand Index (ARI) \cite{steinley2004properties} between the morphological segmentation masks and the spatial domains. We compute the ARI metric only across the locations of predicted gene expression and separately per each tissue section. Specifically, for a given axial section in the 3D image, we first obtain the morphological cluster assignment for each spot by referring to the corresponding morphological segmentation class. ARI metric is then computed between the morphological clusters and the spatial domains across all the spots within the tissue section. This process is repeated for every axial section in the 3D tissue image.

\hheading{3D visualization}

3D visualization was used for visualzing the $\ours$-predicted volumetric gene expression and the corresponding 3D morphological data. 3D renders were generated using Napari from 2D image stacks (along the z-axis) that represents the 3D tissue image data. 
The 3D ST prediction visualizations were generated from the stacks of 2D tissue images of gene predictions. For the prostate cohort, the \textit{3D+VOI} model was used for prediction of the spatial gene expression. 
To visualize spatial gene expression at high resolution, TESLA algorithm was used with a resolution factor of 15\cite{hu2023deciphering}. For each gene, the predicted transcriptomics expression levels were clipped at the 1st and 99th percentiles of the predicted expression levels in the central plane of the volume. This ensures a consistent scale of intensity across the volume. The data was then overlaid onto each plane with a transparency value of 0.7.

\hheading{3D morphological segmentation}

To generate 3D segmentations of microCT images, we leveraged the Segment Anything Model 2 (SAM2), a state-of-the-art video segmentation model trained on spatiotemporal datasets\cite{ravi2024sam2segmentimages}. Following the previously proposed methodology\cite{shen2024interactive3dmedicalimage}, we treated the sequential planes of CT volumes as video frames, enabling SAM2 to propagate segmentation masks annotated on a subset of planes to the entire 3D volume. The initial annotations were provided by a pathologist (A.K.) who labeled two evenly distributed H\&E tissue planes within each volume using polygon masks, capturing key anatomical structures. The annotations were then transferred to the registered microCT planes for each H\&E section. The adaptation and extension of a video-based segmentation paradigm to medical imaging significantly reduced the annotation burden without requiring domain-specific model retraining.

Within our pipeline, minimal pre- and post-processing steps (including normalization, clipping, morphological closing, and the application of a threshold-based foreground mask) were applied to refine the propagated masks, ensuring cleaner and more continuous boundaries across planes. For each sections annotated by the pathologist, the model propagated the segmentations both forward and backward, either to the next annotated section or to the end of the volume, whichever came first. This approach produced two sets of predictions for planes located between annotated sections. To ensure consistency in the model’s predictions when transitioning between annotated planes, we combined the two sets of predictions by taking a pixel-wise weighted average of the output logits for each class. The weights were scaled linearly, starting at one at the annotated plane where the propagation began and linearly decreasing to 0 at the next annotated plane, seamlessly blending the propagated segmentations.

\hheading{Evaluation metrics}

We evaluate $\ours$ using five metrics: Pearson Correlation Coefficient (PCC), Structural Similarity Index Measure (SSIM), Spearman's $\rho$, Moran's I, and Geary's C. These metrics are computed on a per-plane basis. When multiple planes with ground truth ST are available for a given VOI, we calculate the metrics for each plane, and then average values across all planes across the given cohort.

\noindent\textbf{PCC} offers insights into the linear relationship
between predicted and ground truth values, both in strength
and direction. It is one of the most common metrics for evaluating the quality of ST prediction from morphology\cite{zhang2024inferring,he2020integrating,chung2024accurate,wang2025benchmark}. We evaluate PCC for each gene across all spots in the plane. We compute the average across $M$ genes as
\begin{equation}
\operatorname{PCC} = \frac{1}{M} \sum_{i=1}^{M} \operatorname{PCC}_i.
\end{equation}

\noindent\textbf{SSIM} measures the similarity between the spatial structures of the ground truth and the predicted gene expression values. SSIM is an image similarity metric and has been applied to evaluate ST prediction tasks\cite{zhang2024inferring,wang2025benchmark}. A higher SSIM
indicates a higher degree of similarity between two images. For a given gene in each tissue section, we generate two single-channel images, one for the ground truth and the other for predicted ST expression values. The ST spot coordinates are first downscaled by the factor equivalent to the center-to-center distance of ST spots, which yields a dense 2D pixel grid. Each pixel corresponds to a ST spot, with the expression values scaled to $[0,1]$ using min-max normalization. 
SSIM is then computed for each gene and averaged across gene sets, following a procedure similar to that used for the PCC metric.

\noindent\textbf{Spearman's $\mathbf{\rho}$} To better assess how effectively the different $\ours$ training strategies recapitulate the variance of the genes being predicted, we compare the variance of each gene in the ground truth ST data to that of the predicted gene expression, following previous work\cite{xie2024spatially}.
Genes are ranked based on their ground truth variance (from smallest to largest), and curves are generated for both the original and predicted gene expression variances. Spearman's rank correlation coefficient (Spearman's $\rho$) is then computed to quantify the similarity between the two distributions.

\noindent\textbf{Moran's I and Geary's C} While PCC and SSIM evaluate the correct prediction of each gene mean expression, Spearman's $\rho$ assesses the variance of the predictions. In addition, to evaluate how well $\ours$'s training strategies capture the distribution of gene expression patterns across spatial locations in the tissue, we consider Moran's I \cite{moran1950notes} and Geary's C \cite{geary1954contiguity}, two classical spatial autocorrelation metrics widely employed to identify spatially variable genes. We evaluate these metrics per gene and we compute the average across all genes. We then report the difference of each metric between the ground truth and $\ours$-prediction, with smaller values indicating better captures of the expression heterogeneity.

\hheading{Statistical analysis}

For each training scenario (\textit{2D}, \textit{3D} and \textit{3D+VOI}), we evaluate the model performance on each of the two sections available per patient (P1$\sim$P5) separately. We report the mean performance and the standard deviations across all 10 planes. We use a one-sided Wilcoxon signed-rank test to evaluate the statistical significance between the three settings, for all evaluation metrics.

\hheading{Computational hardware and software.}
3D Spatial Transcriptomics on volumetric images via $\ours$ was performed on AMD Ryzen multicore CPUs (central processing units). Two NVIDIA GeForce RTX 3090 GPUs (graphics processing units) were used for the 2D pretraining, and one GPU of the same specifications was used for the following training stages (3D pretraining and VOI fine-tuning). $\ours$ was implemented in Python (version 3.10.13). All deep learning implementations were performed with PyTorch (version 2.1.2). The implementation of scGPT from 
(\texttt{\href{https://github.com/bowang-lab/scGPT}
{\url{https://github.com/bowang-lab/scGPT}}}) was used, which required flash-attn (version 1.0.4). The loss function for contrastive alignment was adapted from (\texttt{\href{ https://github.com/moein-shariatnia/OpenAI-CLIP.git}{\url{ https://github.com/moein-shariatnia/OpenAI-CLIP.git}}}). Processing and analysis of spatial transcriptomics data was performed using scanpy (version 1.10.1). Generation of ST super-resolution data for visualization was generated with TESLA from (\texttt{\href{https://github.com/jianhuupenn/TESLA}{\url{ https://github.com/jianhuupenn/TESLA}}}). Valis-wsi (version 1.0.4) was used for serial section registration and pyRANSAC from (\texttt{\href{https://github.com/leomariga/pyRANSAC-3D/}{\url{https://github.com/leomariga/pyRANSAC-3D/}}}) was used for 3D point cloud fitting during cross-modal registration.
Evaluation metrics for $\ours$ predictive performance used numpy (version 1.26.4), and scikit-image (version 0.19.3).
Other Python libraries used to support data analysis include slideio (version 2.5.0), tifffile (version 2024.5.10), pandas (version 2.2.2), scipy (version 1.13.0), pillow (version 9.5.0), opencv-python (version 4.9.0), torchvision (version 0.16.0), and timm (version 1.0.3). Plots were generated in Python using matplotlib (version 3.9.0). 3D visualization was accomplished via napari (version 0.4.16). The interactive demo website was developed using THREE.js (version 0.152.2) and jQuery (version 3.6.0).

\paragraph{Data and code availability} 

$\ours$ code will be made available upon publication acceptance. We will release all microCT 3D tissue images and their corresponding Visium ST measurements for research and education use.

\section*{Author Contributions}
C.A.P, A.H.S, and F.M conceived the study and designed the experiments. A.H.S, D.F.K.W, and B.C imaged the microCT dataset. A.K provided annotations for all volumes. C.A.P, A.H.S, A.K, and D.F.K.W identified patient blocks for Visium sequencing. A.H.S worked with the University of Michigan Advanced Genomics Core for ST sequencing on samples. L.W developed 3D morphological segmentation framework and the interactive demo. C.A.P and A.H.S created and ran all the experiments. G.J, K.H, M.Y.L, and K.S helped analyze experiment results. C.A.P, A.H.S, and F.M prepared the manuscript. All authors contributed to the writing. F.M supervised the research.

\section*{Acknowledgements}
We thank the University of Michigan Advanced Genomics Core for assistance with Visium spatial transcriptomics sequencing.
This work was funded in part by the Brigham and Women’s Hospital (BWH) President’s Fund, Mass General Hospital (MGH) Pathology and by the National Institute of Health (NIH) National Institute of General Medical Sciences (NIGMS) through R35GM138216. M.Y.L was supported by the Tau Beta Pi Fellowship and the Siebel Foundation. The content is solely the responsibility of the authors and does not reflect the official views of the NIH, and NIGMS.


\end{spacing}

\clearpage

\newpage
\setcounter{figure}{0}
\renewcommand{\figurename}{\textbf{Extended Data Figure}}
\renewcommand{\thefigure}{\arabic{figure}}

\setcounter{table}{0}
\renewcommand{\tablename}{\textbf{Extended Data Table}}

\clearpage

\begin{figure*}[!ht]
\centering  
\includegraphics[width=\textwidth]{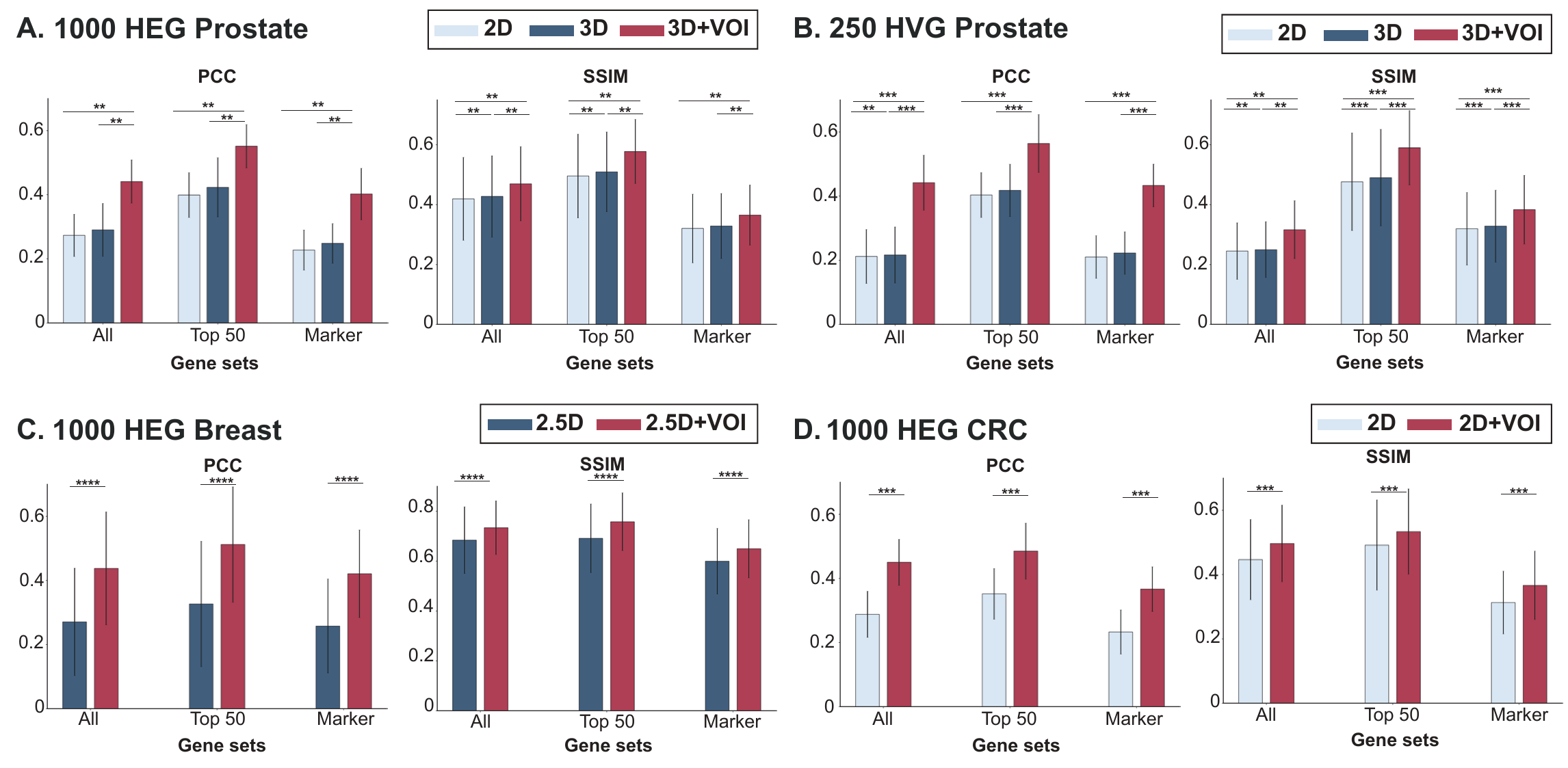}
\caption{\textbf{$\ours$ ST prediction analysis on additional gene sets}. In addition to the analysis for 250 highly-expressed genes (HEG) in \textbf{Figure~\ref{fig:prostate}}, we analyze $\ours$ for gene sets with \textbf{(a)} 1,000 HEG and \textbf{(b)} 250 highly variable genes (HVG) over three different scenarios. Error bars indicate one standard deviation from the mean, over ten sections across five patients. In addition to the analysis for 250 highly-expressed genes (HEG) in \textbf{Extended Figure \ref{fig:breast_crc}}, we analyze $\ours$ for gene sets with 1,000 HEG over two different scenarios for \textbf{(c)} the breast cancer cohort and \textbf{(d)} the colorectal cancer cohort. Statistical significance was assessed with the Wilcoxon signed-rank test. $^{\ast\ast}p\leq 0.01$, $^{\ast\ast\ast}p\leq 0.001$, $^{\ast\ast\ast\ast}p\leq 0.0001$. PCC: Pearson Correlation Coefficient. SSIM: Structural Similarity Index Measure.}
\label{fig:ext_1000HEG_and_HVG}
\end{figure*}

\clearpage

\begin{figure*}[!ht]
\centering  
\includegraphics[width=\textwidth]{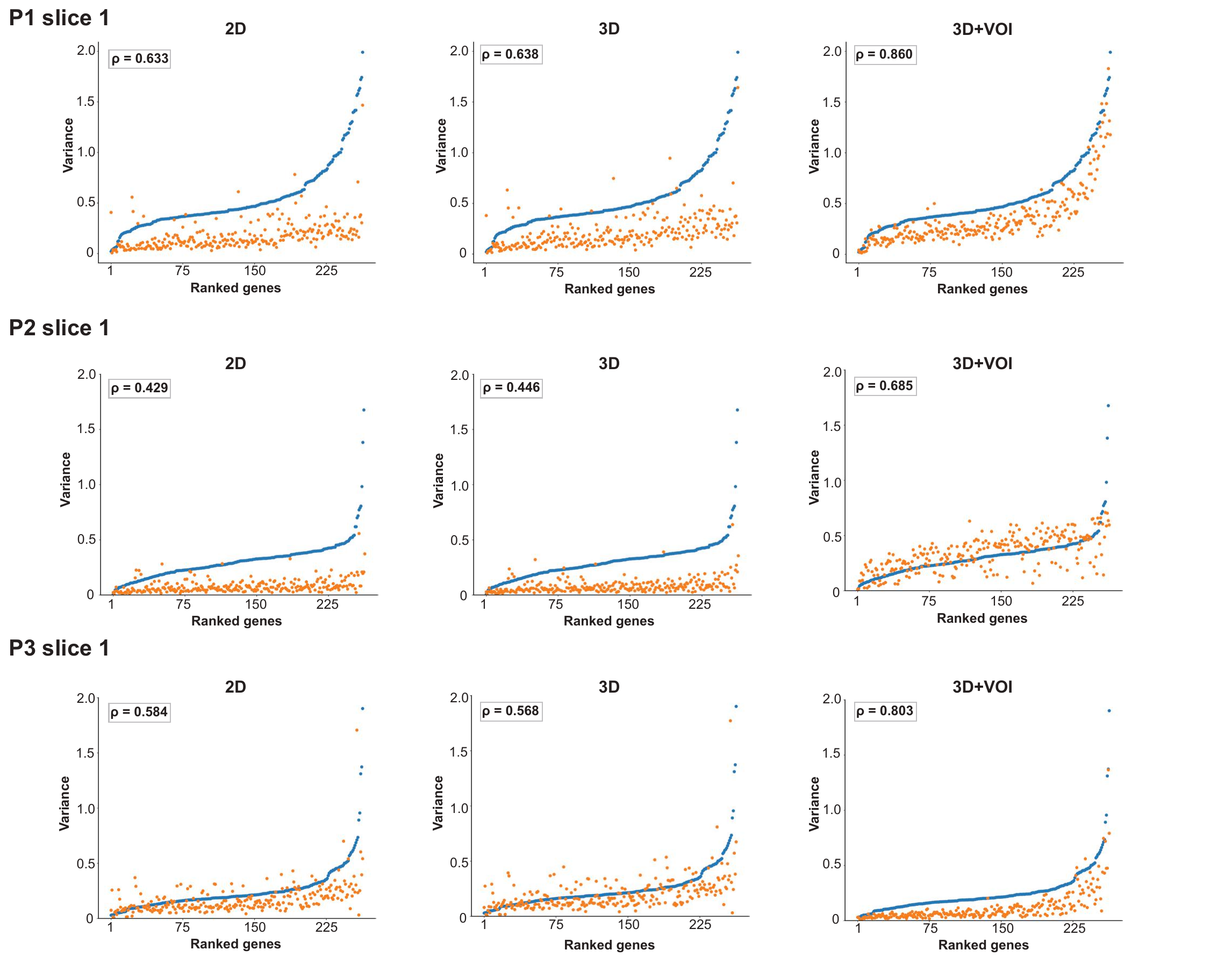}
\caption{\textbf{$\ours$ ST prediction analysis on gene expression variance}. The correlation Spearman's $\rho$ between the variance of $\ours$-predicted expression levels (orange) and the variance of measured ST expression levels (blue) across all Visium ST spots in each tissue section (refer to \textbf{Online Methods} in section \textbf{ST spot filtering and expression normalization}). Genes are ranked based on measured ST expression variance, from the smallest to the largest. The variance measures are shown across three different scenarios for three exemplar sections.}
\label{fig:ext_variance_ablation}
\end{figure*}

\clearpage

\begin{figure*}[!ht]
\centering  
\includegraphics[width=\textwidth]{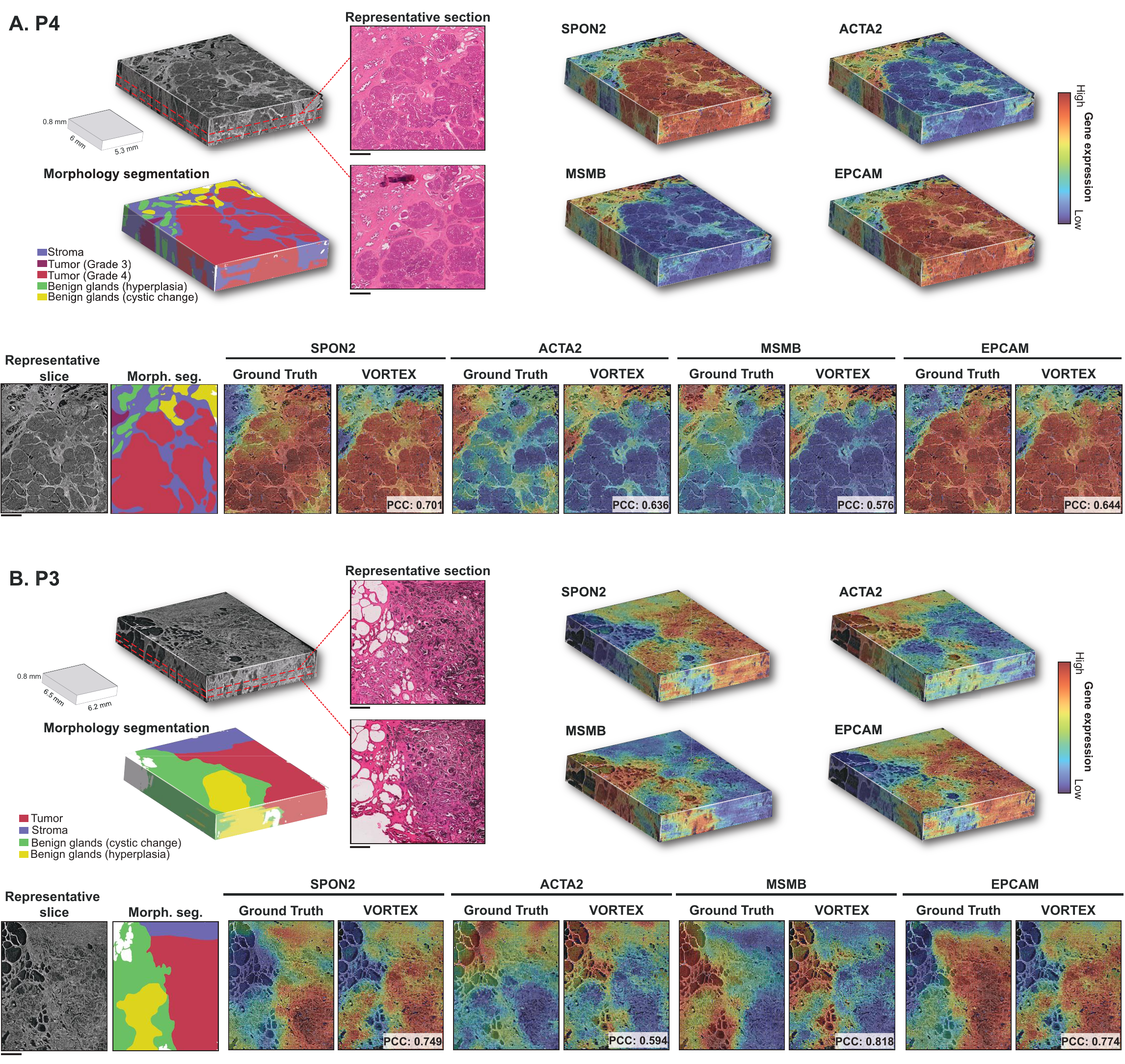}
\caption{\textbf{$\ours$ analysis on prostate cancer cohort}. The predicted 3D ST heatmaps for four representative genes (\textit{SPON2}, \textit{ACTA2}, \textit{MSMB}, and \textit{EPCAM}) along with 3D tissue images captured with microCT, 3D morphological segmentation, and representative
2D H\&E-stained histological sections for each of the two patients on which Visium ST is measured. The bottom row for each patient shows the measured ST expression and $\ours$-predicted expression (\textit{3D+VOI} setting) along with Pearson Correlation Coefficient (PCC) for evaluating prediction capacity. During \textit{VOI} fine-tuning, ST data from one representative section is first used for training while evaluation is performed on the second section. The process is then reversed, with the second plane used for training and the first plane for evaluation. We observe that \textit{SPON2} and \textit{EPCAM} genes are overexpressed in tumor regions, \textit{MSMB} gene is downregulated in prostatic tumor glands compared to benign glands, and \textit{ACTA2} gene is overexpressed in stromal tissue regions, aligning with previous findings in literature\cite{berglund2018spatial,song2022single}. Examples for P1 are in \textbf{Figure~\ref{fig:prostate}}. Scalebar is 1 mm.}
\label{fig:ext_prostate_3D}
\end{figure*}

\clearpage

\begin{figure*}[!ht]
\centering  
\includegraphics[width=\textwidth]{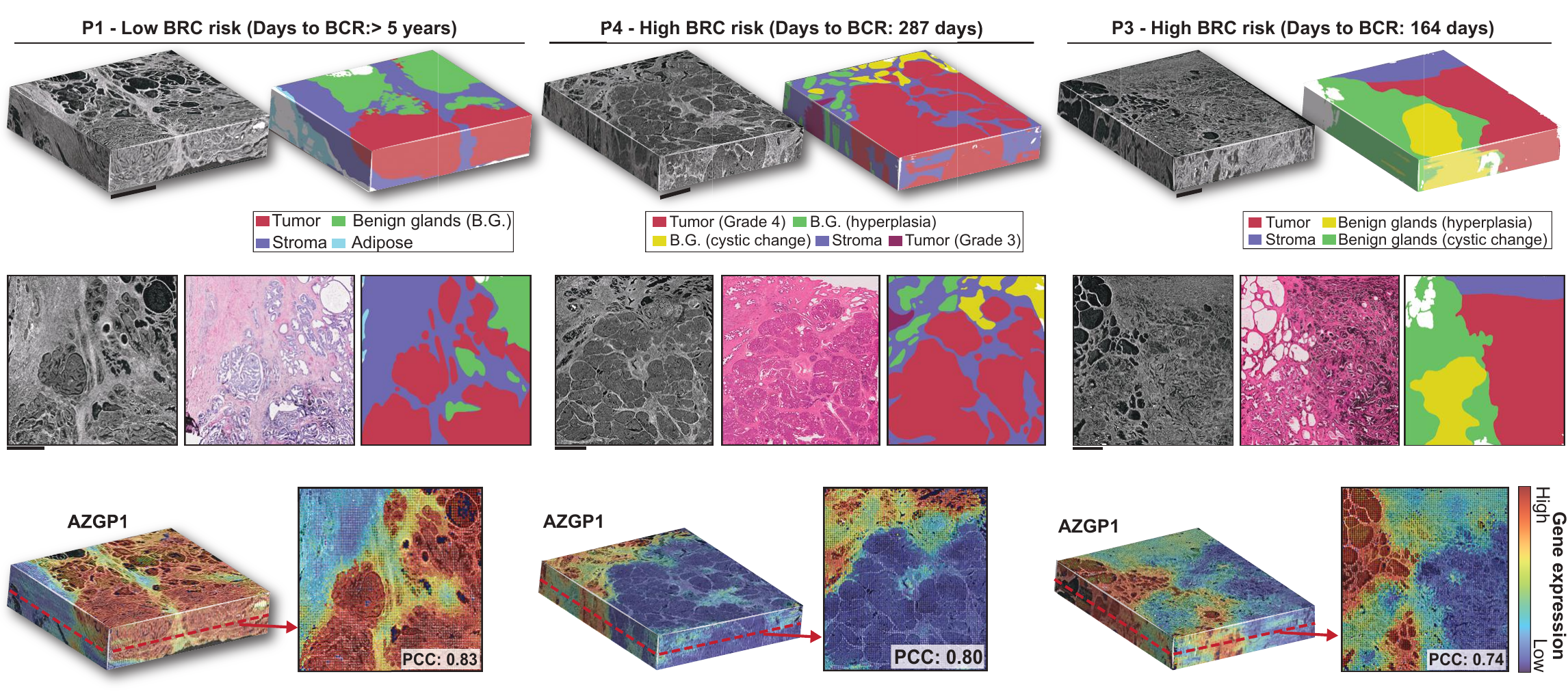}
\caption{\textbf{$\ours$ captures inter-tumoral heterogeneity}. 3D ST prediction by VORTEX of \textit{AZGP1} gene on three samples with different biochemical recurrence (BCR) status. \textit{AZGP1} downregulation in prostate adenocarcinoma is associated with shorter time to BCR\cite{kristensen2019predictive,burdelski2016reduced}. 
$\ours$ captures the inter-tumoral heterogeneity that agrees with patient BCR status and predicts high expression of \textit{AZGP1} in tumoral regions for low-risk sample P1, and low expression for high-risk samples P3 and P4. Scalebar is 1 mm.}
\label{fig:ext_intertumoral}
\end{figure*}

\clearpage
\begin{figure*}[!ht]
\centering  
\includegraphics[width=\textwidth]{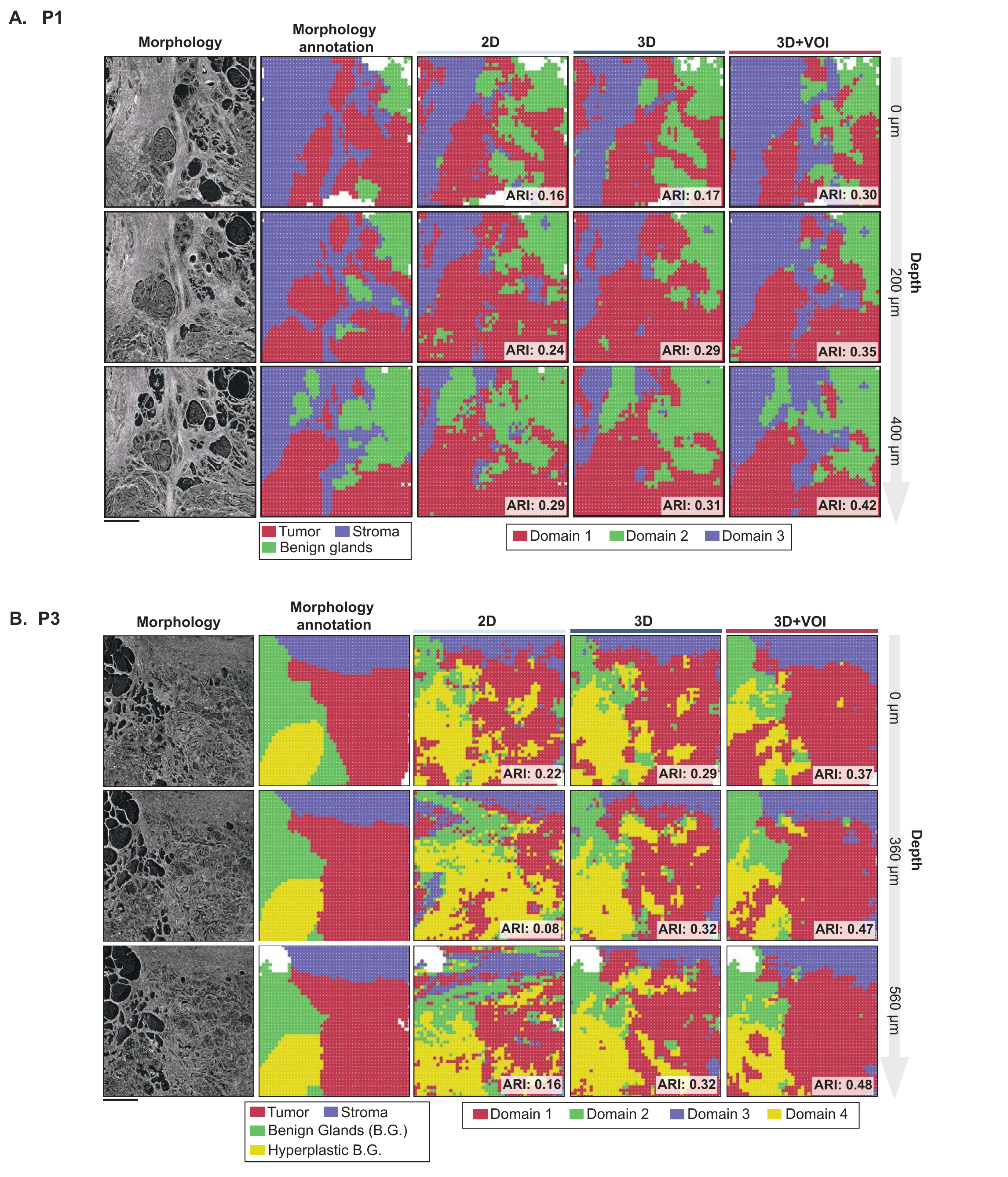}
\caption{\textbf{3D Spatial Domain identification with $\ours$}. Spatial domains across the tissue volumes for two patients (P1 and P3). The \textit{3D + VOI} setting shows higher degree of agreement with the manually annotated morphology by a pathologist. Adjusted Rand Index (ARI) scores are displayed. Scalebar is 1~mm.}
\label{fig:ext_spat_cluster}
\end{figure*}

\clearpage

\begin{figure*}[!ht]
\centering  
\includegraphics[width=\textwidth]{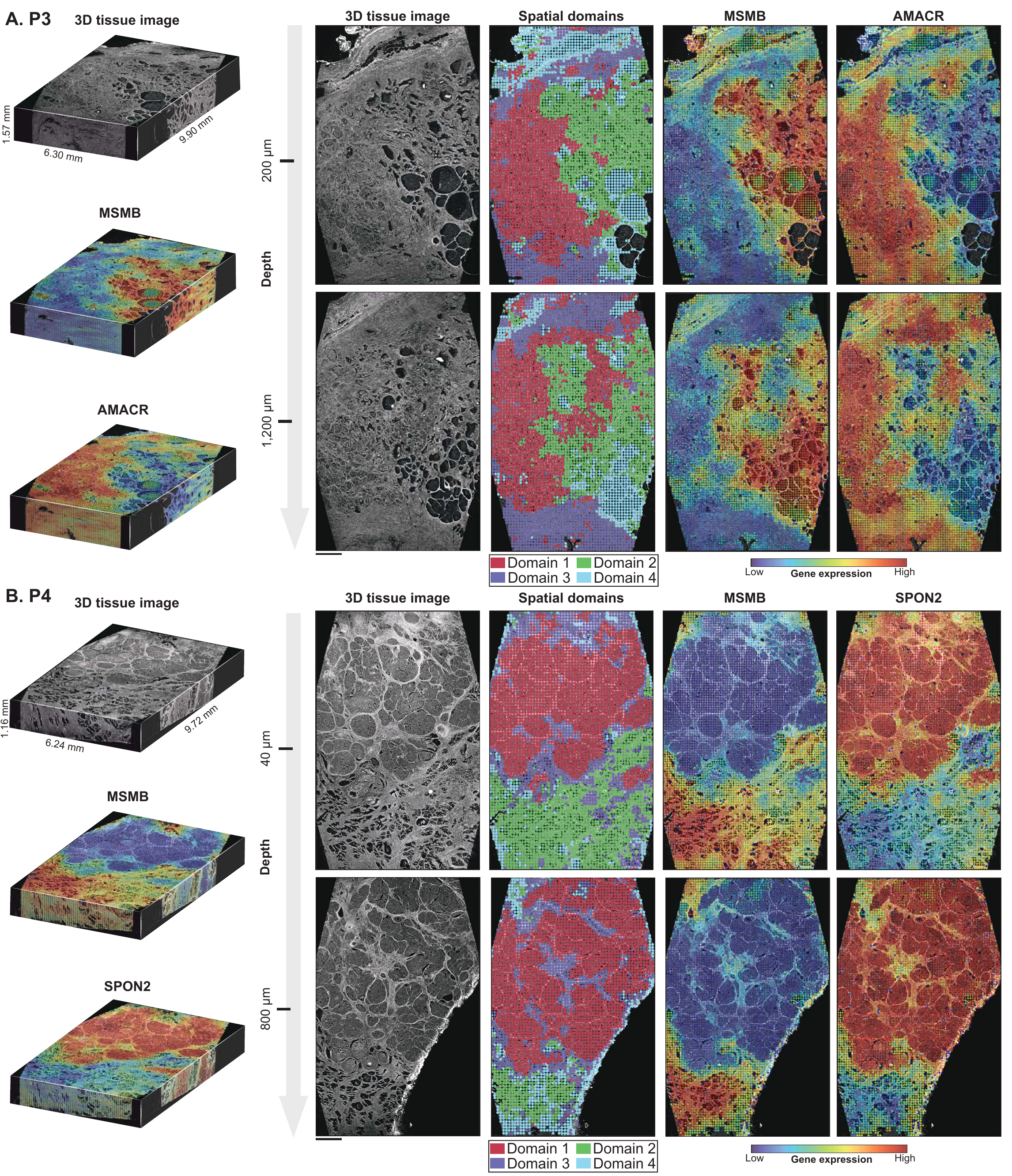}
\caption{\textbf{$\ours$ on large prostate cancer tissue}. 
3D ST prediction by $\ours$ on large prostate cancer tissue volumes for \textit{MSMB} and \textit{AMACR} genes in sample P3 and 
\textit{SPON2} and \textit{MSMB} in sample P4. Cross-sections at different depths are shown, along with the spatial domains identified by $\ours$. In P3, spatial domain (S.D.) 1 predominantly corresponds to adenocarcinoma, S.D. 2 to hyperplastic benign glands, S.D. 3 to stroma, and S.D. 4 to luminal areas of benign glands with cystic change and adventitia. In P4, S.D. 1 predominantly corresponds to adenocarcinoma, S.D. 2 to benign prostatic glands, S.D. 3 to intratumoral stroma and S.D. 4 to luminal areas and tissue edges. Scalebar is 1 mm.}
\label{fig:ext_prostate_3D_largeFOV}
\end{figure*}

\clearpage
\begin{figure*}[!ht]
\centering  
\includegraphics[width=0.9\textwidth]{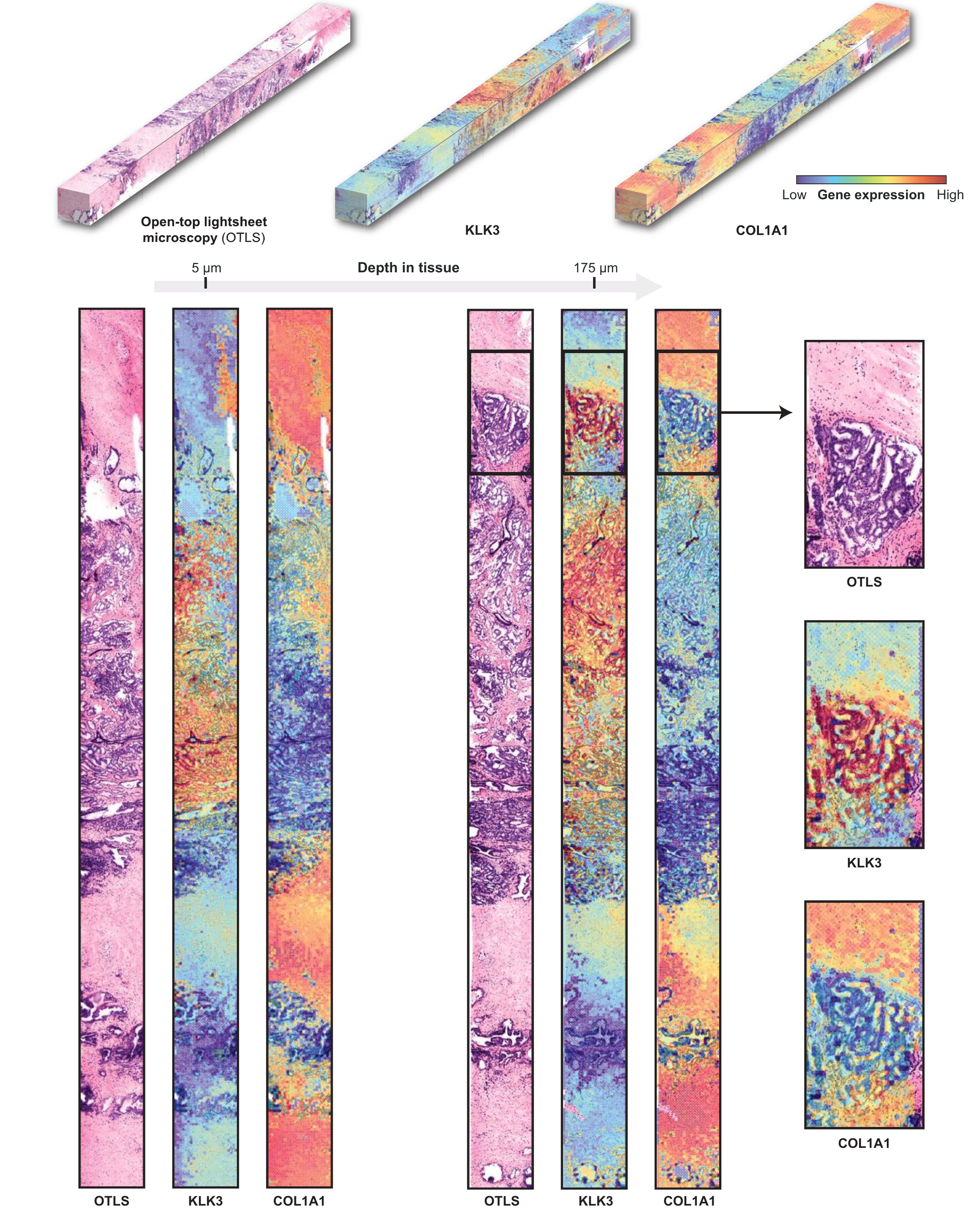}
\caption{\textbf{$\ours$ with prostate cancer sample imaged with open-top lightsheet microscopy (OTLS)}. Simulated prostate core needle biopsy was imaged with OTLS at $1 \mu m$/voxel, which was converted to provide H\&E-like appearance of the 3D tissue sample. $\ours$ pretrained on 2D H\&E image and ST pairs is applied to OTLS. The agreement with the tumoral and stromal regions with up-regulation of \textit{KLK3} and \textit{COL1A1} respectively demonstrates its generalizability across imaging modalities.}
\label{fig:ext_otls}
\end{figure*}

\clearpage

\begin{figure*}[]
\centering
\includegraphics[width=\textwidth]{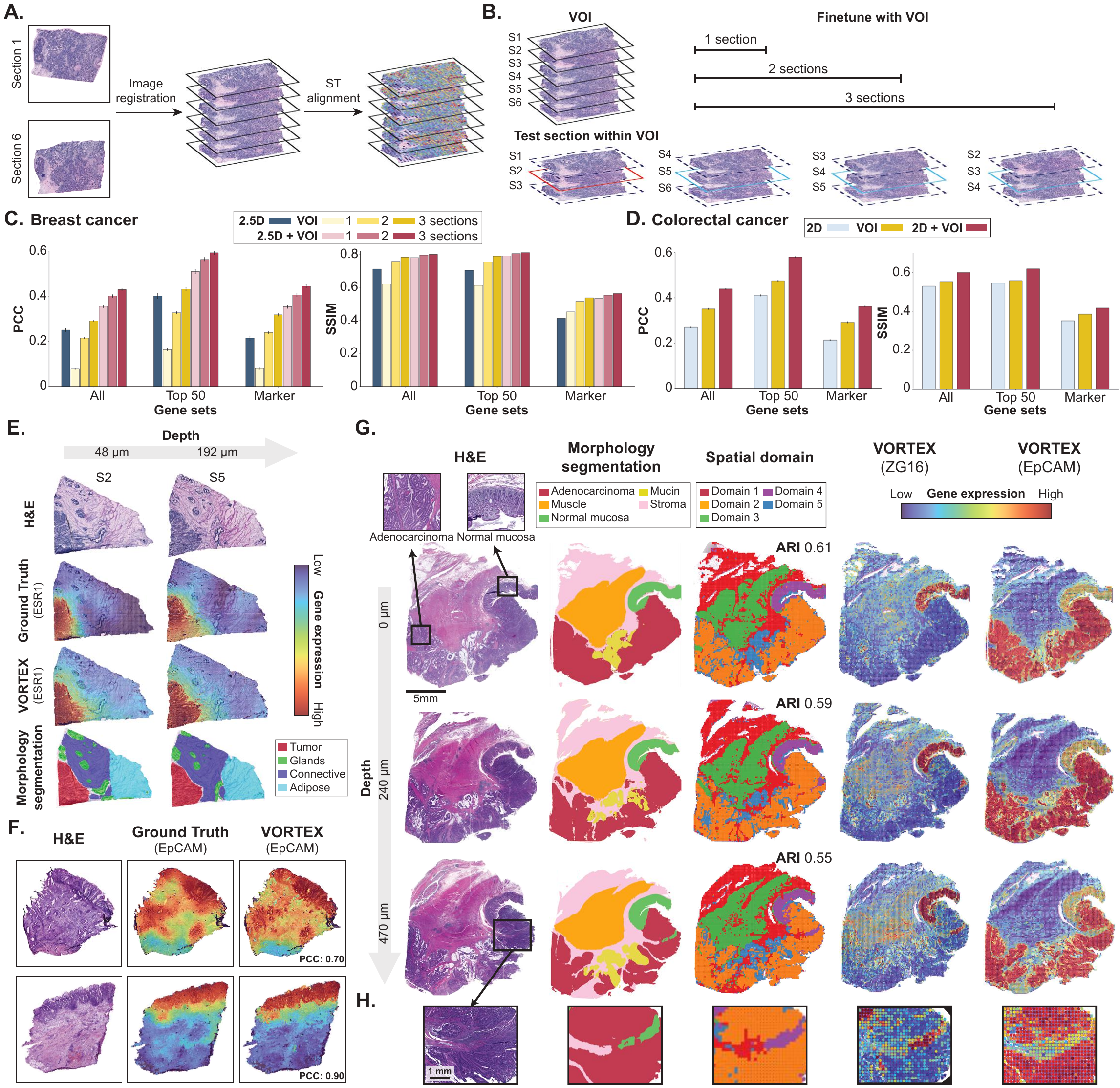}
\caption{\textbf{$\ours$ analysis on breast and colorectal cancer cohort}. 
\textbf{(a)} H\&E tissue images and ST are obtained from serial tissue sections sampled from breast cancer volumes. 2.5D tissue image is constructed by registering serial sections\cite{gatenbee2023virtual}.
\textbf{(b)} Schematic for evaluation of performance change with increasing ST sections within VOI.
\textbf{(c)} PCC and SSIM between the predicted and the measured ST expression for four breast cancer patients. For fine-tuning with VOI, performance is shown over an increasing number of sections with ST measurements without cohort-level pretraining (yellow) and with pretraining (red) as illustrated in (b).
\textbf{(d)} PCC and SSIM between the predicted and the measured ST expression for three gene sets averaged across six colorectal cancer patients with two sections each.
\textbf{(e)} 2.5D ST heatmap of \textit{ESR1} with measured expression, H\&E, and morphology segmentation.
\textbf{(f)} H\&E, measured, and predicted \textit{EpCAM} expression from two CRC samples. PCC between the measured and predicted values is displayed. 
\textbf{(g)} 2.5D ST heatmaps for large colorectal cancer volume with 22 serial sections\cite{lin2023multiplexed}, morphology segmentation, spatial domains, and predicted expression profiles. Additional examples can be found in \textbf{Extended Data Figure~\ref{fig:ext_CRC_3D}}. 
\textbf{(h)} Zoomed-in region with `cord-like' structure of stroma (left half) and normal mucosa (right half). S: Section. PCC: Pearson correlation coefficient. SSIM: Structural Similarity Index Measure. VOI: Volume of interest.}
\label{fig:breast_crc}
\end{figure*} 

\clearpage

\begin{figure*}[!ht]
\centering  
\includegraphics[width=\textwidth]{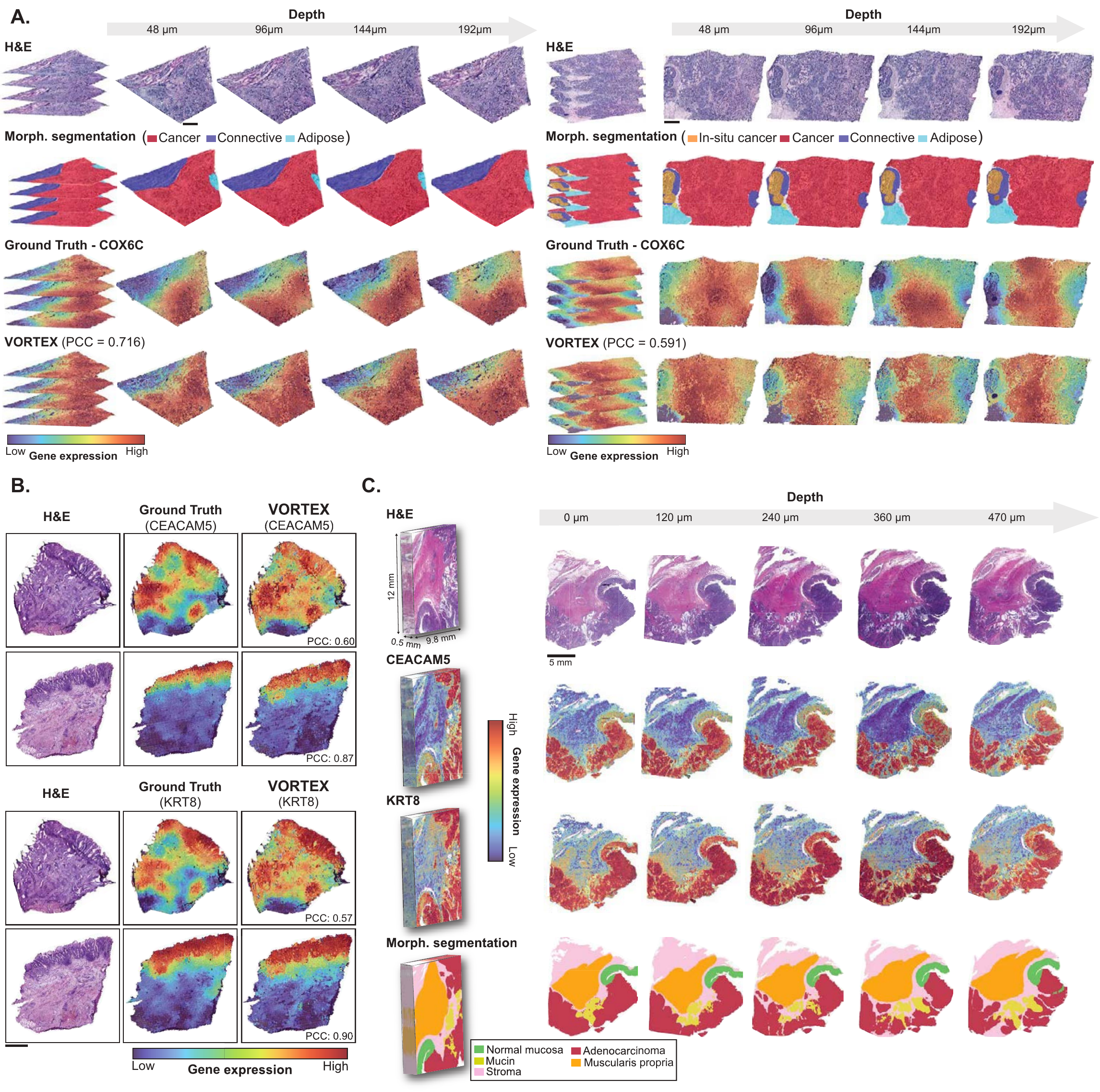}
\caption{\textbf{$\ours$ analysis on breast and colorectal cancer cohorts: additional visualizations}. \textbf{(a)} 2.5D ST heatmaps of predicted and measured gene \textit{COX6C}, which plays a crucial role in
the identification of hormone-responsive breast cancer\cite{west2001predicting}. The four central sections, with 2.5D morphological context (three sections) comprised of a section of interest and a neighboring section above and below, out of the six total planes for each patient are shown.
\textbf{(b)} 2.5D ST heatmaps of predicted and measured gene \textit{CEACAM5} and \textit{KRT8}, which are upregulated in tumoral tissue compared to normal colonic mucosa\cite{xiao2024integrating}. PCC between the measured and the predicted ST expressions is displayed. \textbf{(c)} 2.5D ST heatmaps obtained with $\ours$ for publicly-available CRC sample with 22 serial tissue sections and tissue segmentation with representative 2D axial section. 
Unless specified otherwise, scalebar is 1 mm. PCC: Pearson correlation coefficient. Morph. segmentation: morphology segmentation.}
\label{fig:ext_CRC_3D}
\end{figure*}

\clearpage

\begin{figure*}[!ht]
\centering  
\includegraphics[width=\textwidth]{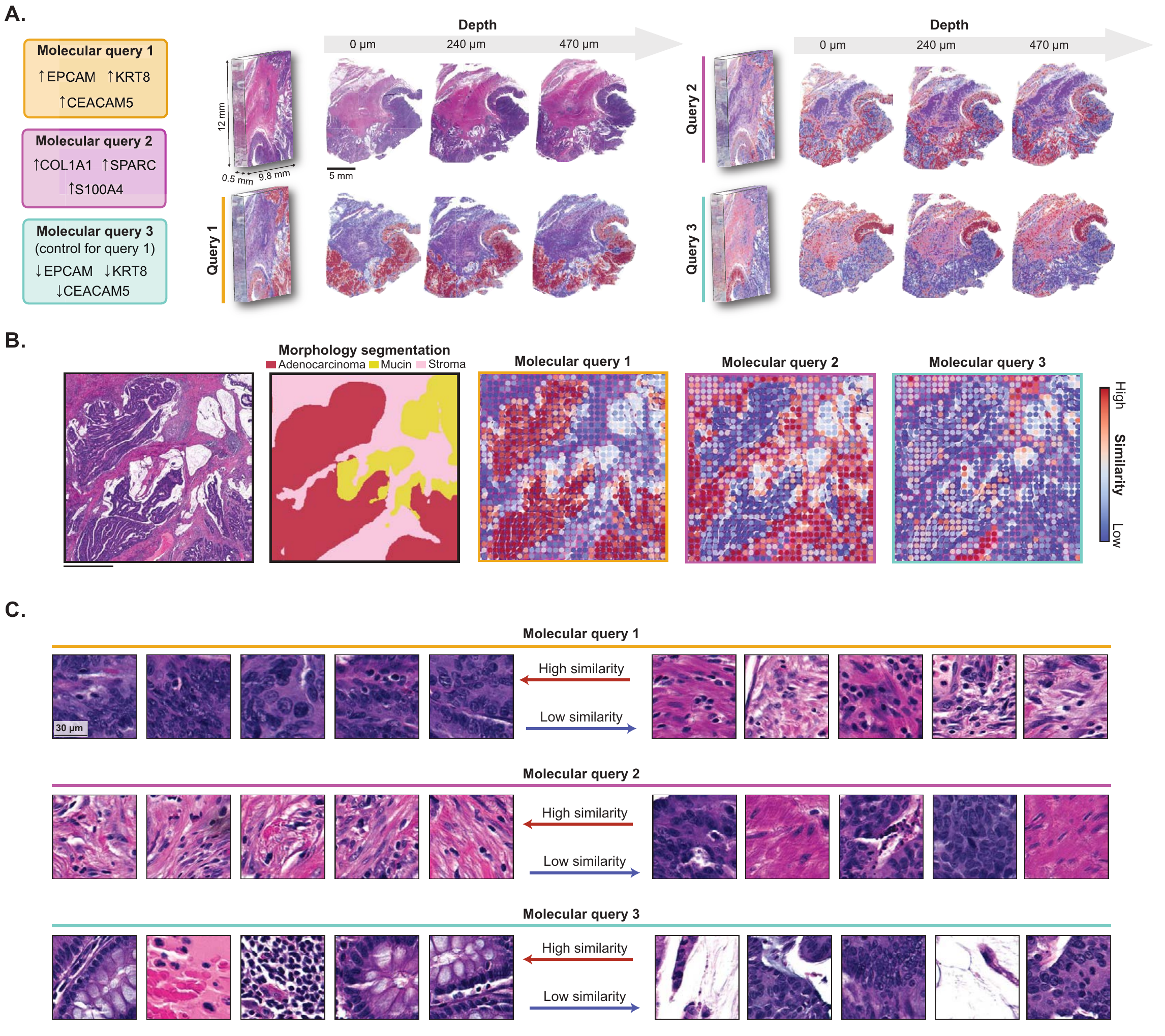}
\caption{\textbf{Cross-modal morphology retrieval with $\ours$ on colorectal cancer cohort.} \textbf{(a)} Molecular query analysis for large 2.5D colorectal cancer tissue volume.
We design two molecular queries defined by up-regulation of \textit{EPCAM}, \textit{KRT8}, \textit{CEACAM5} and up-regulation of \textit{COL1A1}, \textit{SPARC}, \textit{S100A4}. The first and second molecular queries largely correspond to adenocarcinoma and stroma, respectively, as can be seen with the complementary similarity heatmap around adenocarcinoma and the top similar patches. Additionally, we introduce a third molecular query as a control for the first query, by using low-expression of \textit{EPCAM}, \textit{KRT8}, \textit{CEACAM5}. \textbf{(b)} Close-up view of an image region containing colon adenocarcinoma, mucin, and stroma. Complementary heatmap of molecular query 1 and 2 is observed between adenocarcinoma and surrounding stroma. \textbf{(c)} Examples of top similar and dissimilar 2D patches for each molecular query are shown. For molecular query 1, tumor patches characterize the high similarity group while low similarity is observed in patches containing stroma and smooth muscle (muscularis propria). For molecular query 2, stroma patches represent the high similarity group while other tissue components such as adenocarcinoma and muscularis propria constitute the low similarity set. For molecular query 3 (control for molecular query 1), high similarity patches are composed of heterogeneous morphology, while tumor and secreted mucin constitute the low similarity group. Unless specified otherwise, scalebar is 1 mm.
}
\label{fig:ext_CRC_query}
\end{figure*}

\clearpage
\begin{table}[!h]
\centering
\caption{\textbf{Dataset for $\ours$ training and evaluation}.}
\tiny
\begin{tabular}{|l|l|p{2cm}|l|l|l|l|l|}
\hline
\textbf{Organ/}\textbf{Dataset}                                                      & \textbf{Disease state}                                                                                                                                         & \textbf{Data source}                                                                                          & \textbf{Imaging modality}                                                                                         & \textbf{Dataset dimensions}                                                                                                                                                                                                                                                & \textbf{Protocol}                                                               & \multicolumn{1}{c|}{\textbf{\begin{tabular}[c]{@{}c@{}}Spot diameter \\ (µm)\end{tabular}}} & \multicolumn{1}{c|}{\textbf{\begin{tabular}[c]{@{}c@{}}Distance between \\ spots (µm)\end{tabular}}} \\ \hline
Prostate I                                                            & Prostate adenocarcinoma                                                                                                                                        & Internal                                                                                                      & \begin{tabular}[c]{@{}l@{}}FFPE H\&E-stained \\ histological sections and \\ microCT images\end{tabular}          & \begin{tabular}[c]{@{}l@{}}-Num. Spots: 65,715\\ -Num. Sections: 16\\ -Num. Patients: 11\\ -Num. Genes: 17,943\end{tabular}                                                                        & Visium                                                                          & 55                                                                                          & 100                                                                                                  \\ \hline
Prostate II                                                           & \begin{tabular}[c]{@{}l@{}}-Healthy (N=1)\\ -Prostate acinar cell \\  carcinoma (N=1)\\ -Prostate adenocarcinoma \\ with invasive carcinoma (N=1)\end{tabular} & 10x Genomics (blocks 1E333\_Tn17, 1E500\_Tp12 Section 1, and 1D1061-Tp11 Section 1)                                            & \begin{tabular}[c]{@{}l@{}}FFPE H\&E \\ histological sections\end{tabular}                                        & \begin{tabular}[c]{@{}l@{}}-Num. Spots: 9,957 \\ -Num. Sections: 3 \\ -Num. Patients: 3\\ -Num. Genes: 17,943\end{tabular}                                                                                                                                                 & Visium                                                                          & 55                                                                                          & 100                                                                                                  \\ \hline
Prostate III                                                            & Prostate adenocarcinoma                                                                                                                                        & \href{https://data.mendeley.com/datasets/svw96g68dv/1}{\url{https://data.mendeley.com/datasets/svw96g68dv/1}}                                                               & \begin{tabular}[c]{@{}l@{}}Frozen H\&E-stained \\ histological sections\end{tabular}                              & \begin{tabular}[c]{@{}l@{}}-Num. Spots: 58,906\\ -Num. Sections: 22\\ -Num. Patients: 2 \\ -Num. Genes: 33,538\end{tabular}                                                                                                                                                & Visium                                                                          & 55                                                                                          & 100                                                                                                  \\ \hline
Prostate IV                                                            & Prostate adenocarcinoma                                                                                                                                        & \href{https://data.mendeley.com/datasets/mdt8n2xgf4/1}{\url{https://data.mendeley.com/datasets/mdt8n2xgf4/1}}                                                              & \begin{tabular}[c]{@{}l@{}}Frozen H\&E-stained \\ histological sections from\\  core needle biopsies\end{tabular} & \begin{tabular}[c]{@{}l@{}}-Num. Spots: 3,969\\ -Num. Sections: 24\\ -Num. Patients: 3\\ -Num. Genes (mean): 24,945 \\ (Range: 19,537 - 30,157)\end{tabular}                                                                                                               & \begin{tabular}[c]{@{}l@{}}Spatial \\ Transcriptomics\end{tabular}              & 100                                                                                         & 200                                                                                                  \\ \hline
Breast I                                                              & HER2+ breast cancer                                                                                                                                            & \begin{tabular}[c]{@{}l@{}}Andersson et al. \tiny \cite{andersson2021spatial} \\ Data retrieved from \\ \href{https://huggingface.co/datasets/MahmoodLab/hest}{HEST-1k database} \end{tabular} & \begin{tabular}[c]{@{}l@{}}Frozen H\&E-stained \\ histological sections\end{tabular}                              & \begin{tabular}[c]{@{}l@{}}-Num. Spots: 13,299\\ -Num. Spots with associated \\  3D morphology: 6,577\\ -Num. Sections: 36\\ -Num. Sections with associated \\ 3D morphology: 16\\ -Num. Patients: 8\\ -Num. Genes (mean): 15,364 \\ (Range: 14,861 - 15,842)\end{tabular} & \begin{tabular}[c]{@{}l@{}}Spatial \\ Transcriptomics\end{tabular}              & 100                                                                                         & 200                                                                                                  \\ \hline
Breast II                                                              & Invasive ductal carcinoma                                                                                                                                      & \begin{tabular}[c]{@{}l@{}}B. He et al. \cite{he2020integrating}, \\ Data retrieved from \\ HEST-1k database.\end{tabular}        & \begin{tabular}[c]{@{}l@{}}Frozen H\&E-stained \\ histological sections\end{tabular}                              & \begin{tabular}[c]{@{}l@{}}-Num. Spots: 23,699\\ -Num. Sections: 56\\ -Num. Patients: N/A\\ -Num. Genes (mean): 18,010 \\ (Range: 16,744 - 19,729)\end{tabular}                                                                                                            & \begin{tabular}[c]{@{}l@{}}Spatial \\ Transcriptomics\end{tabular}              & 100                                                                                         & 200                                                                                                  \\ \hline
Breast III                                                              & Invasive ductal carcinoma                                                                                                                                      & \begin{tabular}[c]{@{}l@{}}10x Genomics. \\ Data retrieved from \\ HEST-1k database.\end{tabular}                 & \begin{tabular}[c]{@{}l@{}}Frozen (N=2) \& \\ FFPE (N=2) H\&E-stained \\ histological sections\end{tabular}       & \begin{tabular}[c]{@{}l@{}}-Num. Spots: 20,236\\ -Num. Sections: 4\\ -Num. Patients: 3\\ -Num. Genes (mean): 27,971 \\ (Range: 17,943 - 36,601)\end{tabular}                                                                                                               & Visium                                                                          & 55                                                                                          & 100                                                                                                  \\ \hline
Breast IV                                                              & Invasive ductal carcinoma                                                                                                                                      & \begin{tabular}[c]{@{}l@{}}PL Stalh et al \cite{staahl2016visualization}. \\ Data retrieved from \\ HEST-1k database.\end{tabular}        & \begin{tabular}[c]{@{}l@{}}H\&E-stained \\ histological sections\end{tabular}                                     & \begin{tabular}[c]{@{}l@{}}-Num. Spots: 1,029\\ -Num. Sections: 4\\ -Num. Patients: 1\\ -Num. Genes: 14851 \\ (Range: 14789 - 14929)\end{tabular}                                                                                                                          & \begin{tabular}[c]{@{}l@{}}Spatial \\ Transcriptomics\end{tabular}              & 100                                                                                         & 200                                                                                                  \\ \hline
CRC I                                                        & Colorectal cancer                                                                                                                                              & \begin{tabular}[c]{@{}l@{}}A. Valdeolivas et al. \cite{valdeolivas2024profiling}, \\ Data retrieved from \\ HEST-1k database.\end{tabular}  & \begin{tabular}[c]{@{}l@{}}Frozen H\&E-stained \\ histological sections\end{tabular}                              & \begin{tabular}[c]{@{}l@{}}-Num. Spots: 20,708\\ -Num. Sections: 14\\ -Num. Patients: 7\\ -Num. Genes: 36,601\end{tabular}                                                                                                                                                 & Visium                                                                          & 55                                                                                          & 100                                                                                                  \\ \hline
\begin{tabular}[c]{@{}l@{}}CRC II\end{tabular} & Healthy                                                                                                                                                        & \begin{tabular}[c]{@{}l@{}}R. Mirzazadeh et al \cite{mirzazadeh2023spatially}, \\ Data retrieved from \\ HEST-1k database.\end{tabular}   & \begin{tabular}[c]{@{}l@{}}Frozen H\&E-stained \\ histological sections\end{tabular}                              & \begin{tabular}[c]{@{}l@{}}-Num. Spots: 11,049\\ -Num. Sections: 5\\ -Num. Patients: N/A\\ -Num. Genes (mean): 21,674 \\ (Range: 17,943 - 36,601)\end{tabular}                                                                                                             & \begin{tabular}[c]{@{}l@{}}RNA-Rescue \\ Spatial\\ Transcriptomics\end{tabular} & 55                                                                                          & 100                                                                                                  \\ \hline
CRC III                                                        & \begin{tabular}[c]{@{}l@{}}-Colon adenocarcinoma (N=6) \\ -Colorectal\\  adenocarcinoma (N=1)\end{tabular}                                                     & \begin{tabular}[c]{@{}l@{}}10x Genomics. \\ Data retrieved from \\ HEST-1k database.\end{tabular}                 & \begin{tabular}[c]{@{}l@{}}Frozen (N=1) \& \\ FFPE (N=6) H\&E-stained \\ histological sections\end{tabular}       & \begin{tabular}[c]{@{}l@{}}-Num. Spots: 40,285\\ -Num. Sections: 7\\ -Num. Patients: 4\\ -Num. Genes (mean): 20,709 \\ (Range: 17,943 - 36,601)\end{tabular}                                                                                                               & Visium                                                                          & 55                                                                                          & 100                                                                                                  \\ \hline
\end{tabular}
\label{tab:dataset}
\end{table}


\clearpage

\begin{table}[h!]
    \centering
    \caption{Marker genes for different cancer cohorts.}
    \begin{tabular}{|p{3cm}|p{12cm}|} 
        \toprule
        \textbf{Dataset} & \textbf{Marker Genes} \\
        \midrule
        Prostate & \textbf{OncotypeDX}: AZGP1, KLK2, FAM13C, FLNC, GSN, TPM2, BGN, COL1A1, SFRP4 \\ 
                 & \textbf{Decipher}: CAMK2N1, G6PD, ATM, LASP1, PRDX4, PCNA, STMN1, ERG, NFIB, ANO7 \\
        \midrule
        Breast & \textbf{HER2DX}: RRM2, FGFR4, BAG1, PHGDH, MLPH, MYC \\
        & \textbf{Additional genes}: CDH1, ERBB3, ESR1, KMT2C, MAP2K4, MDM2, PTEN, ARID1A, ERBB2, FGFR1, GATA3, KMT2D, MAP3K1, TP53 \\
        \midrule
        Colorectal & APC, TP53, KRAS, SMAD4, BRAF, VCAN, ARID1A, SOX9, NRAS, KDR, FBXW7, MET, PTEN, BIRC6, ACVR2A, RNF43, UBR5, SETD1B, KMT2C, ZFP36L2, BMPR2, EFEMP2, FBN1, SPARC, SCD, RNF43, MMP1, PLAU, CXCL14, AXIN2 \\
        \bottomrule
    \end{tabular}
    \label{tab:marker_genes}
\end{table}

\clearpage
\begin{table}[h]
    \centering
    \caption{\textbf{Hyperparameters in $\ours$ pretraining (stage I).} CoCa training between H\&E and ST.}
    \scalebox{0.95}{
        \begin{tabular}{l|l}
            \toprule
            Hyperparameter & Value \\
            \midrule
            \midrule
            GPU & 2$\times$ 24GB GeForce RTX 3090 \\
            Batch size per GPU & 256 \\
            AdamW $\beta$ & (0.9, 0.999) \\
            Num. trainable blocks H\&E encoder & 3 \\
            Num. trainable blocks ST encoder & 3 \\
            H\&E encoder learning rate & 0.00001 \\
            ST encoder learning rate & 0.00001 \\
            ST predictor learning rate & 0.00001 \\
            Adversarial discriminator (batch correction) learning rate & 0.0001 \\ 
            H\&E encoder weight decay & 0.01 \\
            ST encoder weight decay & 0.01 \\
            ST predictor weight decay & 0.01 \\
            Adversarial discriminator (batch correction) weight decay & 0.01 \\
            Learning rate schedule & Linear (warmup period)-cosine \\
            Learning rate (start) & 0 \\
            Learning rate (post warmup) & 1e-5\\
            Learning rate (final) & 0 \\
            Warmup epochs & 5 \\
            Total epochs & 25 \\
            Contrastive Loss Temperature & 0.1 \\
            Automatic mixed precision & bfloat16 \\
            \bottomrule
            \end{tabular}}
\label{tab:stageI}
\end{table}

\clearpage
\begin{table}[h]
    \centering
    \caption{\textbf{Hyperparameters in $\ours$ training (stage II).} CoCa training between 3D tissue image, H\&E and ST.}
    \scalebox{0.95}{
        \begin{tabular}{l|l}
            \toprule
            Hyperparameter & Value \\
            \midrule
            \midrule
            GPU & 1$\times$ 24GB GeForce RTX 3090 \\
            Batch size per GPU & 128 \\
            AdamW $\beta$ & (0.9, 0.999) \\
            Num. trainable blocks H\&E encoder & 0 \\
            Num. trainable blocks volumetric image encoder & 0/3 (MicroCT/Serial Sections) \\
            Num. trainable blocks ST encoder & 0 \\
            Volumetric image encoder encoder learning rate & 0.00001 \\
            ST predictor learning rate & 0.00001 \\
            Volumetric image encoder weight decay & 0.01 \\
            ST predictor weight decay & 0.01 \\
            Learning rate schedule & Cosine \\
            Learning rate (start) & 0 \\
            Learning rate (post warmup) & 1e-5\\
            Learning rate (final) & 0 \\
            Epochs & 15 \\
            Contrastive Loss Temperature & 0.1 \\
            Automatic mixed precision & bfloat16 \\
            \bottomrule
            \end{tabular}
            }
\label{tab:stageII}
\end{table}

\clearpage
\begin{table}[h]
    \centering
    \caption{\textbf{Hyperparameters in $\ours$ fine-tuning (stage III).} CoCa training between volumetric image, H\&E and ST.}
    \scalebox{0.95}{
        \begin{tabular}{l|l}
            \toprule
            Hyperparameter & Value \\
            \midrule
            \midrule
            GPU & 1$\times$ 24GB GeForce RTX 3090 \\
            Batch size per GPU & 16 \\
            AdamW $\beta$ & (0.9, 0.999) \\
            Num. trainable blocks H\&E encoder & 3 \\
            Num. trainable blocks volumetric image encoder & 0/3 (MicroCT/Serial Sections) \\
            Num. trainable blocks ST encoder & 3 \\
            H\&E encoder learning rate & 0.00001 \\
            Volumetric image encoder encoder learning rate & 0.00001 \\
            ST encoder learning rate & 0.00001 \\
            ST predictor learning rate & 0.00001 \\
             H\&E encoder weight decay & 0.01 \\
             Volumetric image encoder weight decay & 0.01 \\
            ST encoder weight decay & 0.01 \\
            ST predictor weight decay & 0.01 \\
            Learning rate schedule & Cosine \\
            Learning rate (start) & 0 \\
            Learning rate (post warmup) & 1e-5\\
            Learning rate (final) & 0 \\
            Epochs & 10 \\
            Contrastive Loss Temperature & 0.1 \\
            Automatic mixed precision & bfloat16 \\
            \bottomrule
            \end{tabular}
            }
\label{tab:stageIII}
\end{table}

\clearpage

\begin{table}[h!]
    \centering
    \caption{\textit{Spatial filter genes} for molecular query design in the prostate cancer cohort.}
    \begin{tabular}{|p{4cm}|p{12cm}|} 
        \hline
        \textbf{Molecular Query} & \textbf{Gene Sets} \\
        \hline
        \multirow{2}{4cm}{Molecular Query 1: \textit{PI3K/AKT/mTOR} pathway} 
        & \textbf{\textit{Genes of interest}}: SPON2, TFF3 \\  
        & \textit{\textbf{Correlated genes}}: TSPAN1, SERP1, ERGIC1, SPDEF, STEAP2, NKX3-1, HSPA5, CORO1B, NCAPD3, TMBIM6, KRT18, CAMKK2, FOXA1, DHRS7, SERF2, ABHD2, ALDH1A3, APRT, CANT1, P4HB, DCXR, FASN, HMG20B, FXYD3, KLK3, KLK2, FAM3B, SMS \\  
        \hline
        \multirow{2}{4cm}{Molecular Query 2: \textit{Myogenesis}}  
        & \textbf{\textit{Genes of interest}}: TPM2, TAGLN \\  
        & \textit{\textbf{Correlated genes}}: LMOD1, EMILIN1, ACTG2, LIMS2, FN1, DES, COL6A3, MYLK, WDR1, SYNPO2, PALLD, PDLIM3, MAP1B, DPYSL3, SPARC, ACTB, CALD1, CLU, OGN, GSN, PTGDS, SVIL, ACTA2, SORBS1, ILK, ACTN1, TPM1, SYNM, TGFB1I1, KANK2, CNN1, HSPB6, DSTN, MYL9, JPH2, PCP4, SMTN, LGALS1, FLNA \\  
        \hline
        \multirow{2}{4cm}{Molecular Query 3: \textit{Tumor suppression}}  
        & \textbf{\textit{Genes of interest}}: MSMB, ACPP \\  
        & \textit{\textbf{Correlated genes}}: ADIRF, SCD, RDH11, NEFH \\  
        \hline
    \end{tabular}
    \label{tab:query_genes}
\end{table}


\clearpage

\begin{nolinenumbers}
\Heading{References}
\bibliographystyle{nature}
\bibliography{main}

\begin{thebibliography}{100}
\expandafter\ifx\csname url\endcsname\relax
  \def\url#1{\texttt{#1}}\fi
\expandafter\ifx\csname urlprefix\endcsname\relax\def\urlprefix{URL }\fi
\providecommand{\bibinfo}[2]{#2}
\providecommand{\eprint}[2][]{\url{#2}}

\bibitem{erturk2024deep}
\bibinfo{author}{Ert{\"u}rk, A.}
\newblock \bibinfo{title}{{Deep 3D histology powered by tissue clearing, omics and AI}}.
\newblock \emph{\bibinfo{journal}{Nature Methods}} \textbf{\bibinfo{volume}{21}}, \bibinfo{pages}{1153--1165} (\bibinfo{year}{2024}).

\bibitem{staahl2016visualization}
\bibinfo{author}{St{\aa}hl, P.~L.} \emph{et~al.}
\newblock \bibinfo{title}{Visualization and analysis of gene expression in tissue sections by spatial transcriptomics}.
\newblock \emph{\bibinfo{journal}{Science}} \textbf{\bibinfo{volume}{353}}, \bibinfo{pages}{78--82} (\bibinfo{year}{2016}).

\bibitem{marx2021method}
\bibinfo{author}{Marx, V.}
\newblock \bibinfo{title}{Method of the year: spatially resolved transcriptomics}.
\newblock \emph{\bibinfo{journal}{Nature methods}} \textbf{\bibinfo{volume}{18}}, \bibinfo{pages}{9--14} (\bibinfo{year}{2021}).

\bibitem{moses2022museum}
\bibinfo{author}{Moses, L.} \& \bibinfo{author}{Pachter, L.}
\newblock \bibinfo{title}{Museum of spatial transcriptomics}.
\newblock \emph{\bibinfo{journal}{Nature methods}} \textbf{\bibinfo{volume}{19}}, \bibinfo{pages}{534--546} (\bibinfo{year}{2022}).

\bibitem{wang2023construction}
\bibinfo{author}{Wang, G.} \emph{et~al.}
\newblock \bibinfo{title}{{Construction of a 3D whole organism spatial atlas by joint modelling of multiple slices with deep neural networks}}.
\newblock \emph{\bibinfo{journal}{Nature Machine Intelligence}} \textbf{\bibinfo{volume}{5}}, \bibinfo{pages}{1200--1213} (\bibinfo{year}{2023}).

\bibitem{tang2024search}
\bibinfo{author}{Tang, Z.} \emph{et~al.}
\newblock \bibinfo{title}{Search and match across spatial omics samples at single-cell resolution}.
\newblock \emph{\bibinfo{journal}{Nature methods}} \bibinfo{pages}{1--12} (\bibinfo{year}{2024}).

\bibitem{schott2024open}
\bibinfo{author}{Schott, M.} \emph{et~al.}
\newblock \bibinfo{title}{{Open-ST: High-resolution spatial transcriptomics in 3D}}.
\newblock \emph{\bibinfo{journal}{Cell}} \textbf{\bibinfo{volume}{187}}, \bibinfo{pages}{3953--3972} (\bibinfo{year}{2024}).

\bibitem{mo2024tumour}
\bibinfo{author}{Mo, C.-K.} \emph{et~al.}
\newblock \bibinfo{title}{{Tumour evolution and microenvironment interactions in 2D and 3D space}}.
\newblock \emph{\bibinfo{journal}{Nature}} \textbf{\bibinfo{volume}{634}}, \bibinfo{pages}{1178--1186} (\bibinfo{year}{2024}).

\bibitem{song2023artificial}
\bibinfo{author}{Song, A.~H.} \emph{et~al.}
\newblock \bibinfo{title}{Artificial intelligence for digital and computational pathology}.
\newblock \emph{\bibinfo{journal}{Nature Reviews Bioengineering}} \textbf{\bibinfo{volume}{1}}, \bibinfo{pages}{930--949} (\bibinfo{year}{2023}).

\bibitem{marusyk2020intratumor}
\bibinfo{author}{Marusyk, A.}, \bibinfo{author}{Janiszewska, M.} \& \bibinfo{author}{Polyak, K.}
\newblock \bibinfo{title}{Intratumor heterogeneity: the rosetta stone of therapy resistance}.
\newblock \emph{\bibinfo{journal}{Cancer cell}} \textbf{\bibinfo{volume}{37}}, \bibinfo{pages}{471--484} (\bibinfo{year}{2020}).

\bibitem{vitale2021intratumoral}
\bibinfo{author}{Vitale, I.}, \bibinfo{author}{Shema, E.}, \bibinfo{author}{Loi, S.} \& \bibinfo{author}{Galluzzi, L.}
\newblock \bibinfo{title}{Intratumoral heterogeneity in cancer progression and response to immunotherapy}.
\newblock \emph{\bibinfo{journal}{Nature medicine}} \textbf{\bibinfo{volume}{27}}, \bibinfo{pages}{212--224} (\bibinfo{year}{2021}).

\bibitem{fu2021spatial}
\bibinfo{author}{Fu, T.} \emph{et~al.}
\newblock \bibinfo{title}{Spatial architecture of the immune microenvironment orchestrates tumor immunity and therapeutic response}.
\newblock \emph{\bibinfo{journal}{Journal of hematology \& oncology}} \textbf{\bibinfo{volume}{14}}, \bibinfo{pages}{98} (\bibinfo{year}{2021}).

\bibitem{bagaev2021conserved}
\bibinfo{author}{Bagaev, A.} \emph{et~al.}
\newblock \bibinfo{title}{Conserved pan-cancer microenvironment subtypes predict response to immunotherapy}.
\newblock \emph{\bibinfo{journal}{Cancer cell}} \textbf{\bibinfo{volume}{39}}, \bibinfo{pages}{845--865} (\bibinfo{year}{2021}).

\bibitem{arora2023spatial}
\bibinfo{author}{Arora, R.} \emph{et~al.}
\newblock \bibinfo{title}{Spatial transcriptomics reveals distinct and conserved tumor core and edge architectures that predict survival and targeted therapy response}.
\newblock \emph{\bibinfo{journal}{Nature Communications}} \textbf{\bibinfo{volume}{14}}, \bibinfo{pages}{5029} (\bibinfo{year}{2023}).

\bibitem{rao2021exploring}
\bibinfo{author}{Rao, A.}, \bibinfo{author}{Barkley, D.}, \bibinfo{author}{Fran{\ifmmode\mbox{\c{c}}\else\c{c}\fi}a, G.~S.} \& \bibinfo{author}{Yanai, I.}
\newblock \bibinfo{title}{{Exploring tissue architecture using spatial transcriptomics}}.
\newblock \emph{\bibinfo{journal}{Nature}} \textbf{\bibinfo{volume}{596}}, \bibinfo{pages}{211--220} (\bibinfo{year}{2021}).

\bibitem{rodriques2019slide}
\bibinfo{author}{Rodriques, S.~G.} \emph{et~al.}
\newblock \bibinfo{title}{Slide-seq: A scalable technology for measuring genome-wide expression at high spatial resolution}.
\newblock \emph{\bibinfo{journal}{Science}} \textbf{\bibinfo{volume}{363}}, \bibinfo{pages}{1463--1467} (\bibinfo{year}{2019}).

\bibitem{palla2022spatial}
\bibinfo{author}{Palla, G.}, \bibinfo{author}{Fischer, D.~S.}, \bibinfo{author}{Regev, A.} \& \bibinfo{author}{Theis, F.~J.}
\newblock \bibinfo{title}{Spatial components of molecular tissue biology}.
\newblock \emph{\bibinfo{journal}{Nature Biotechnology}} \textbf{\bibinfo{volume}{40}}, \bibinfo{pages}{308--318} (\bibinfo{year}{2022}).

\bibitem{ren2024spatial}
\bibinfo{author}{Ren, J.}, \bibinfo{author}{Luo, S.}, \bibinfo{author}{Shi, H.} \& \bibinfo{author}{Wang, X.}
\newblock \bibinfo{title}{{Spatial omics advances for in situ RNA biology}}.
\newblock \emph{\bibinfo{journal}{Molecular Cell}} \textbf{\bibinfo{volume}{84}}, \bibinfo{pages}{3737--3757} (\bibinfo{year}{2024}).

\bibitem{liu2021harnessing}
\bibinfo{author}{Liu, J.~T.} \emph{et~al.}
\newblock \bibinfo{title}{{Harnessing non-destructive 3D pathology}}.
\newblock \emph{\bibinfo{journal}{Nature biomedical engineering}} \textbf{\bibinfo{volume}{5}}, \bibinfo{pages}{203--218} (\bibinfo{year}{2021}).

\bibitem{braxton20243d}
\bibinfo{author}{Braxton, A.~M.} \emph{et~al.}
\newblock \bibinfo{title}{{3D genomic mapping reveals multifocality of human pancreatic precancers}}.
\newblock \emph{\bibinfo{journal}{Nature}} \bibinfo{pages}{1--9} (\bibinfo{year}{2024}).

\bibitem{wang20243d}
\bibinfo{author}{Wang, L.}, \bibinfo{author}{Li, M.} \& \bibinfo{author}{Hwang, T.~H.}
\newblock \bibinfo{title}{The 3d revolution in cancer discovery}.
\newblock \emph{\bibinfo{journal}{Cancer discovery}} \textbf{\bibinfo{volume}{14}}, \bibinfo{pages}{625--629} (\bibinfo{year}{2024}).

\bibitem{mathur2024glioblastoma}
\bibinfo{author}{Mathur, R.} \emph{et~al.}
\newblock \bibinfo{title}{{Glioblastoma evolution and heterogeneity from a 3D whole-tumor perspective}}.
\newblock \emph{\bibinfo{journal}{Cell}} \textbf{\bibinfo{volume}{187}}, \bibinfo{pages}{446--463} (\bibinfo{year}{2024}).

\bibitem{withers2021x}
\bibinfo{author}{Withers, P.~J.} \emph{et~al.}
\newblock \bibinfo{title}{X-ray computed tomography}.
\newblock \emph{\bibinfo{journal}{Nature Reviews Methods Primers}} \textbf{\bibinfo{volume}{1}}, \bibinfo{pages}{18} (\bibinfo{year}{2021}).
\newblock \bibinfo{note}{\lowercase{https://doi.org/10.1038/s43586-021-00015-4}}.

\bibitem{bishop2024end}
\bibinfo{author}{Bishop, K.~W.} \emph{et~al.}
\newblock \bibinfo{title}{{An end-to-end workflow for nondestructive 3D pathology}}.
\newblock \emph{\bibinfo{journal}{Nature Protocols}} \bibinfo{pages}{1--27} (\bibinfo{year}{2024}).

\bibitem{song2024analysis}
\bibinfo{author}{Song, A.~H.} \emph{et~al.}
\newblock \bibinfo{title}{{Analysis of 3D pathology samples using weakly supervised AI}}.
\newblock \emph{\bibinfo{journal}{Cell}} \textbf{\bibinfo{volume}{187}}, \bibinfo{pages}{2502--2520} (\bibinfo{year}{2024}).

\bibitem{xie2022prostate}
\bibinfo{author}{Xie, W.} \emph{et~al.}
\newblock \bibinfo{title}{Prostate cancer risk stratification via nondestructive 3d pathology with deep learning--assisted gland analysis}.
\newblock \emph{\bibinfo{journal}{Cancer research}} \textbf{\bibinfo{volume}{82}}, \bibinfo{pages}{334--345} (\bibinfo{year}{2022}).

\bibitem{wang2018three}
\bibinfo{author}{Wang, X.} \emph{et~al.}
\newblock \bibinfo{title}{Three-dimensional intact-tissue sequencing of single-cell transcriptional states}.
\newblock \emph{\bibinfo{journal}{Science}} \textbf{\bibinfo{volume}{361}}, \bibinfo{pages}{eaat5691} (\bibinfo{year}{2018}).

\bibitem{wang2021easi}
\bibinfo{author}{Wang, Y.} \emph{et~al.}
\newblock \bibinfo{title}{Easi-fish for thick tissue defines lateral hypothalamus spatio-molecular organization}.
\newblock \emph{\bibinfo{journal}{Cell}} \textbf{\bibinfo{volume}{184}}, \bibinfo{pages}{6361--6377} (\bibinfo{year}{2021}).

\bibitem{fang2024three}
\bibinfo{author}{Fang, R.} \emph{et~al.}
\newblock \bibinfo{title}{Three-dimensional single-cell transcriptome imaging of thick tissues}.
\newblock \emph{\bibinfo{journal}{Elife}} \textbf{\bibinfo{volume}{12}}, \bibinfo{pages}{RP90029} (\bibinfo{year}{2024}).

\bibitem{sui2024scalable}
\bibinfo{author}{Sui, X.} \emph{et~al.}
\newblock \bibinfo{title}{Scalable spatial single-cell transcriptomics and translatomics in 3d thick tissue blocks}.
\newblock \emph{\bibinfo{journal}{bioRxiv}} \bibinfo{pages}{2024--08} (\bibinfo{year}{2024}).

\bibitem{doi:10.1126/science.adq2084}
\bibinfo{author}{Gandin, V.} \emph{et~al.}
\newblock \bibinfo{title}{Deep-tissue transcriptomics and subcellular imaging at high spatial resolution}.
\newblock \emph{\bibinfo{journal}{Science}} \textbf{\bibinfo{volume}{0}}, \bibinfo{pages}{eadq2084}.
\newblock \urlprefix\url{https://www.science.org/doi/abs/10.1126/science.adq2084}.
\newblock \eprint{https://www.science.org/doi/pdf/10.1126/science.adq2084}.

\bibitem{dong2022deciphering}
\bibinfo{author}{Dong, K.} \& \bibinfo{author}{Zhang, S.}
\newblock \bibinfo{title}{Deciphering spatial domains from spatially resolved transcriptomics with an adaptive graph attention auto-encoder}.
\newblock \emph{\bibinfo{journal}{Nature communications}} \textbf{\bibinfo{volume}{13}}, \bibinfo{pages}{1739} (\bibinfo{year}{2022}).

\bibitem{vickovic2022three}
\bibinfo{author}{Vickovic, S.} \emph{et~al.}
\newblock \bibinfo{title}{Three-dimensional spatial transcriptomics uncovers cell type localizations in the human rheumatoid arthritis synovium}.
\newblock \emph{\bibinfo{journal}{Communications Biology}} \textbf{\bibinfo{volume}{5}}, \bibinfo{pages}{129} (\bibinfo{year}{2022}).

\bibitem{zeira2022alignment}
\bibinfo{author}{Zeira, R.}, \bibinfo{author}{Land, M.}, \bibinfo{author}{Strzalkowski, A.} \& \bibinfo{author}{Raphael, B.~J.}
\newblock \bibinfo{title}{Alignment and integration of spatial transcriptomics data}.
\newblock \emph{\bibinfo{journal}{Nature Methods}} \textbf{\bibinfo{volume}{19}}, \bibinfo{pages}{567--575} (\bibinfo{year}{2022}).

\bibitem{zhou2023integrating}
\bibinfo{author}{Zhou, X.}, \bibinfo{author}{Dong, K.} \& \bibinfo{author}{Zhang, S.}
\newblock \bibinfo{title}{Integrating spatial transcriptomics data across different conditions, technologies and developmental stages}.
\newblock \emph{\bibinfo{journal}{Nature Computational Science}} \textbf{\bibinfo{volume}{3}}, \bibinfo{pages}{894--906} (\bibinfo{year}{2023}).

\bibitem{lin2023multiplexed}
\bibinfo{author}{Lin, J.-R.} \emph{et~al.}
\newblock \bibinfo{title}{{Multiplexed 3D atlas of state transitions and immune interaction in colorectal cancer}}.
\newblock \emph{\bibinfo{journal}{Cell}} \textbf{\bibinfo{volume}{186}}, \bibinfo{pages}{363--381} (\bibinfo{year}{2023}).

\bibitem{shu2024efficient}
\bibinfo{author}{Shu, H.} \emph{et~al.}
\newblock \bibinfo{title}{Efficient integration of multiple spatial transcriptomics data for 3d domain detection, matching, and alignment with stmsa}.
\newblock \emph{\bibinfo{journal}{bioRxiv}} \bibinfo{pages}{2024--07} (\bibinfo{year}{2024}).

\bibitem{edsgard2018identification}
\bibinfo{author}{Edsg{\"a}rd, D.}, \bibinfo{author}{Johnsson, P.} \& \bibinfo{author}{Sandberg, R.}
\newblock \bibinfo{title}{Identification of spatial expression trends in single-cell gene expression data}.
\newblock \emph{\bibinfo{journal}{Nature methods}} \textbf{\bibinfo{volume}{15}}, \bibinfo{pages}{339--342} (\bibinfo{year}{2018}).

\bibitem{svensson2018spatialde}
\bibinfo{author}{Svensson, V.}, \bibinfo{author}{Teichmann, S.~A.} \& \bibinfo{author}{Stegle, O.}
\newblock \bibinfo{title}{Spatialde: identification of spatially variable genes}.
\newblock \emph{\bibinfo{journal}{Nature methods}} \textbf{\bibinfo{volume}{15}}, \bibinfo{pages}{343--346} (\bibinfo{year}{2018}).

\bibitem{sun2020statistical}
\bibinfo{author}{Sun, S.}, \bibinfo{author}{Zhu, J.} \& \bibinfo{author}{Zhou, X.}
\newblock \bibinfo{title}{Statistical analysis of spatial expression patterns for spatially resolved transcriptomic studies}.
\newblock \emph{\bibinfo{journal}{Nature methods}} \textbf{\bibinfo{volume}{17}}, \bibinfo{pages}{193--200} (\bibinfo{year}{2020}).

\bibitem{binder2021morphological}
\bibinfo{author}{Binder, A.} \emph{et~al.}
\newblock \bibinfo{title}{Morphological and molecular breast cancer profiling through explainable machine learning}.
\newblock \emph{\bibinfo{journal}{Nature Machine Intelligence}} \textbf{\bibinfo{volume}{3}}, \bibinfo{pages}{355--366} (\bibinfo{year}{2021}).

\bibitem{ash2021joint}
\bibinfo{author}{Ash, J.~T.}, \bibinfo{author}{Darnell, G.}, \bibinfo{author}{Munro, D.} \& \bibinfo{author}{Engelhardt, B.~E.}
\newblock \bibinfo{title}{Joint analysis of expression levels and histological images identifies genes associated with tissue morphology}.
\newblock \emph{\bibinfo{journal}{Nature communications}} \textbf{\bibinfo{volume}{12}}, \bibinfo{pages}{1609} (\bibinfo{year}{2021}).

\bibitem{hu2021spagcn}
\bibinfo{author}{Hu, J.} \emph{et~al.}
\newblock \bibinfo{title}{{SpaGCN: Integrating gene expression, spatial location and histology to identify spatial domains and spatially variable genes by graph convolutional network}}.
\newblock \emph{\bibinfo{journal}{Nature methods}} \textbf{\bibinfo{volume}{18}}, \bibinfo{pages}{1342--1351} (\bibinfo{year}{2021}).

\bibitem{jaume2024hest}
\bibinfo{author}{Jaume, G.} \emph{et~al.}
\newblock \bibinfo{title}{{HEST-1k}: A dataset for spatial transcriptomics and histology image analysis}.
\newblock In \emph{\bibinfo{booktitle}{The Thirty-eight Conference on Neural Information Processing Systems Datasets and Benchmarks Track}} (\bibinfo{year}{2024}).

\bibitem{chen2024stimagekm}
\bibinfo{author}{Chen, J.} \emph{et~al.}
\newblock \bibinfo{title}{{STimage-1K4M: A histopathology image-gene expression dataset for spatial transcriptomics}}.
\newblock In \emph{\bibinfo{booktitle}{The Thirty-eight Conference on Neural Information Processing Systems Datasets and Benchmarks Track}} (\bibinfo{year}{2024}).

\bibitem{he2020integrating}
\bibinfo{author}{He, B.} \emph{et~al.}
\newblock \bibinfo{title}{Integrating spatial gene expression and breast tumour morphology via deep learning}.
\newblock \emph{\bibinfo{journal}{Nature biomedical engineering}} \textbf{\bibinfo{volume}{4}}, \bibinfo{pages}{827--834} (\bibinfo{year}{2020}).

\bibitem{bergenstraahle2022super}
\bibinfo{author}{Bergenstr{\aa}hle, L.} \emph{et~al.}
\newblock \bibinfo{title}{Super-resolved spatial transcriptomics by deep data fusion}.
\newblock \emph{\bibinfo{journal}{Nature biotechnology}} \textbf{\bibinfo{volume}{40}}, \bibinfo{pages}{476--479} (\bibinfo{year}{2022}).

\bibitem{xie2024spatially}
\bibinfo{author}{Xie, R.} \emph{et~al.}
\newblock \bibinfo{title}{Spatially resolved gene expression prediction from histology images via bi-modal contrastive learning}.
\newblock \emph{\bibinfo{journal}{Advances in Neural Information Processing Systems}} \textbf{\bibinfo{volume}{36}} (\bibinfo{year}{2024}).

\bibitem{chung2024accurate}
\bibinfo{author}{Chung, Y.}, \bibinfo{author}{Ha, J.~H.}, \bibinfo{author}{Im, K.~C.} \& \bibinfo{author}{Lee, J.~S.}
\newblock \bibinfo{title}{Accurate spatial gene expression prediction by integrating multi-resolution features}.
\newblock In \emph{\bibinfo{booktitle}{Proceedings of the IEEE/CVF Conference on Computer Vision and Pattern Recognition}}, \bibinfo{pages}{11591--11600} (\bibinfo{year}{2024}).

\bibitem{coleman2024unlocking}
\bibinfo{author}{Coleman, K.}, \bibinfo{author}{Schroeder, A.} \& \bibinfo{author}{Li, M.}
\newblock \bibinfo{title}{Unlocking the power of spatial omics with ai}.
\newblock \emph{\bibinfo{journal}{Nature Methods}} \textbf{\bibinfo{volume}{21}}, \bibinfo{pages}{1378--1381} (\bibinfo{year}{2024}).

\bibitem{kueckelhaus2024inferring}
\bibinfo{author}{Kueckelhaus, J.} \emph{et~al.}
\newblock \bibinfo{title}{Inferring histology-associated gene expression gradients in spatial transcriptomic studies}.
\newblock \emph{\bibinfo{journal}{Nature Communications}} \textbf{\bibinfo{volume}{15}}, \bibinfo{pages}{7280} (\bibinfo{year}{2024}).

\bibitem{zhang2024inferring}
\bibinfo{author}{Zhang, D.} \emph{et~al.}
\newblock \bibinfo{title}{Inferring super-resolution tissue architecture by integrating spatial transcriptomics with histology}.
\newblock \emph{\bibinfo{journal}{Nature biotechnology}} \bibinfo{pages}{1--6} (\bibinfo{year}{2024}).

\bibitem{lee2024Path}
\bibinfo{author}{Lee, Y.}, \bibinfo{author}{Liu, X.}, \bibinfo{author}{Hao, M.}, \bibinfo{author}{Liu, T.} \& \bibinfo{author}{Regev, A.}
\newblock \bibinfo{title}{{PathOmCLIP: Connecting tumor histology with spatial gene expression via locally enhanced contrastive learning of Pathology and Single-cell foundation model}}.
\newblock \emph{\bibinfo{journal}{bioRxiv}}  (\bibinfo{year}{2024}).

\bibitem{palermo2025investigating}
\bibinfo{author}{Palermo, F.} \emph{et~al.}
\newblock \bibinfo{title}{Investigating gut alterations in alzheimer’s disease: In-depth analysis with micro- and nano-3d x-ray phase contrast tomography}.
\newblock \emph{\bibinfo{journal}{Science Advances}} \textbf{\bibinfo{volume}{11}}, \bibinfo{pages}{eadr8511} (\bibinfo{year}{2025}).

\bibitem{glaser2017light}
\bibinfo{author}{Glaser, A.~K.} \emph{et~al.}
\newblock \bibinfo{title}{Light-sheet microscopy for slide-free non-destructive pathology of large clinical specimens}.
\newblock \emph{\bibinfo{journal}{Nature biomedical engineering}} \textbf{\bibinfo{volume}{1}}, \bibinfo{pages}{1--10} (\bibinfo{year}{2017}).

\bibitem{kiemen2022coda}
\bibinfo{author}{Kiemen, A.~L.} \emph{et~al.}
\newblock \bibinfo{title}{{CODA}: quantitative 3{D} reconstruction of large tissues at cellular resolution}.
\newblock \emph{\bibinfo{journal}{Nature Methods}} \bibinfo{pages}{1--10} (\bibinfo{year}{2022}).

\bibitem{li2023feasibility}
\bibinfo{author}{Li, K. Y.~C.} \emph{et~al.}
\newblock \bibinfo{title}{{Feasibility and safety of synchrotron-based X-ray phase contrast imaging as a technique complementary to histopathology analysis}}.
\newblock \emph{\bibinfo{journal}{Histochemistry and cell biology}} \textbf{\bibinfo{volume}{160}}, \bibinfo{pages}{377--389} (\bibinfo{year}{2023}).

\bibitem{stuart2019comprehensive}
\bibinfo{author}{Stuart, T.} \emph{et~al.}
\newblock \bibinfo{title}{Comprehensive integration of single-cell data}.
\newblock \emph{\bibinfo{journal}{cell}} \textbf{\bibinfo{volume}{177}}, \bibinfo{pages}{1888--1902} (\bibinfo{year}{2019}).

\bibitem{korsunsky2019fast}
\bibinfo{author}{Korsunsky, I.} \emph{et~al.}
\newblock \bibinfo{title}{Fast, sensitive and accurate integration of single-cell data with harmony}.
\newblock \emph{\bibinfo{journal}{Nature methods}} \textbf{\bibinfo{volume}{16}}, \bibinfo{pages}{1289--1296} (\bibinfo{year}{2019}).

\bibitem{kirillov2023segment}
\bibinfo{author}{Kirillov, A.} \emph{et~al.}
\newblock \bibinfo{title}{Segment anything}.
\newblock In \emph{\bibinfo{booktitle}{Proceedings of the IEEE/CVF International Conference on Computer Vision}}, \bibinfo{pages}{4015--4026} (\bibinfo{year}{2023}).

\bibitem{lu2024visual}
\bibinfo{author}{Lu, M.~Y.} \emph{et~al.}
\newblock \bibinfo{title}{A visual-language foundation model for computational pathology}.
\newblock \emph{\bibinfo{journal}{Nature Medicine}} \textbf{\bibinfo{volume}{30}}, \bibinfo{pages}{863--874} (\bibinfo{year}{2024}).

\bibitem{cui2024scgpt}
\bibinfo{author}{Cui, H.} \emph{et~al.}
\newblock \bibinfo{title}{{scGPT: toward building a foundation model for single-cell multi-omics using generative AI}}.
\newblock \emph{\bibinfo{journal}{Nature Methods}} \bibinfo{pages}{1--11} (\bibinfo{year}{2024}).

\bibitem{yu2022coca}
\bibinfo{author}{Yu, J.} \emph{et~al.}
\newblock \bibinfo{title}{{CoCa: Contrastive Captioners are Image-Text Foundation Models}}.
\newblock \emph{\bibinfo{journal}{Transactions on Machine Learning Research}}  (\bibinfo{year}{2022}).

\bibitem{li2024high}
\bibinfo{author}{Li, S.}, \bibinfo{author}{Gai, K.}, \bibinfo{author}{Dong, K.}, \bibinfo{author}{Zhang, Y.} \& \bibinfo{author}{Zhang, S.}
\newblock \bibinfo{title}{{High-density generation of spatial transcriptomics with STAGE}}.
\newblock \emph{\bibinfo{journal}{Nucleic Acids Research}} \textbf{\bibinfo{volume}{52}}, \bibinfo{pages}{4843--4856} (\bibinfo{year}{2024}).

\bibitem{lin2024bridging}
\bibinfo{author}{Lin, S.} \emph{et~al.}
\newblock \bibinfo{title}{{Bridging the Dimensional Gap from Planar Spatial Transcriptomics to 3D Cell Atlases}}.
\newblock \emph{\bibinfo{journal}{bioRxiv}}  (\bibinfo{year}{2024}).

\bibitem{bay2006surf}
\bibinfo{author}{Bay, H.}, \bibinfo{author}{Tuytelaars, T.} \& \bibinfo{author}{Van~Gool, L.}
\newblock \bibinfo{title}{Surf: Speeded up robust features}.
\newblock In \emph{\bibinfo{booktitle}{European Conference on Computer Vision}}, \bibinfo{pages}{404--417} (\bibinfo{organization}{Springer}, \bibinfo{year}{2006}).

\bibitem{fischler1981random}
\bibinfo{author}{Fischler, M.~A.} \& \bibinfo{author}{Bolles, R.~C.}
\newblock \bibinfo{title}{Random sample consensus: a paradigm for model fitting with applications to image analysis and automated cartography}.
\newblock \emph{\bibinfo{journal}{Communications of the ACM}} \textbf{\bibinfo{volume}{24}}, \bibinfo{pages}{381--395} (\bibinfo{year}{1981}).

\bibitem{gatenbee2023virtual}
\bibinfo{author}{Gatenbee, C.~D.} \emph{et~al.}
\newblock \bibinfo{title}{Virtual alignment of pathology image series for multi-gigapixel whole slide images}.
\newblock \emph{\bibinfo{journal}{Nature communications}} \textbf{\bibinfo{volume}{14}}, \bibinfo{pages}{4502} (\bibinfo{year}{2023}).

\bibitem{chicherova2014histology}
\bibinfo{author}{Chicherova, N.}, \bibinfo{author}{Fundana, K.}, \bibinfo{author}{M{\"u}ller, B.} \& \bibinfo{author}{Cattin, P.~C.}
\newblock \bibinfo{title}{{Histology to $\mu$ CT data matching using landmarks and a density biased RANSAC}}.
\newblock In \emph{\bibinfo{booktitle}{Medical Image Computing and Computer-Assisted Intervention}}, \bibinfo{pages}{243--250} (\bibinfo{organization}{Springer}, \bibinfo{year}{2014}).

\bibitem{wang2004image}
\bibinfo{author}{Wang, Z.}, \bibinfo{author}{Bovik, A.~C.}, \bibinfo{author}{Sheikh, H.~R.} \& \bibinfo{author}{Simoncelli, E.~P.}
\newblock \bibinfo{title}{Image quality assessment: from error visibility to structural similarity}.
\newblock \emph{\bibinfo{journal}{IEEE transactions on image processing}} \textbf{\bibinfo{volume}{13}}, \bibinfo{pages}{600--612} (\bibinfo{year}{2004}).

\bibitem{wang2025benchmark}
\bibinfo{author}{Wang, C.} \emph{et~al.}
\newblock \bibinfo{title}{Benchmarking the translational potential of spatial gene expression prediction from histology}.
\newblock \emph{\bibinfo{journal}{Nature Communications}} \textbf{\bibinfo{volume}{16}}, \bibinfo{pages}{1544} (\bibinfo{year}{2025}).

\bibitem{klein201417}
\bibinfo{author}{Klein, E.~A.} \emph{et~al.}
\newblock \bibinfo{title}{A 17-gene assay to predict prostate cancer aggressiveness in the context of gleason grade heterogeneity, tumor multifocality, and biopsy undersampling}.
\newblock \emph{\bibinfo{journal}{European urology}} \textbf{\bibinfo{volume}{66}}, \bibinfo{pages}{550--560} (\bibinfo{year}{2014}).

\bibitem{cullen2015biopsy}
\bibinfo{author}{Cullen, J.} \emph{et~al.}
\newblock \bibinfo{title}{A biopsy-based 17-gene genomic prostate score predicts recurrence after radical prostatectomy and adverse surgical pathology in a racially diverse population of men with clinically low-and intermediate-risk prostate cancer}.
\newblock \emph{\bibinfo{journal}{European urology}} \textbf{\bibinfo{volume}{68}}, \bibinfo{pages}{123--131} (\bibinfo{year}{2015}).

\bibitem{klein2016decipher}
\bibinfo{author}{Klein, E.~A.} \emph{et~al.}
\newblock \bibinfo{title}{Decipher genomic classifier measured on prostate biopsy predicts metastasis risk}.
\newblock \emph{\bibinfo{journal}{Urology}} \textbf{\bibinfo{volume}{90}}, \bibinfo{pages}{148--152} (\bibinfo{year}{2016}).

\bibitem{moran1950notes}
\bibinfo{author}{Moran, P.~A.}
\newblock \bibinfo{title}{Notes on continuous stochastic phenomena}.
\newblock \emph{\bibinfo{journal}{Biometrika}} \textbf{\bibinfo{volume}{37}}, \bibinfo{pages}{17--23} (\bibinfo{year}{1950}).

\bibitem{geary1954contiguity}
\bibinfo{author}{Geary, R.~C.}
\newblock \bibinfo{title}{The contiguity ratio and statistical mapping}.
\newblock \emph{\bibinfo{journal}{The incorporated statistician}} \textbf{\bibinfo{volume}{5}}, \bibinfo{pages}{115--146} (\bibinfo{year}{1954}).

\bibitem{hu2023deciphering}
\bibinfo{author}{Hu, J.} \emph{et~al.}
\newblock \bibinfo{title}{Deciphering tumor ecosystems at super resolution from spatial transcriptomics with tesla}.
\newblock \emph{\bibinfo{journal}{Cell systems}} \textbf{\bibinfo{volume}{14}}, \bibinfo{pages}{404--417} (\bibinfo{year}{2023}).

\bibitem{ravi2024sam2segmentimages}
\bibinfo{author}{Ravi, N.} \emph{et~al.}
\newblock \bibinfo{title}{{SAM 2: Segment Anything in Images and Videos}} (\bibinfo{year}{2024}).
\newblock \urlprefix\url{https://arxiv.org/abs/2408.00714}.
\newblock \eprint{2408.00714}.

\bibitem{whitaker2010rs10993994}
\bibinfo{author}{Whitaker, H.~C.} \emph{et~al.}
\newblock \bibinfo{title}{The rs10993994 risk allele for prostate cancer results in clinically relevant changes in microseminoprotein-beta expression in tissue and urine}.
\newblock \emph{\bibinfo{journal}{PloS one}} \textbf{\bibinfo{volume}{5}}, \bibinfo{pages}{e13363} (\bibinfo{year}{2010}).

\bibitem{qian2012spondin}
\bibinfo{author}{Qian, X.} \emph{et~al.}
\newblock \bibinfo{title}{{Spondin-2 (SPON2), a more prostate-cancer-specific diagnostic biomarker}}.
\newblock \emph{\bibinfo{journal}{PloS one}} \textbf{\bibinfo{volume}{7}}, \bibinfo{pages}{e37225} (\bibinfo{year}{2012}).

\bibitem{berglund2018spatial}
\bibinfo{author}{Berglund, E.} \emph{et~al.}
\newblock \bibinfo{title}{Spatial maps of prostate cancer transcriptomes reveal an unexplored landscape of heterogeneity}.
\newblock \emph{\bibinfo{journal}{Nature communications}} \textbf{\bibinfo{volume}{9}}, \bibinfo{pages}{2419} (\bibinfo{year}{2018}).

\bibitem{lapointe2004gene}
\bibinfo{author}{Lapointe, J.} \emph{et~al.}
\newblock \bibinfo{title}{Gene expression profiling identifies clinically relevant subtypes of prostate cancer}.
\newblock \emph{\bibinfo{journal}{Proceedings of the National Academy of Sciences}} \textbf{\bibinfo{volume}{101}}, \bibinfo{pages}{811--816} (\bibinfo{year}{2004}).

\bibitem{burdelski2016reduced}
\bibinfo{author}{Burdelski, C.} \emph{et~al.}
\newblock \bibinfo{title}{{Reduced AZGP1 expression is an independent predictor of early PSA recurrence and associated with ERG-fusion positive and PTEN deleted prostate cancers}}.
\newblock \emph{\bibinfo{journal}{International journal of cancer}} \textbf{\bibinfo{volume}{138}}, \bibinfo{pages}{1199--1206} (\bibinfo{year}{2016}).

\bibitem{kristensen2019predictive}
\bibinfo{author}{Kristensen, G.} \emph{et~al.}
\newblock \bibinfo{title}{{Predictive value of AZGP1 following radical prostatectomy for prostate cancer: a cohort study and meta-analysis}}.
\newblock \emph{\bibinfo{journal}{Journal of Clinical Pathology}} \textbf{\bibinfo{volume}{72}}, \bibinfo{pages}{696--704} (\bibinfo{year}{2019}).

\bibitem{andersson2021spatial}
\bibinfo{author}{Andersson, A.} \emph{et~al.}
\newblock \bibinfo{title}{{Spatial deconvolution of HER2-positive breast cancer delineates tumor-associated cell type interactions}}.
\newblock \emph{\bibinfo{journal}{Nature communications}} \textbf{\bibinfo{volume}{12}}, \bibinfo{pages}{6012} (\bibinfo{year}{2021}).

\bibitem{valdeolivas2024profiling}
\bibinfo{author}{Valdeolivas, A.} \emph{et~al.}
\newblock \bibinfo{title}{Profiling the heterogeneity of colorectal cancer consensus molecular subtypes using spatial transcriptomics}.
\newblock \emph{\bibinfo{journal}{NPJ precision oncology}} \textbf{\bibinfo{volume}{8}}, \bibinfo{pages}{10} (\bibinfo{year}{2024}).

\bibitem{mirzazadeh2023spatially}
\bibinfo{author}{Mirzazadeh, R.} \emph{et~al.}
\newblock \bibinfo{title}{Spatially resolved transcriptomic profiling of degraded and challenging fresh frozen samples}.
\newblock \emph{\bibinfo{journal}{Nature Communications}} \textbf{\bibinfo{volume}{14}}, \bibinfo{pages}{509} (\bibinfo{year}{2023}).

\bibitem{prat2020multivariable}
\bibinfo{author}{Prat, A.} \emph{et~al.}
\newblock \bibinfo{title}{{A multivariable prognostic score to guide systemic therapy in early-stage HER2-positive breast cancer: a retrospective study with an external evaluation}}.
\newblock \emph{\bibinfo{journal}{The Lancet Oncology}} \textbf{\bibinfo{volume}{21}}, \bibinfo{pages}{1455--1464} (\bibinfo{year}{2020}).

\bibitem{smith2021her2+}
\bibinfo{author}{Smith, A.~E.} \emph{et~al.}
\newblock \bibinfo{title}{{HER2+ breast cancers evade anti-HER2 therapy via a switch in driver pathway}}.
\newblock \emph{\bibinfo{journal}{Nature Communications}} \textbf{\bibinfo{volume}{12}}, \bibinfo{pages}{6667} (\bibinfo{year}{2021}).

\bibitem{kim2017epithelial}
\bibinfo{author}{Kim, G.-E.}, \bibinfo{author}{Lee, J.~S.}, \bibinfo{author}{Park, M.~H.} \& \bibinfo{author}{Yoon, J.~H.}
\newblock \bibinfo{title}{Epithelial periostin expression is correlated with poor survival in patients with invasive breast carcinoma}.
\newblock \emph{\bibinfo{journal}{PLoS One}} \textbf{\bibinfo{volume}{12}}, \bibinfo{pages}{e0187635} (\bibinfo{year}{2017}).

\bibitem{xiao2024integrating}
\bibinfo{author}{Xiao, J.} \emph{et~al.}
\newblock \bibinfo{title}{Integrating spatial and single-cell transcriptomics reveals tumor heterogeneity and intercellular networks in colorectal cancer}.
\newblock \emph{\bibinfo{journal}{Cell Death \& Disease}} \textbf{\bibinfo{volume}{15}}, \bibinfo{pages}{326} (\bibinfo{year}{2024}).

\bibitem{spizzo2011epcam}
\bibinfo{author}{Spizzo, G.} \emph{et~al.}
\newblock \bibinfo{title}{Epcam expression in primary tumour tissues and metastases: an immunohistochemical analysis}.
\newblock \emph{\bibinfo{journal}{Journal of clinical pathology}} \textbf{\bibinfo{volume}{64}}, \bibinfo{pages}{415--420} (\bibinfo{year}{2011}).

\bibitem{xu2020identification}
\bibinfo{author}{Xu, H.} \emph{et~al.}
\newblock \bibinfo{title}{Identification and verification of core genes in colorectal cancer}.
\newblock \emph{\bibinfo{journal}{BioMed Research International}} \textbf{\bibinfo{volume}{2020}}, \bibinfo{pages}{8082697} (\bibinfo{year}{2020}).

\bibitem{liu2018overexpression}
\bibinfo{author}{Liu, J.} \emph{et~al.}
\newblock \bibinfo{title}{Overexpression of tff3 is involved in prostate carcinogenesis via blocking mitochondria-mediated apoptosis}.
\newblock \emph{\bibinfo{journal}{Experimental \& molecular medicine}} \textbf{\bibinfo{volume}{50}}, \bibinfo{pages}{1--11} (\bibinfo{year}{2018}).

\bibitem{zhang2024biological}
\bibinfo{author}{Zhang, J.} \emph{et~al.}
\newblock \bibinfo{title}{The biological functions and related signaling pathways of spon2}.
\newblock \emph{\bibinfo{journal}{Frontiers in Oncology}} \textbf{\bibinfo{volume}{13}}, \bibinfo{pages}{1323744} (\bibinfo{year}{2024}).

\bibitem{liberzon2015molecular}
\bibinfo{author}{Liberzon, A.} \emph{et~al.}
\newblock \bibinfo{title}{The molecular signatures database hallmark gene set collection}.
\newblock \emph{\bibinfo{journal}{Cell systems}} \textbf{\bibinfo{volume}{1}}, \bibinfo{pages}{417--425} (\bibinfo{year}{2015}).

\bibitem{ma2023prostate}
\bibinfo{author}{Ma, C.} \emph{et~al.}
\newblock \bibinfo{title}{The prostate stromal transcriptome in aggressive and lethal prostate cancer}.
\newblock \emph{\bibinfo{journal}{Molecular Cancer Research}} \textbf{\bibinfo{volume}{21}}, \bibinfo{pages}{253--260} (\bibinfo{year}{2023}).

\bibitem{sjoblom2016microseminoprotein}
\bibinfo{author}{Sj{\"o}blom, L.} \emph{et~al.}
\newblock \bibinfo{title}{Microseminoprotein-beta expression in different stages of prostate cancer}.
\newblock \emph{\bibinfo{journal}{PloS one}} \textbf{\bibinfo{volume}{11}}, \bibinfo{pages}{e0150241} (\bibinfo{year}{2016}).

\bibitem{veeramani2005cellular}
\bibinfo{author}{Veeramani, S.} \emph{et~al.}
\newblock \bibinfo{title}{Cellular prostatic acid phosphatase: a protein tyrosine phosphatase involved in androgen-independent proliferation of prostate cancer}.
\newblock \emph{\bibinfo{journal}{Endocrine-Related Cancer}} \textbf{\bibinfo{volume}{12}}, \bibinfo{pages}{805--822} (\bibinfo{year}{2005}).

\bibitem{shorning2020pi3k}
\bibinfo{author}{Shorning, B.~Y.}, \bibinfo{author}{Dass, M.~S.}, \bibinfo{author}{Smalley, M.~J.} \& \bibinfo{author}{Pearson, H.~B.}
\newblock \bibinfo{title}{The pi3k-akt-mtor pathway and prostate cancer: at the crossroads of ar, mapk, and wnt signaling}.
\newblock \emph{\bibinfo{journal}{International journal of molecular sciences}} \textbf{\bibinfo{volume}{21}}, \bibinfo{pages}{4507} (\bibinfo{year}{2020}).

\bibitem{dakhova2014genes}
\bibinfo{author}{Dakhova, O.}, \bibinfo{author}{Rowley, D.} \& \bibinfo{author}{Ittmann, M.}
\newblock \bibinfo{title}{Genes upregulated in prostate cancer reactive stroma promote prostate cancer progression in vivo}.
\newblock \emph{\bibinfo{journal}{Clinical Cancer Research}} \textbf{\bibinfo{volume}{20}}, \bibinfo{pages}{100--109} (\bibinfo{year}{2014}).

\bibitem{bjartell2007association}
\bibinfo{author}{Bjartell, A.~S.} \emph{et~al.}
\newblock \bibinfo{title}{Association of cysteine-rich secretory protein 3 and $\beta$-microseminoprotein with outcome after radical prostatectomy}.
\newblock \emph{\bibinfo{journal}{Clinical cancer research}} \textbf{\bibinfo{volume}{13}}, \bibinfo{pages}{4130--4138} (\bibinfo{year}{2007}).

\bibitem{dahlman2011evaluation}
\bibinfo{author}{Dahlman, A.} \emph{et~al.}
\newblock \bibinfo{title}{Evaluation of the prognostic significance of msmb and crisp3 in prostate cancer using automated image analysis}.
\newblock \emph{\bibinfo{journal}{Modern pathology}} \textbf{\bibinfo{volume}{24}}, \bibinfo{pages}{708--719} (\bibinfo{year}{2011}).

\bibitem{chen2024towards}
\bibinfo{author}{Chen, R.~J.} \emph{et~al.}
\newblock \bibinfo{title}{Towards a general-purpose foundation model for computational pathology}.
\newblock \emph{\bibinfo{journal}{Nature Medicine}} \textbf{\bibinfo{volume}{30}}, \bibinfo{pages}{850--862} (\bibinfo{year}{2024}).

\bibitem{zimmermann2024virchow}
\bibinfo{author}{Zimmermann, E.} \emph{et~al.}
\newblock \bibinfo{title}{Virchow 2: Scaling self-supervised mixed magnification models in pathology}.
\newblock \emph{\bibinfo{journal}{arXiv preprint arXiv:2408.00738}}  (\bibinfo{year}{2024}).

\bibitem{campanella2024clinical}
\bibinfo{author}{Campanella, G.} \emph{et~al.}
\newblock \bibinfo{title}{A clinical benchmark of public self-supervised pathology foundation models}.
\newblock \emph{\bibinfo{journal}{arXiv preprint arXiv:2407.06508}}  (\bibinfo{year}{2024}).

\bibitem{hao2024large}
\bibinfo{author}{Hao, M.} \emph{et~al.}
\newblock \bibinfo{title}{Large-scale foundation model on single-cell transcriptomics}.
\newblock \emph{\bibinfo{journal}{Nature Methods}} \bibinfo{pages}{1--11} (\bibinfo{year}{2024}).

\bibitem{frohn20203d}
\bibinfo{author}{Frohn, J.} \emph{et~al.}
\newblock \bibinfo{title}{3d virtual histology of human pancreatic tissue by multiscale phase-contrast x-ray tomography}.
\newblock \emph{\bibinfo{journal}{Journal of Synchrotron Radiation}} \textbf{\bibinfo{volume}{27}}, \bibinfo{pages}{1707--1719} (\bibinfo{year}{2020}).

\bibitem{walsh2021imaging}
\bibinfo{author}{Walsh, C.} \emph{et~al.}
\newblock \bibinfo{title}{Imaging intact human organs with local resolution of cellular structures using hierarchical phase-contrast tomography}.
\newblock \emph{\bibinfo{journal}{Nature methods}} \textbf{\bibinfo{volume}{18}}, \bibinfo{pages}{1532--1541} (\bibinfo{year}{2021}).

\bibitem{kim2024holotomography}
\bibinfo{author}{Kim, G.} \emph{et~al.}
\newblock \bibinfo{title}{Holotomography}.
\newblock \emph{\bibinfo{journal}{Nature Reviews Methods Primers}} \textbf{\bibinfo{volume}{4}}, \bibinfo{pages}{51} (\bibinfo{year}{2024}).

\bibitem{comiter2023inference}
\bibinfo{author}{Comiter, C.} \emph{et~al.}
\newblock \bibinfo{title}{Inference of single cell profiles from histology stains with the single-cell omics from histology analysis framework (schaf)}.
\newblock \emph{\bibinfo{journal}{BioRxiv}} \bibinfo{pages}{2023--03} (\bibinfo{year}{2023}).

\bibitem{chadoutaud2024scellst}
\bibinfo{author}{Chadoutaud, L.} \emph{et~al.}
\newblock \bibinfo{title}{scellst: a multiple instance learning approach to predict single-cell gene expression from h\&e images using spatial transcriptomics}.
\newblock \emph{\bibinfo{journal}{bioRxiv}} \bibinfo{pages}{2024--11} (\bibinfo{year}{2024}).

\bibitem{erickson2022spatially}
\bibinfo{author}{Erickson, A.} \emph{et~al.}
\newblock \bibinfo{title}{Spatially resolved clonal copy number alterations in benign and malignant tissue}.
\newblock \emph{\bibinfo{journal}{Nature}} \textbf{\bibinfo{volume}{608}}, \bibinfo{pages}{360--367} (\bibinfo{year}{2022}).

\bibitem{marklund2022spatio}
\bibinfo{author}{Marklund, M.} \emph{et~al.}
\newblock \bibinfo{title}{Spatio-temporal analysis of prostate tumors in situ suggests pre-existence of treatment-resistant clones}.
\newblock \emph{\bibinfo{journal}{Nature Communications}} \textbf{\bibinfo{volume}{13}}, \bibinfo{pages}{5475} (\bibinfo{year}{2022}).

\bibitem{nunes2024prognostic}
\bibinfo{author}{Nunes, L.} \emph{et~al.}
\newblock \bibinfo{title}{Prognostic genome and transcriptome signatures in colorectal cancers}.
\newblock \emph{\bibinfo{journal}{Nature}} \textbf{\bibinfo{volume}{633}}, \bibinfo{pages}{137--146} (\bibinfo{year}{2024}).

\bibitem{min2024multimodal}
\bibinfo{author}{Min, W.}, \bibinfo{author}{Shi, Z.}, \bibinfo{author}{Zhang, J.}, \bibinfo{author}{Wan, J.} \& \bibinfo{author}{Wang, C.}
\newblock \bibinfo{title}{Multimodal contrastive learning for spatial gene expression prediction using histology images}.
\newblock \emph{\bibinfo{journal}{Briefings in Bioinformatics}} \textbf{\bibinfo{volume}{25}}, \bibinfo{pages}{bbae551} (\bibinfo{year}{2024}).

\bibitem{filiot2025distilling}
\bibinfo{author}{Filiot, A.} \emph{et~al.}
\newblock \bibinfo{title}{Distilling foundation models for robust and efficient models in digital pathology}.
\newblock \emph{\bibinfo{journal}{arXiv preprint arXiv:2501.16239}}  (\bibinfo{year}{2025}).

\bibitem{ganin2015unsupervised}
\bibinfo{author}{Ganin, Y.} \& \bibinfo{author}{Lempitsky, V.}
\newblock \bibinfo{title}{Unsupervised domain adaptation by backpropagation}.
\newblock In \emph{\bibinfo{booktitle}{International conference on machine learning}}, \bibinfo{pages}{1180--1189} (\bibinfo{organization}{PMLR}, \bibinfo{year}{2015}).

\bibitem{vaidya2024demographic}
\bibinfo{author}{Vaidya, A.} \emph{et~al.}
\newblock \bibinfo{title}{Demographic bias in misdiagnosis by computational pathology models}.
\newblock \emph{\bibinfo{journal}{Nature Medicine}} \textbf{\bibinfo{volume}{30}}, \bibinfo{pages}{1174--1190} (\bibinfo{year}{2024}).

\bibitem{alayrac2022flamingo}
\bibinfo{author}{Alayrac, J.-B.} \emph{et~al.}
\newblock \bibinfo{title}{Flamingo: a visual language model for few-shot learning}.
\newblock In \bibinfo{editor}{Oh, A.~H.}, \bibinfo{editor}{Agarwal, A.}, \bibinfo{editor}{Belgrave, D.} \& \bibinfo{editor}{Cho, K.} (eds.) \emph{\bibinfo{booktitle}{Advances in Neural Information Processing Systems}} (\bibinfo{year}{2022}).

\bibitem{elosua2021spotlight}
\bibinfo{author}{Elosua-Bayes, M.}, \bibinfo{author}{Nieto, P.}, \bibinfo{author}{Mereu, E.}, \bibinfo{author}{Gut, I.} \& \bibinfo{author}{Heyn, H.}
\newblock \bibinfo{title}{Spotlight: seeded nmf regression to deconvolute spatial transcriptomics spots with single-cell transcriptomes}.
\newblock \emph{\bibinfo{journal}{Nucleic acids research}} \textbf{\bibinfo{volume}{49}}, \bibinfo{pages}{e50--e50} (\bibinfo{year}{2021}).

\bibitem{vaidya2025molecular}
\bibinfo{author}{Vaidya, A.} \emph{et~al.}
\newblock \bibinfo{title}{Molecular-driven foundation model for oncologic pathology}.
\newblock \emph{\bibinfo{journal}{arXiv preprint arXiv:2501.16652}}  (\bibinfo{year}{2025}).

\bibitem{lee2024pathomclip}
\bibinfo{author}{Lee, Y.}, \bibinfo{author}{Liu, X.}, \bibinfo{author}{Hao, M.}, \bibinfo{author}{Liu, T.} \& \bibinfo{author}{Regev, A.}
\newblock \bibinfo{title}{Pathomclip: Connecting tumor histology with spatial gene expression via locally enhanced contrastive learning of pathology and single-cell foundation model}.
\newblock \emph{\bibinfo{journal}{bioRxiv}} \bibinfo{pages}{2024--12} (\bibinfo{year}{2024}).

\bibitem{yang2021multi}
\bibinfo{author}{Yang, K.~D.} \emph{et~al.}
\newblock \bibinfo{title}{Multi-domain translation between single-cell imaging and sequencing data using autoencoders}.
\newblock \emph{\bibinfo{journal}{Nature communications}} \textbf{\bibinfo{volume}{12}}, \bibinfo{pages}{31} (\bibinfo{year}{2021}).

\bibitem{radford2021learning}
\bibinfo{author}{Radford, A.} \emph{et~al.}
\newblock \bibinfo{title}{Learning transferable visual models from natural language supervision}.
\newblock In \emph{\bibinfo{booktitle}{International conference on machine learning}}, \bibinfo{pages}{8748--8763} (\bibinfo{organization}{PMLR}, \bibinfo{year}{2021}).

\bibitem{ding2024multimodalslidefoundationmodel}
\bibinfo{author}{Ding, T.} \emph{et~al.}
\newblock \bibinfo{title}{Multimodal whole slide foundation model for pathology} (\bibinfo{year}{2024}).
\newblock \urlprefix\url{https://arxiv.org/abs/2411.19666}.
\newblock \eprint{2411.19666}.

\bibitem{steinley2004properties}
\bibinfo{author}{Steinley, D.}
\newblock \bibinfo{title}{Properties of the hubert-arable adjusted rand index.}
\newblock \emph{\bibinfo{journal}{Psychological methods}} \textbf{\bibinfo{volume}{9}}, \bibinfo{pages}{386} (\bibinfo{year}{2004}).

\bibitem{shen2024interactive3dmedicalimage}
\bibinfo{author}{Shen, C.}, \bibinfo{author}{Li, W.}, \bibinfo{author}{Shi, Y.} \& \bibinfo{author}{Wang, X.}
\newblock \bibinfo{title}{{Interactive 3D Medical Image Segmentation with SAM 2}} (\bibinfo{year}{2024}).
\newblock \urlprefix\url{https://arxiv.org/abs/2408.02635}.
\newblock \eprint{2408.02635}.

\bibitem{song2022single}
\bibinfo{author}{Song, H.} \emph{et~al.}
\newblock \bibinfo{title}{Single-cell analysis of human primary prostate cancer reveals the heterogeneity of tumor-associated epithelial cell states}.
\newblock \emph{\bibinfo{journal}{Nature communications}} \textbf{\bibinfo{volume}{13}}, \bibinfo{pages}{141} (\bibinfo{year}{2022}).

\bibitem{west2001predicting}
\bibinfo{author}{West, M.} \emph{et~al.}
\newblock \bibinfo{title}{Predicting the clinical status of human breast cancer by using gene expression profiles}.
\newblock \emph{\bibinfo{journal}{Proceedings of the national academy of Sciences}} \textbf{\bibinfo{volume}{98}}, \bibinfo{pages}{11462--11467} (\bibinfo{year}{2001}).

\end{thebibliography}
\end{nolinenumbers}

\end{document}